%% file: main.tex
\documentclass{article}

\PassOptionsToPackage{numbers, compress}{natbib}

\usepackage[preprint]{neurips_2025}

\usepackage[utf8]{inputenc} 
\usepackage[T1]{fontenc}    
\usepackage{hyperref}       
\usepackage{nccmath}
\usepackage{pgfplots}
\pgfplotsset{compat=1.18}
\hypersetup{
     colorlinks=true,
     linkcolor=blue,
     filecolor=blue,
     citecolor=purple,
     urlcolor=blue,
     }
\usepackage{url}
\usepackage{wrapfig}
\usepackage{booktabs}       
\usepackage{amsfonts}       
\usepackage{nicefrac}       
\usepackage{microtype}      
\usepackage{xcolor}         
\usepackage{multicol}
\usepackage{graphicx}
\usepackage{amsthm}
\usepackage{bbm}
\usepackage{comment}
\usepackage{enumitem}
\usepackage{bm}
\usepackage{booktabs}
\usepackage{subfigure}
\usepackage{mathtools}
\usepackage[export]{adjustbox}
\usepackage{stmaryrd}
\usepackage{booktabs} 
\usepackage{xifthen}
\usepackage{xargs}
\usepackage{mathtools}
\usepackage{amsmath,amssymb}
\usepackage{algorithm,algorithmic}
\usepackage{caption}
\usepackage{cleveref}
\usepackage{titlesec}
\usepackage{multirow}
\usepackage{subfigure}
\usepackage{mathrsfs}
\usepackage{colortbl}
\titleformat*{\subparagraph}{\itshape}
\newtheorem{definition}{Definition}
\newtheorem{example}{Example}
\newtheorem{proposition}{Proposition}
\newtheorem*{proprestate}{Proposition} 
\newtheorem{lemma}{Lemma}
\newtheorem{corollary}{Corollary}

\numberwithin{equation}{section}

\usepackage[textsize=tiny]{todonotes}
\usepackage[framemethod=TikZ]{mdframed}
\mdfdefinestyle{propFrame}{%
    linecolor=white,
    outerlinewidth=1pt,
    roundcorner=6pt,
    innertopmargin=5.5pt,
    innerbottommargin=2.5pt,
    innerrightmargin=3.5pt,
    innerleftmargin=3.5pt,
    backgroundcolor=black!3!white
}

\definecolor{AlgoHighlight}{RGB}{255,220,220}

\newcommand\blfootnote[1]{%
  \begingroup
  \renewcommand\thefootnote{}\footnote{#1}%
  \addtocounter{footnote}{-1}%
  \endgroup
}
\newcounter{hypA}
\newenvironment{hypA}{\refstepcounter{hypA}\begin{itemize}
  \item[({\bf A\arabic{hypA}})]}{\end{itemize}}
\input{def.tex}
\renewcommand{\citep}{\cite}
\renewcommand{\citet}{\cite}

\title{Conditional Diffusion Models with Classifier-Free Gibbs-like Guidance}

\author{Badr Moufad$^{*, 1}$
\quad
Yazid Janati$^{*, 1}$ \quad Alain Durmus$^{1}$\\
\hspace{-.4cm} \textbf{
Ahmed Ghorbel$^1$ \quad
Eric Moulines$^{1, 3}$
\quad
Jimmy Olsson$^2$} \\
$^1$ CMAP, Ecole polytechnique \quad $^2$ KTH Royal Institute of Technology \quad $^3$ MBZUAI}

\begin{document}

\maketitle

\blfootnote{* Equal contribution}
\blfootnote{Corresponding authors: \texttt{$\{$badr.moufad, yazid.janati$\}$@polytechnique.edu}}

\begin{abstract}
    Classifier-Free Guidance (CFG) is a widely used technique for improving conditional diffusion models by linearly combining the outputs of conditional and unconditional denoisers. While CFG enhances visual quality and improves alignment with prompts, it often reduces sample diversity, leading to a challenging trade-off between quality and diversity. To address this issue, we make two key contributions.
    First, CFG generally does not correspond to a well-defined denoising diffusion model (DDM). In particular, contrary to common intuition, CFG does not yield samples from the target distribution associated with the limiting CFG score as the noise level approaches zero—where the data distribution is tilted by a power $w>1$ of the conditional distribution. We identify the missing component: a Rényi divergence term that acts as a repulsive force and is required to correct CFG and render it consistent with a proper DDM. Our analysis shows that this correction term vanishes in the low-noise limit.
    Second, motivated by this insight, we propose a Gibbs-like sampling procedure to draw samples from the desired tilted distribution. This method starts with an initial sample from the conditional diffusion model without CFG and iteratively refines it, preserving diversity while progressively enhancing sample quality. We evaluate our approach on both image and text-to-audio generation tasks, demonstrating substantial improvements over CFG across all considered metrics. The code is available at \url{https://github.com/yazidjanati/cfgig}
\end{abstract}
    
    \section{Introduction}
    Diffusion models \cite{sohl2015deep,song2019generative,ho2020denoising} have emerged as a powerful framework for generative modeling, achieving state-of-the-art performance in a variety of tasks such as text-to-image generation \cite{rombach2022high,podell2023sdxl}, video \cite{blattmann2023align} and audio generation \cite{kong2020diffwave}. 
    The success of these models can be partly attributed to their ability to produce powerful conditional generative models through guidance. Among various methods, classifier-free guidance (CFG) \cite{ho2022classifier} has become a very popular method for sample generation, as it allows strengthening the alignment to the conditioning context via a temperature-like parameter $\scale > 1$ that acts as a guidance scale. Beyond improving alignment, CFG plays a crucial role in ensuring high-quality samples, as unguided diffusion models typically produce subpar outputs, limiting their practical use \cite{dieleman2022guidance,karras2024guiding}. 

    CFG is implemented by linearly combining the outputs of a conditional denoiser---one that takes the conditioning context as additional input—and an unconditional denoiser, with the combination controlled by a scaling parameter $\scale$.
    Previous works have demonstrated that although this simple combination effectively enhances the performance of diffusion models, it often results in overly simplistic images and a substantial reduction in output diversity \cite{karras2024guiding,kynkanniemi2024applying}. This is essentially due to the fact that a linear combination of the conditional and unconditional denoisers does not yield a valid denoiser, thereby breaking the correspondence with any underlying diffusion process \cite{karras2024guiding,bradley2024classifier,chidambaram2024what}.  
    
\begin{figure}[t]
    \vspace*{-4mm}
    \centering
    \begin{subfigure}
        \centering
        \includegraphics[width=0.46\textwidth]{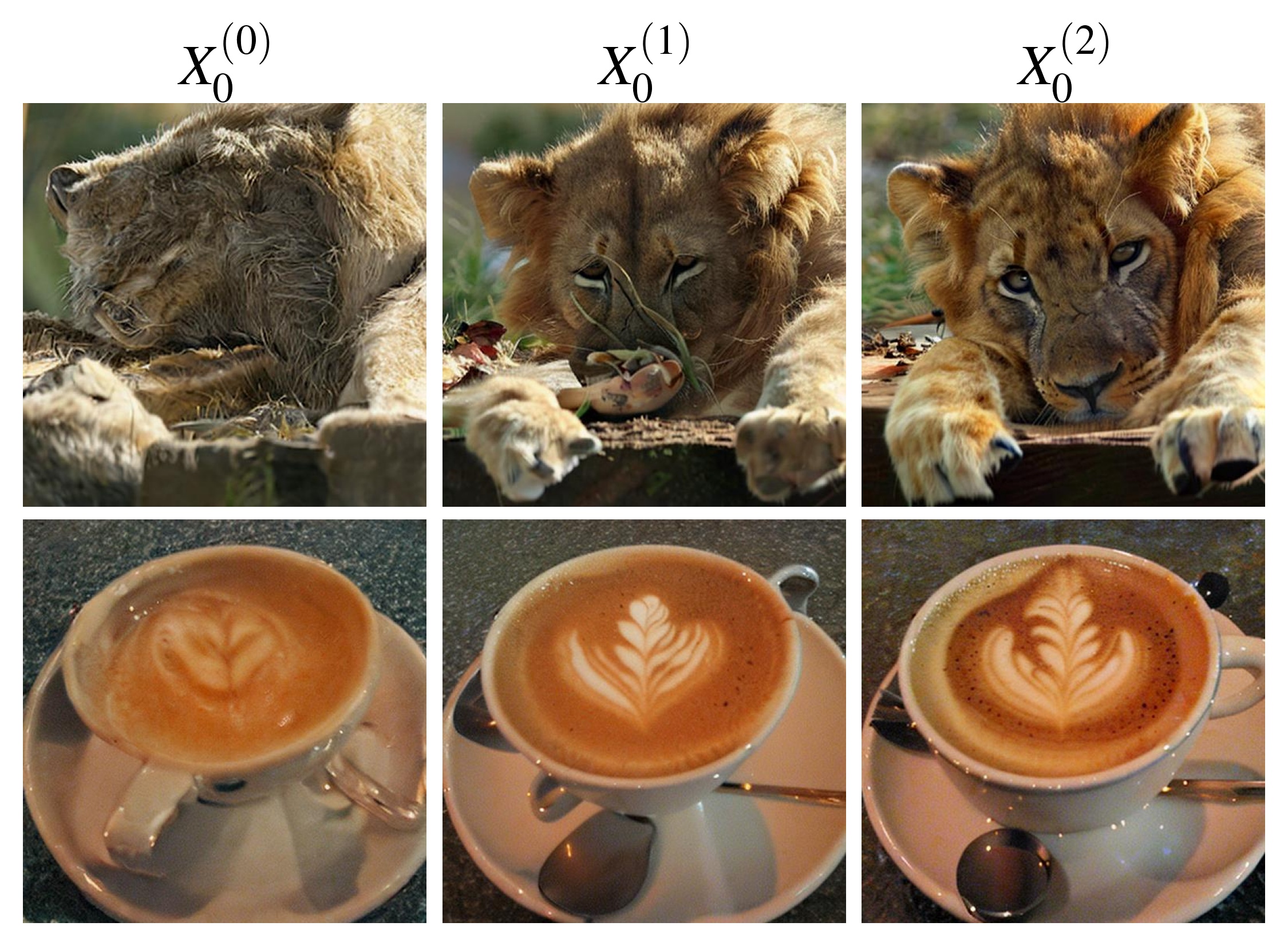}
    \end{subfigure}
    \begin{subfigure}
        \centering
        \includegraphics[width=0.46\textwidth]{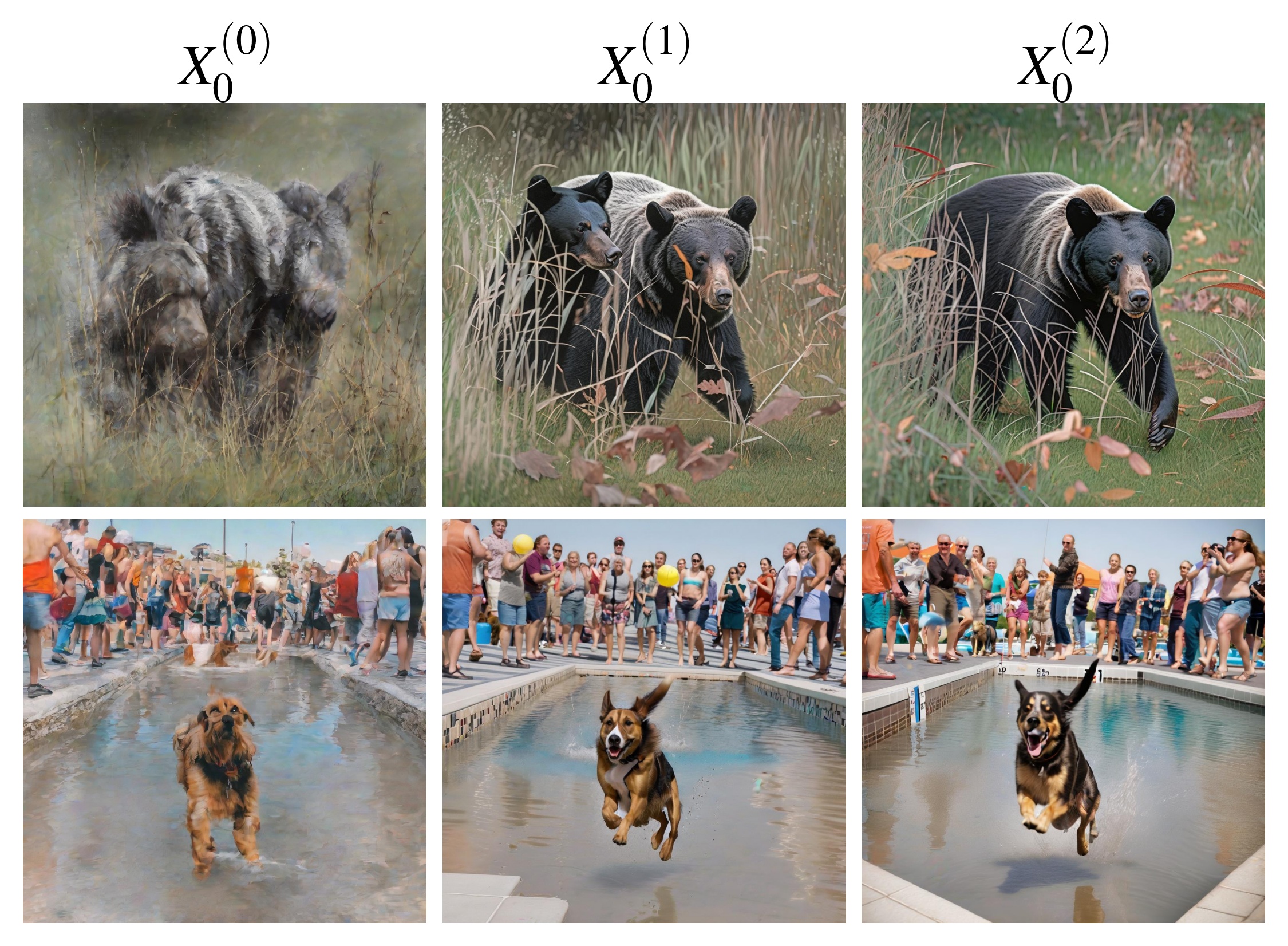}
    \end{subfigure}
    \captionsetup{font=small}
    \caption{Illustration of sample refinement across Gibbs iterations.
    Left, samples generated using EDM-XXL for two \imagenet\ classes: $291$ (top) and $967$ (bottom).
    Right, samples generated using Stable Diffusion XL (SDXL) for the prompts: \textit{``A black bear walking in the grass and leaves.''} (top) and
    \textit{``A dog jumping through the air above a pool of water that has been marked for distance, with people watching in the distance.''} (bottom).
    Each row displays an initial sample $\iX{0}{0}$ alongside two subsequent iterates $\iX{1}{0}, \iX{2}{0}$.
    }
    \label{fig:gibbs_reps}
    \vspace*{-2mm}
\end{figure}

\paragraph{Contributions.}  The aim of this paper is to propose a procedure that enhances the quality of samples generated by conditional diffusion models without compromising output diversity. To this end, we revisit guidance from a probabilistic perspective, focusing on the problem of sampling from the data distribution tilted by the conditional distribution of the context given the data raised to some power $w>1$. Our first contribution is to show that the denoisers used in CFG are neither posterior-mean estimators for this tilted target distribution nor valid denoisers for any other data distribution; see \Cref{ex:gaussian-cfg}. We then demonstrate (\Cref{prop:tilted-score}) that for CFG to target the tilted distribution of interest, it must include an additional term---the gradient of a R\'enyi divergence---which acts as a repulsive force that encourages output diversity, as illustrated in \Cref{fig:illustration-renyi}.  
Our analysis further shows that this term vanishes in the low-noise regime and can thus be safely discarded (\Cref{prop:renyi-expansion}). This insight motivates our second contribution: an iterative sampling procedure, akin to a Gibbs sampler \cite{gelfand2000gibbs} and referred to as \emph{Classifier-Free Gibbs-Like Guidance} (\algoname), which generates approximate samples from the  tilted distribution. The proposed algorithm (\Cref{algo:algo}) is straightforward and begins by first drawing a sample from the conditional distribution, without CFG, then iteratively refines it via  alternating noising and CFG denoising steps. 
A key advantage of this approach is its ability to preserve the sample diversity of the prior model while improving generation quality.
Some examples illustrating the iterative refinement procedure are found in \Cref{fig:gibbs_reps}. We also analyze the algorithm in a simple Gaussian setting, where we are able to quantify the bias introduced by omitting the R\'enyi divergence gradient (\Cref{prop:variance-bias}). Finally, as an additional contribution of independent interest, we derive a new expression for the tilted target scores that involves two different noise levels, providing a theoretical  justification for recent guidance methods \cite{sadat2025no,li2024self}. 

We validate \algoname\ on image generation using \imagenet-$512$ \cite{karras2024edm2} and text-to-audio generation \citet{liu2024audioldm2},
showing that it significantly outperforms CFG and is competitive with the state-of-the art method proposed recently in  \citet{kynkanniemi2024applying}.

    \section{Background}
    \label{sec:background}
    \input{background}
    \section{Gibbs-like guidance}
    \label{sec:gibbs-like-guidance}
    \input{cf2g.tex}
    \input{experiments.tex}

    \vspace*{-1mm}
    \section{Conclusion}
    \vspace*{-1mm}
    We have provided a detailed analysis of CFG and identified a crucial missing term---the gradient of a Rényi divergence---which naturally promotes diversity in generated samples. Our theoretical analysis demonstrates that this additional term becomes negligible in the low-noise regime, motivating a new sampling algorithm, \algoname, that leverages this insight. Through extensive experiments on both image and audio generation tasks, we have confirmed the effectiveness and practical benefits of the proposed method.
    The insights gained in this work suggest several promising directions for future research. One especially promising avenue involves utilizing our newly derived score expressions to design novel training procedures for conditional diffusion models. In particular, deriving an efficient and stable training loss to explicitly account for the Rényi divergence term could alleviate the need for heuristic hyperparameter tuning—such as selecting noise thresholds ($\stds$ in our case or $\sigma_\text{hi}$ in \limitedCFG)—and enable the application of guidance across the full sampling interval without sacrificing diversity or perceptual quality.
\vspace{-.3cm}
    \paragraph{Broader impact.} The proposed method improves both sample quality and diversity in conditional diffusion models, offering promising applications in content creation and accessibility. However, it also raises concerns regarding potential misuse for generating synthetic media, underscoring the importance of developing reliable detection methods and responsible deployment practices.

    \section*{Acknowledgements}
    The authors thank Gabriel V. Cardoso for insightful discussions and contributions during the initial phase of this project. The work of Y.J. and B.M. has been supported by Technology Innovation Institute (TII), project Fed2Learn. The work of Eric Moulines has been partly funded by the European Union (ERC-2022-SYG-OCEAN-101071601). Views and opinions expressed are however those of the author(s) only and do not necessarily reflect those of the European Union or the European Research Council Executive Agency. Neither the European Union nor the granting authority can be held responsible for them. This work was granted access to the HPC resources of IDRIS under the allocations 2025-AD011015980 and 2025-AD011016484 made by GENCI.
\bibliographystyle{plain}
\bibliography{bibliography}

\clearpage
\newpage
    \appendix
    \section{Proofs}
    \input{appendix/proofs.tex}
    \label{apdx-sec:proofs}
    \input{appendix/related.tex}
    \section{Experiments details}
    \input{appendix/exp_details.tex}
    \section{Delayed guidance}
    \label{apdx:delayed-guidance}   
    \input{appendix/delayed_guidance.tex}
    \clearpage
    \input{appendix/qualitative.tex}

 \end{document}

%% file: def.tex
\def\rset{\mathbb{R}}
\def\nset{\mathbb{N}}
\def\rmd{\mathrm{d}}

\def\normpdf{\mathrm{N}}
\def\eqsp{\,}
\def\Id{\mathbf{I}}
\def\zero{0}
\def\pV{\mathbb{V}}
\def\pE{\mathbb{E}}
\def\calR{\mathcal{R}}

\def\indic{\mathbbm{1}}
\def\eqdef{\vcentcolon=}

\def\wrt{w.r.t.}

\def\law{\mathsf{Law}}

\newcommand{\pclass}[1]{\mathcal{Q}_3(\rset^{#1})}

\newcommand{\assp}[1]{(\textbf{A}\textbf{#1})}

\def\ie{\textit{i.e.}}

\newcommand{\intset}[2]{\llbracket #1, #2 \rrbracket}

\newcommand{\renyi}[3]{\mathsf{R}_{#1}(#2 \| #3)}

 \def\rset{{\mathbb{R}}}

 \def\rmd{\mathrm{d}}
 \def\eqsp{\;}
 \def\eqdef{\coloneqq}

 \def\cond{\vec{C}}

\def\cfg{\mathsf{cfg}}

\newcommandx\targ[4][4=]{
    \ifthenelse{\equal{#2}{}}{
        \ifthenelse{\equal{#3}{}}{
            p^{#4} _{#1}
        }{
            p^{#4} _{#1}(#3)
            }
    }{
        \ifthenelse{\equal{#3}{}}{
            p^{#4} _{#1}(\cdot|#2)
        }{
            p^{#4} _{#1}(#3|#2)
        }
    }}

\newcommandx\interp[4][4=\interpscale]{
    \ifthenelse{\equal{#2}{}}{
        \ifthenelse{\equal{#3}{}}{
            \pi^{#4} _{#1}
        }{
            \pi^{#4} _{#1}(#3)
            }
    }{
        \ifthenelse{\equal{#3}{}}{
            \pi^{#4} _{#1}(\cdot|#2)
        }{
            \pi^{#4} _{#1}(#3|#2)
        }
    }}

\newcommand\pdata[3]{
    \ifthenelse{\equal{#2}{}}{
        \ifthenelse{\equal{#3}{}}{
            p_{#1}
        }{
            p_{#1}(#3)
        }
    }{
        \ifthenelse{\equal{#3}{}}{
            p_{#1}(\cdot|#2)
        }{
            p_{#1}(#3|#2)
        }
    }}

\newcommand\jpdata[3]{
    \ifthenelse{\equal{#2}{}}{
        \ifthenelse{\equal{#3}{}}{
            \bar{p}_{#1}
        }{
            \bar{p}_{#1}(#3)
        }
    }{
        \ifthenelse{\equal{#3}{}}{
            \bar{p}_{#1}(\cdot|#2)
        }{
            \bar{p}_{#1}(#3|#2)
        }
    }}

\newcommandx\cpdata[4][4=]{
    \ifthenelse{\equal{#2}{}}{
        \ifthenelse{\equal{#3}{}}{
            \pi^{#4} _{#1}
        }{
            \pi^{#4} _{#1}(#3)
        }
    }{
        \ifthenelse{\equal{#3}{}}{
            \pi^{#4} _{#1}(\cdot|#2)
        }{
            \pi^{#4} _{#1}(#3|#2)
        }
    }}

  \newcommandx\potg[4][4=]{
    \ifthenelse{\equal{#2}{}}{
        \ifthenelse{\equal{#3}{}}{
            g^{#4} _{#1}
        }{
            g^{#4} _{#1}(#3)
        }
    }{
        \ifthenelse{\equal{#3}{}}{
            g^{#4} _{#1}(\cdot|#2)
        }{
            g^{#4} _{#1}(#3|#2)
        }
    }}

\newcommandx\jcpdata[4][4=]{
    \ifthenelse{\equal{#2}{}}{
        \ifthenelse{\equal{#3}{}}{
            \bar\pi^{#4} _{#1}
        }{
            \bar\pi^{#4} _{#1}(#3)
        }
    }{
        \ifthenelse{\equal{#3}{}}{
            \bar\pi^{#4} _{#1}(\cdot|#2)
        }{
            \bar\pi^{#4} _{#1}(#3|#2)
        }
    }}

\newcommandx\tilt[4][4=]{
    \ifthenelse{\equal{#2}{}}{
        \ifthenelse{\equal{#3}{}}{
            \pi^{#4} _{#1}
        }{
            \pi^{#4} _{#1}(#3)
        }
    }{
        \ifthenelse{\equal{#3}{}}{
            \pi^{#4} _{#1}(\cdot|#2)
        }{
            \pi^{#4} _{#1}(#3|#2)
        }
    }}

\newcommand\updata[3]{
    \ifthenelse{\equal{#2}{}}{
        \ifthenelse{\equal{#3}{}}{
            \bar{p}_{#1}
        }{
            \bar{p}_{#1}(#3)
        }
    }{
        \ifthenelse{\equal{#3}{}}{
            \bar{p}_{#1}(\cdot|#2)
        }{
            \bar{p}_{#1}(#3|#2)
        }
    }}

\newcommand\phat[3]{
    \ifthenelse{\equal{#2}{}}{
        \ifthenelse{\equal{#3}{}}{
            \hat{p}_{#1}
        }{
            \hat{p}_{#1}(#3)
        }
    }{
        \ifthenelse{\equal{#3}{}}{
            \hat{p}_{#1}(\cdot|#2)
        }{
            \hat{p}_{#1}(#3|#2)
        }
    }}
\newcommandx\fw[4][4=]{
        \ifthenelse{\equal{#3}{}}{
            q^{#4} _{\smash{#1}}(\cdot|#2)
        }{
            q^{#4} _{\smash{#1}}(#3|#2)
        }
    }
\newcommandx\denoiser[4][4=]{
    \ifthenelse{\equal{#2}{}}{
        \ifthenelse{\equal{#3}{}}{
            D^{#4}_{#1}
        }{
            D^{#4} _{#1}(#3)
        }
    }{
        \ifthenelse{\equal{#3}{}}{
            D^{#4} _{#1}(\cdot|#2)
        }{
            D^{#4} _{#1}(#3|#2)
        }
    }}

    \newcommandx\hdenoiser[4][4=]{
    \ifthenelse{\equal{#2}{}}{
        \ifthenelse{\equal{#3}{}}{
            \hat{D}^{#4}_{#1}
        }{
            \hat{D}^{#4} _{#1}(#3)
        }
    }{
        \ifthenelse{\equal{#3}{}}{
            \hat{D}^{#4} _{#1}(\cdot|#2)
        }{
            \hat{D}^{#4} _{#1}(#3|#2)
        }
    }}

\newcommandx\epspred[4][4=]{
    \ifthenelse{\equal{#2}{}}{
        \ifthenelse{\equal{#3}{}}{
            \varepsilon^{#4}_{#1}
        }{
            \varepsilon^{#4} _{#1}(#3)
        }
    }{
        \ifthenelse{\equal{#3}{}}{
            \varepsilon^{#4} _{#1}(\cdot|#2)
        }{
            \varepsilon^{#4} _{#1}(#3|#2)
        }
    }}

\newcommand\clf[3]{
    \ifthenelse{\equal{#2}{}}{
        g_{#1}(#3|\cdot)
    }{
        g _{#1}(#3|#2)
    }
}

\newcommand\cfgdist[3]{
    \ifthenelse{\equal{#2}{}}{
        p^w _{#1}(#3|\cdot)
    }{
        p^w _{#1}(#3|#2)
    }
}

\newcommand\cscore[3]{
    \ifthenelse{\equal{#2}{}}{
        \ifthenelse{\equal{#3}{}}{
            s _{#1}
        }{
            s _{#1}(#3)
        }
    }{
        \ifthenelse{\equal{#3}{}}{
            s _{#1}(\cdot|#2)
        }{
            s _{#1}(#3|#2)
        }
    }}

\def\gauss{\mathcal{N}}

\def\nulltok{\varnothing}

\def\scale{w}
\def\interpscale{\eta}
\newcommand{\stdmax}{\sigma_{\tiny{\mbox{max}}}}
\def\std{\sigma}
\def\stdmax{\std_\text{max}}
\def\stds{{\sigma_*}}
\def\stdplus{{\sigma_{\tiny\mathmbox{+}}}}
\def\stdminus{{\sigma_{\tiny\mathmbox{-}}}}
\def\vars{{\sigma^2 _*}}

\def\bx{\mathbf{x}}
\def\by{\mathbf{y}}
\newcommand{\iX}[2]{\smash{X^{\tiny\mathmbox{(#1)}} _{#2}}}
\def\bz{\mathbf{z}}
\def\bX{X}
\def\cond{\mathbf{c}}

\def\algoname{{\sc{CFGiG}}}
\def\delayedcfg{{\sc{D-CFGiG}}}
\def\CFGPP{{\sc{CFG++}}}
\def\CFG{{\sc{CFG}}}
\def\limitedCFG{{\sc{LI-CFG}}}

\def\imagenet{{\texttt{ImageNet}}}
\def\audiocaps{{\texttt{AudioCaps}}}

\def\param{\theta}

\def\dimx{{d}}

\newcommandx{\hpredx}[3][2=0,3=\param]{\smash{m^{#3} _{#2|#1}}}
\newcommandx{\predx}[2][2=0]{\smash{m _{#2|#1}}}
\newcommandx{\prednoise}[2][2=\param]{\smash{\epsilon^{#2} _{#1}}}
\newcommandx{\score}[2][2=\param]{s^{#2} _{#1}}

\newcommand{\ccint}[1]{\left[#1\right]}

\def\constNc{\mathscr{Z}(\cond)}
\def\Leb{\mbox{Leb}}
\def\msa{\mathsf{A}}
\def\inv{\leftarrow}
\renewcommand{\det}{\mbox{det}}

%% file: background.tex
\paragraph{Diffusion models.} Denoising diffusion models (DDMs) \cite{song2019generative,song2021ddim,song2021score} define a generative procedure targeting a data distribution $\pdata{0}{}{}$. It consists in first sampling from a highly noised distribution $\pdata{\stdmax}{}{}$, that practically resembles a Gaussian, and then iteratively sampling through a sequence of progressively less noisy distributions $\pdata{\std}{}{}$ for decreasing noise levels $\std < \std_\text{max}$.
The distribution $\pdata{\std}{}{}$ is the $\bx_\std$-marginal of $\pdata{0,\std}{}{\bx_0, \bx_\std}$,
the joint distribution of the $(X_0, X_\std) \eqdef (X_0,X_0 + \sigma Z)$,
where $\bX_0$ and $Z$ are drawn independently from $\pdata{0}{}{}$ and $\gauss(0, \Id)$.
Formally, by letting 
$\smash{\fw{\std|0}{\bx_0}{\bx_\std}} \eqdef \normpdf(\bx_\std; \bx_0, \std^2 \Id)$,
the marginals are defined as
$
    \pdata{\std}{}{\bx_\std}
    \eqdef \int \pdata{0,\sigma}{}{\bx_0,\bx_{\sigma}}  \, \rmd \bx_0
    = \int \fw{\std|0}{\bx_0}{\bx_\std} \pdata{0}{}{\bx_0} \, \rmd \bx_0
$. 
Following \cite{song2021score,song2021ddim,karras2022elucidating}, exact sampling of the target $\pdata{0}{}{}$ can be performed by solving backwards in time over $\ccint{0,\stdmax}$
and starting from $\bx_{\stdmax} \sim \pdata{\stdmax}{}{}$,
\begin{equation}
    \label{eq:pf-ode}
    \rmd \bx_\std / \rmd \std = \big(\bx_\std - \denoiser{\std}{}{\bx_\std}\big)/\std \eqsp, \quad
    \denoiser{\std}{}{\bx_\std} = \int \bx_0 \, \pdata{0|\std}{\bx_\std}{\bx_0} \, \rmd \bx_0
    \eqsp,
\end{equation}
where $\denoiser{\std}{}{}$ is the denoiser at noise level $\std$; 
see, e.g. \cite[Eq. 1 and 3]{karras2022elucidating}. The drift in \eqref{eq:pf-ode} can be identified as the score function $(\bx,\sigma) \mapsto \nabla \log \pdata{\std}{}{\bx}$, since using Tweedie's formula \cite{robbins1956empirical}, it follows that $\denoiser{\std}{}{\bx} = \bx + \std^2 \nabla \log \pdata{\std}{}{\bx}$.

The practical implementation of this generative process involves first estimating $\denoiser{\std}{}{}$ using  parametric approximations $\denoiser{\std}{}{}[\param]$, $\theta \in \Theta$, trained by minimizing the weighted denoising loss
$
 \theta \mapsto \int \pE \left[ \| X_0 - \denoiser{\sigma}{}{X_\std}[\param]\|^2 \right] \lambda(\sigma) \, \rmd \sigma 
$,
and $\lambda$ denotes a probability density function over $\rset_+$ that assigns weights to noise levels \cite{karras2022elucidating}. This training procedure also  provides parametric approximations of the score function. Therefore, 
once $(\denoiser{\std}{}{}[\param])_{\std \geq 0}$ are trained, 
a new approximate sample $\smash{\hat{X}_0}$ from $\pdata{0}{}{}$ can be drawn by fixing a decreasing sequence of noise levels $(\std_t)_{t=T}^0$ and then solving the ODE \eqref{eq:pf-ode} backward from $\std_T$ to $\std_0$ using an integration method such as Euler or Heun \cite{karras2022elucidating} starting from $\smash{\hat{X}_{\std_T} \sim \gauss(\zero, \std_T^2 \Id)}$.
In particular, Euler method corresponds to DDIM scheme \cite{song2021ddim} with updates 
\begin{equation}
    \label{eq:ddim}
    \hat{X}_{\std_t}= \left(1 - \std_t / \std_{t+1} \right) \denoiser{\std_{t+1}}{}{\hat{X}_{\std_{t+1}}}[\param] + \big(\std_t / \std_{t+1}\big) \hat{X}_{\std_{t+1}} \eqsp.
\end{equation}

\paragraph{Conditional DDMs.} This generative procedure also extends to sampling conditionally on $C \in \mathcal{C}$, where $\mathcal{C}$ can be a collection of classes or text-prompts.
Denote by $\jcpdata{0}{}{\cond,\bx_0}$ the joint density of $(C,X_0)$.
The goal is to approximately sample from the conditional distribution
$
\cpdata{0}{\cond}{\bx_0} \propto \jcpdata{0}{}{\cond,\bx_0},
$
given training samples from the joint distribution. Similar to the unconditional case, we introduce the joint distribution 
$\jcpdata{0,\std}{}{\cond, \bx_0,\bx_\std} \propto \jcpdata{0}{}{\cond,\bx_0}\,\fw{\std|0}{\bx_0}{\bx_\std}
$, where only the data $X_0$ is noised. 
Conditional DDM sampling from $\cpdata{0}{\cond}{\bx_0}$ thus reduces to estimating the conditional denoiser defined by
$
\smash{\denoiser{\std}{\cond}{\bx_\std} \eqdef \int \bx_0\, \cpdata{0|\std}{\bx_\std, \cond}{\bx_0}\,\rmd \bx_0},
$
by using a parametric family $\{(\bx,\cond,\std) \mapsto \denoiser{\std}{\cond}{\bx}[\param] \,: \, \param \in \Theta \}$ and minimizing the loss 
$
\theta \mapsto \int \pE \left[\|X_0 - \denoiser{\std}{C}{X_\std}[\param]\|^2\right]\lambda(\std)\,\rmd \std,
$
where the expectation is taken \wrt\ the joint law of $(C,X_0, X_\std) \sim \jcpdata{0, \std}{}{}$.
As noted in \cite{ho2022classifier,karras2024edm2}, it is possible to learn simultaneously a conditional and unconditional denoiser by augmenting $\mathcal{C}$ with a null context $\nulltok$ to represent the unconditional case. Then,
during training, $(C, \bX_0)$ is first sampled from $\jcpdata{0}{}{}$ and $C$ is replaced with $\nulltok$ with probability $p_\text{uncd}$.
Henceforth, we denote the unconditional denoisers by $\denoiser{\std}{\nulltok}{} = \denoiser{\std}{}{}$.

\paragraph{Classifier-free guidance (CFG).} In many complex applications---such as text-to-image synthesis and audio generation--- directly using the conditional diffusion models often results in samples that lack the perceptual quality of the training data. This discrepancy is especially evident in terms of  perceived realism, texture fidelity, and fine-grained detail. CFG~\cite{ho2022classifier} has emerged as a standard approach to mitigate this issue, enhancing both the visual fidelity and the alignment of generated samples with their conditioning prompts. 
However, this improvement typically comes at the cost of reduced sample diversity. In CFG, a linear combination of the conditional and conditional denoisers
\begin{equation}
    \label{eq:cfg-denoiser}
    \denoiser{\std}{\cond; \scale}{\bx_\std}[\cfg] \eqdef \scale \denoiser{\std}{\cond}{\bx_\std} + (1 - \scale) \denoiser{\std}{}{\bx_\std}
    \eqsp,
\end{equation}
is used where $\scale > 1$ is a \emph{guidance scale}.
To illustrate the impact of $\scale$,  we denote by $\clf{0}{\bx_0}{\cond} \eqdef \jcpdata{0}{}{\cond,\bx_0} / \pdata{0}{}{\bx_0}$  the conditional distribution of the context  
and set $\clf{\std}{\bx_\std}{\cond} \eqdef \int \clf{0}{\bx_0}{\cond}\,\pdata{0|\std}{\bx_\std}{\bx_0}\,\rmd \bx_0$.
By construction, the conditional distribution of $X_\std$ given the context can be expressed as
\[
\cpdata{\std}{\cond}{\bx_\std}  
\propto \int \jcpdata{0,\std}{}{\cond,\bx_0,\bx_\std} \, \rmd \bx_0 
\propto \int \clf{0}{\bx_0}{\cond}\,\pdata{0}{}{\bx_0}\,\fw{\std|0}{\bx_0}{\bx_\std} \, \rmd \bx_0 
\propto
\clf{\std}{\bx_\std}{\cond}\,\pdata{\std}{}{\bx_\std} \eqsp,
    \]
where we used that $\pdata{0}{}{\bx_0} \fw{\std|0}{\bx_\std}{\bx_0}= \pdata{\std}{}{\bx_\std} \pdata{0|\std}{\bx_\std}{\bx_0}$. Thus, Tweedie's formula implies that
\begin{align}
\label{eq:cond-denoiser-tweedie}
\denoiser{\std}{\cond}{\bx_\std}  = \bx_\std + \std^2 \nabla \log \cpdata{\std}{\cond}{\bx_\std} 
 = \bx_\std + \std^2 \big( \nabla \log \clf{\std}{\bx_\std}{\cond} + \nabla \log \pdata{\std}{}{\bx_\std}\big) \eqsp.
\end{align}
Substituting back into \eqref{eq:cfg-denoiser}, we obtain
$
\denoiser{\std}{\cond; \scale}{\bx_\std}[\cfg] 
= \bx_\std + \std^2 \nabla \log \cpdata{\std}{\cond; \scale}{\bx_\std}[\cfg]
$, 
where 
\begin{equation}
    \label{eq:cfg-marginal}
    \cpdata{\std}{\cond; \scale}{\bx_\std}[\cfg] \propto 
     \clf{\std}{\bx_\std}{\cond}^\scale \pdata{\std}{}{\bx_\std} \eqsp.
\end{equation}
 Hence, CFG modifies the conditional denoiser \eqref{eq:cond-denoiser-tweedie} solely through the guidance scale $\scale$ applied to the classifier score.
With $\scale > 1$, CFG amplifies the influence of regions where $\clf{\std}{\bx_\std}{\cond}$ is large. In practice, this results in enhanced prompt alignment and improved perceptual quality of generated images, at the cost of reduced sample diversity \cite{kynkanniemi2024applying}. 

Note that generally, $\cpdata{\std}{\cond; \scale}{\bx_\std}[\cfg] \neq \int \fw{\std|0}{\bx_0}{\bx_\std} \cpdata{0}{\cond; \scale}{\bx_0}[\cfg] \, \rmd \bx_0$. 
This discrepancy prompts a fundamental question: does incorporating the CFG denoiser \eqref{eq:cfg-denoiser} into the  ODE \eqref{eq:pf-ode} yield a sampling process that corresponds to a well-defined DDM? Specifically, does there exist $\pi(\cdot | \cond; \scale)$ such that for all $\std > 0$, 
    $
 \cpdata{\std}{\cond; \scale}{\bx_\std}[\cfg] = \int \fw{\std|0}{\bx_0}{\bx_\std} \pi(\bx_0|\cond; \scale) \, \rmd \bx_0
    $? 
    We show in the next example that  this is not the case in general.
    \begin{example}
    \label{ex:gaussian-cfg}
        Let $\pdata{0}{}{} = \gauss(0, 1)$ and $\clf{0}{\bx_0}{\cond} = \normpdf(\cond; \bx_0, \gamma^2)$ is the one-dimensional Gaussian density with mean $\bx_0$ and variance $\gamma^2$; 
        then $\cpdata{\std}{\cond; \scale}{}[\cfg]$ is Gaussian with variance 
        \begin{equation}
            \label{eq:cov_zsig}
            \mathrm{v}(\scale,\sigma^2)\eqdef \frac{(1 + \std^2)\gamma^2 + \std^2}{\scale/(1 + \std^2) + \gamma^2 + \sigma^2 / (1 + \sigma^2)} \eqsp. 
        \end{equation}
        We now show that it does not exist $\pi(\cdot | \cond; \scale)$ such that for all $\std > 0$,
        \begin{equation}
\label{eq:hyp_contradiction}          
            \cpdata{\std}{\cond; \scale}{\bx_\std}[\cfg] = \int \fw{\std|0}{\bx_0}{\bx_\std} \pi(\bx_0|\cond; \scale) \, \rmd \bx_0\eqsp.
        \end{equation}
          The proof is by contradiction. Suppose that \eqref{eq:hyp_contradiction} holds. Then we may show, by letting $\std$ tend to zero, that $\pi(\cdot | \cond; \scale) = \cpdata{0}{\cond; \scale}{}[\cfg]$, where
        $\cpdata{0}{\cond; \scale}{\bx_0}[\cfg] = \normpdf(\bx_0; \scale \cond / (\scale + \gamma^2), \gamma^2 / (\scale + \gamma^2))$
        ; see \Cref{apdx-sec:proofs}. 
        In addition, let $(X_0, X_\sigma)$ be distributed according to the joint distribution with density $\smash{\fw{\std|0}{\bx_0}{\bx_\std} \cpdata{0}{\cond; \scale}{\bx_0}[\cfg]}$.
        Then,  
        $
            \pV(X_\std) 
            = \std^2 + \gamma^2 / (\scale + \gamma^2) .
            \label{eq:cov:bound}
        $
        However, by \eqref{eq:hyp_contradiction}, $X_\std \sim \cpdata{\std}{\cond; \scale}{}[\cfg]$ which implies (see again \Cref{apdx-sec:proofs}) that $\mathrm{v}(\scale,\sigma^2) < \pV(X_\std)$ for all $\std > 0$ and $\scale > 1$, we obtain a contradiction. 
    \end{example}

%% file: cf2g.tex
\begin{figure}[t]
    \vspace*{-3mm}
    \centering
    \includegraphics[width=.8\textwidth]{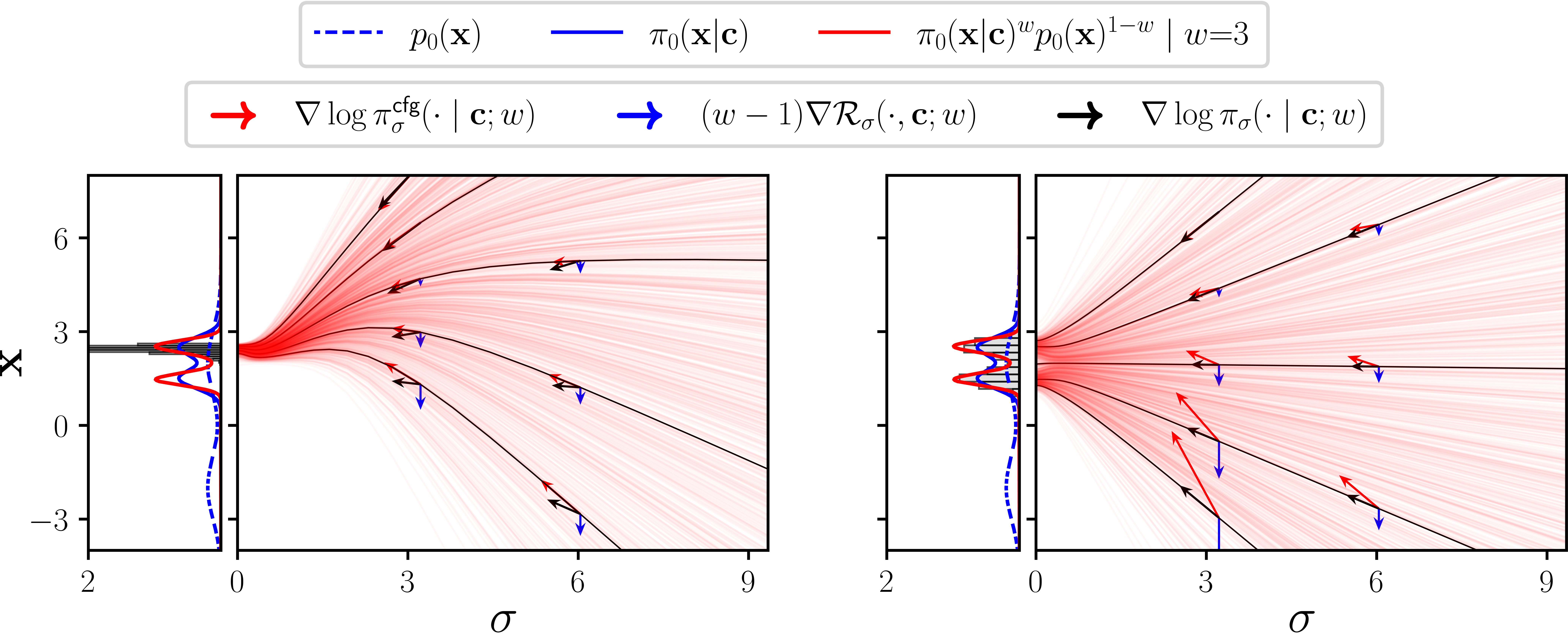}
        \captionsetup{font=small}
        \caption{Left: DDIM sampling with CFG denoiser \eqref{eq:cfg-denoiser}; Right: DDIM sampling with the ideal denoiser \eqref{eq:tilted-score}. The trajectories of $1000$ particles are represented with thin red lines and $5$ selected trajectories are displayed in black thick line along which scores are being depicted with arrows.
        The histogram of the simulated particles is represented in light gray. We also plot the ideal score \eqref{eq:tilted-score} (arrow in black) with the contribution of both the CFG score (arrow in red) and the repulsive term arising from the Rényi divergence (arrow in blue).
        }
        \label{fig:illustration-renyi}
    \end{figure}
    As demonstrated in the previous section, and as we show in \Cref{fig:illustration-renyi}, CFG does not necessarily yield an approximate sample from $\smash{\cpdata{0}{\cond; \scale}{}[\cfg]}$. We now focus on  introducing a DDM specifically designed to target this distribution, 
    which we from now on denote by $\cpdata{0}{\cond; \scale}{}$ (omitting the superscript $\cfg$).
\paragraph{Tilted distribution scores and denoisers.} Define the joint distribution $\jcpdata{0,\std}{\cond; \scale}{\bx_0, \bx_\std} \propto \clf{0}{\bx_0}{\cond}^\scale \pdata{0,\std}{}{\bx_0, \bx_\std}$, of which $\cpdata{0}{\cond; \scale}{}$ is the $\bx_0$-marginal. The next proposition provides a simple and interpretable form of the denoisers and scores of the smoothed marginals $\cpdata{\std}{\cond; \scale}{}=\int \jcpdata{0,\std}{\cond; \scale}{\bx_0, \bx_\std} \rmd \bx_{\std}$. For $\scale > 1$, define the Rényi divergence of order $\scale$ of $p$ from $q$ \cite{van2014renyi} as
\begin{equation}
    \renyi{\scale}{p}{q} \eqdef \frac{1}{\scale - 1} \log \int \frac{p(\bx)^\scale}{q(\bx)^\scale} q(\bx) \, \rmd \bx
    \eqsp.
    \label{eq:def-renyi}
\end{equation}   
For ease of notation, we set $\calR_{\std}(\bx_\std, \cond; \scale) \eqdef \renyi{\scale}{\cpdata{0|\std}{\bx_\std, \cond}{}}{\pdata{0|\std}{\bx_\std}{}}$. 
\begin{mdframed}[style=propFrame]
    \begin{proposition}
        \label{prop:tilted-score}
        For any $\std > 0$,
        the scores associated with $\cpdata{\std}{\cond; \scale}{}$ are
        \begin{equation} 
            \label{eq:tilted-score}
            \nabla \log \cpdata{\std}{\cond; \scale}{\bx_\std} = (\scale - 1) \nabla\, \calR _{\std}(\bx_\std, \cond; \scale) + \nabla \log \cpdata{\std}{\cond; \scale}{\bx_\std}[\cfg] \eqsp. 
        \end{equation} 
    \end{proposition}
    \end{mdframed}
    The proof is postponed to \Cref{subsec-apdx:score-expressions}. The decomposition \eqref{eq:tilted-score} highlights that the CFG score $\nabla \log \cpdata{\sigma}{\cond; \scale}{\bx_\sigma}[\cfg]$
    differs from the \emph{ideal score} $\nabla \log \cpdata{\std}{\cond; \scale}{}$
    by an additional term: the gradient of the Rényi divergence.
    To illustrate \Cref{prop:tilted-score} and the bias introduced by CFG, we consider a one-dimensional toy model (see \Cref{fig:illustration-renyi}). In this example, we compare the trajectories obtained by integrating the ODE \eqref{eq:pf-ode} using the ideal denoiser---defined as $\denoiser{\std}{\cond; \scale}{\bx_\std} \eqdef \bx_\std + \std^2 \nabla \log \cpdata{\std}{\cond; \scale}{\bx_\std}$---and the CFG denoiser \eqref{eq:cfg-denoiser}. The CFG trajectories are observed to collapse 
    onto a single mode, neglecting other regions of significant probability mass, as observed in \cite{kynkanniemi2024applying}. This over-concentration stems from omitting the term $(\scale - 1)\nabla \calR_\std(\bx_\sigma, \cond; \scale)$, which acts as a \emph{repulsive force}. Indeed, since $\scale > 1$, this term
    pushes the sample in the direction where the conditional and unconditional distributions $\cpdata{0|\std}{\bx_\std, \cond}{}$ and $\pdata{0|\std}{\bx_\std}{}$ differ most, thereby counteracting over-concentration. 
    On the other hand, in our next result, we show that the CFG score $\nabla \log \cpdata{\std}{\cond; \scale}{\cdot}[\cfg]$ serves as a good approximation of $ \nabla \log \cpdata{\std}{\cond; \scale}{\cdot}$ for small $\sigma$.
  \begin{mdframed}[style=propFrame]
    \begin{proposition}
        \label{prop:renyi-expansion}
    Under suitable assumptions on $\pdata{0}{}{}$ and $g_0$, 
    it holds for all $\scale > 1$, $\bx \in \rset^\dimx$, $\cond \in \mathcal{C}$,  
    $$
    \nabla \calR _\std(\bx, \cond; \scale) = O(\std^2) \quad \text{as \, $\std \to 0$} \eqsp.
    $$ 
\end{proposition}
\end{mdframed}
  
The CFG denoiser suffers from an intrinsic flaw due to the absence of the term $(\scale - 1)\nabla \calR_\std(\bx_\sigma, \cond; \scale)$, which plays a crucial role at high to medium noise levels in preserving the diversity of the generated samples.
Conversely, the denoiser $\denoiser{\std}{\cond; \scale}{}$ is well approximated by $\denoiser{\std}{\cond; \scale}{}[\cfg]$ in the low-noise regime, where the missing R\'enyi divergence term is effectively negligible.
%
In the following, we present our new  proposed method for generating  approximate samples from the target $\cpdata{0}{\cond; \scale}{\cdot}$, which relies on using the CFG approximation of the denoiser exclusively in the low-noise regime. 
\paragraph{Classifier-free Gibbs-like Guidance.}
\begin{algorithm}[t]
    \caption{\algoname}
    \begin{algorithmic}
       \STATE {\bfseries Require:} Guidance scales $\scale_0 \geq 1$ and $\scale > \scale_0$
       \STATE {\bfseries Require:} Number of repetitions $R$, total steps $T$, and initial steps $T_0$
       \STATE {\bfseries Require:} Standard deviations $\std _*$, $\stdmax$
       \STATE $k \gets T - T_0 \bmod R$
       \STATE $X_{\stdmax} \sim \gauss(\zero, \stdmax^2 \Id)$
       \STATE $X_0 \gets \texttt{ODE\_Solver}(X_{\stdmax}, \denoiser{\std}{\cond; \textcolor{purple}{\scale_0}}{}[\cfg], T_0 + k)$
       \FOR{$r = 1$ {\bfseries to} $R$}
            \STATE $Z \sim \gauss(\zero, \Id)$
            \STATE $X _{\std_*} \gets X _0 + \std_* Z$ 
            \STATE $X_0 \gets \texttt{ODE\_Solver}(X_{\std_*}, \denoiser{\std}{\cond; \textcolor{purple}{\scale}}{}[\cfg], \lfloor 
            (T - T_0)/R
            \rfloor )$
       \ENDFOR
     \STATE {\bfseries Output:} $X_0$
    \end{algorithmic}
    \label{algo:algo}
\end{algorithm}
Our method consists in defining a Markov chain $(\iX{r}{0})_{r \in \nset}$ that admits $\cpdata{0}{\cond; \scale}{}$ as stationary distribution. We rely on a fixed noise level $\stds$ that we assume to be small enough. The chain is generated recursively as follows. 
Given $\iX{r}{0}$ at stage $r$, $\iX{r+1}{0}$ is obtained by 
\begin{enumerate}[wide, labelwidth=!, labelindent=1em,label=Step \arabic*)]
    \item \label{ref:step_1_gibbs_like} sampling an intermediate state $\iX{r}{\stds} \sim \fw{\stds|0}{\iX{r}{0}}{}$;
    \item \label{ref:step_2_gibbs_like} denoising it by integrating the PF-ODE \eqref{eq:pf-ode} with the ideal denoiser $\denoiser{\std}{\cond; \scale}{}$.
\end{enumerate}
These updates can be compactly written as $\iX{r+1}{0} = F_{0|\stds}\big(\iX{r}{0} + \stds Z^{(r+1)}; \scale\big)$, where $(Z^{(r)})_{r \in \nset}$ is a  sequence of i.i.d. standard Gaussian random variables and $F_{0|\stds}(\bx_\stds; \scale)$ is the solution to the PF-ODE with  denoiser $\denoiser{\std}{\cond; \scale}{}$ and initial condition $\bx_\stds$. We can verify that the associated Markov chain admits $\cpdata{0}{\cond; \scale}{}$ as its \emph{unique} stationary distribution under appropriate conditions; see \Cref{apdx-sec:proofs}. Taking $\stds$ small enough, we implement this scheme by using the CFG denoiser \eqref{eq:cfg-denoiser} instead of $\denoiser{\std}{\cond; \scale}{}$ in \ref{ref:step_2_gibbs_like} for integrating the PF-ODE \eqref{eq:pf-ode}. The complete procedure, refered to as \algoname, is detailed in \Cref{algo:algo}. 

\emph{Initial distribution.}\, The initial sample $\iX{0}{0}$ is generated using an ODE solver applied to the PF-ODE \eqref{eq:pf-ode} using either the conditional denoisers $\denoiser{\std}{\cond}{}[\param]$ or the CFG denoisers \eqref{eq:cfg-denoiser} with a moderate guidance scale $1 \leq \scale_0 \ll \scale$. In the case of image generation, this tends to provide initial samples that exhibit high diversity but low perceptual quality \cite{ho2022classifier, karras2024guiding}. These coarse samples are subsequently refined and sharpened in the later stages of the algorithm, where a larger guidance scale $\scale$ is used. This behaviour is illustrated in \Cref{fig:gibbs_reps}. Given a total budget of $T$ function evaluations (NFE) and $R$ Gibbs iterations, we allocate a number $T_0$ of steps between $\lfloor T / 3 \rfloor$ to $\lfloor T / 2 \rfloor$ for generating the initial sample, with the remaining budget evenly distributed across the $R$ refinement stages.

\emph{The Gaussian case.}
We now analyze the behavior of \algoname\ in the Gaussian setting to gain a deeper understanding of its dynamics, with particular focus on the role of the parameter $\stds$. To enable an exact analysis, we consider the simplified Gaussian scenario in \Cref{ex:gaussian-cfg} with $\cond = 0$, which admits an explicit solution to the PF-ODE \eqref{eq:pf-ode} when using the CFG denoisers $\denoiser{\std}{\cond; \scale}{}[\cfg]$. This simplified setting mirrors the one studied in  \cite{bradley2024classifier}. Formally, we consider the following iterative procedure, which corresponds to \Cref{algo:algo} with no discretization error. Let $\iX{0}{0}$ be a square-integrable random variable, meaning we start from the conditional distribution, and define for all $r \geq 1$, $\iX{r}{0} = c(\stds) \big(\iX{r-1}{0} + \stds Z^{(r)})$,
where $c(\stds) \eqdef \gamma^\scale (1 + \std^2 _*)^{(\scale - 1) / 2} \big/ (\gamma^2 + (1 + \gamma^2) \std^2 _*)^{\scale / 2}$; see \Cref{subsec:gaussian_case} and \Cref{lem:ode-solution}. The following proposition compares the limiting distribution of $\iX{r}{0}$ (as $r \to \infty$) to the tilted distribution $\cpdata{0}{\cond; \scale}{} \eqdef \gauss(0, V(\scale))$, where $V(\scale) \eqdef \gamma^2 / (\gamma^2 + \scale)$. Define $V_\infty(\scale) \eqdef \vars c(\vars) / (1 - c(\vars))$. 

\begin{mdframed}[style=propFrame]
\begin{proposition} 
    \label{prop:variance-bias}
    For all $\scale > 1$, $(\iX{r}{0})_{r\in \nset}$ converges to $\gauss(0, V_\infty(\scale))$ exponentially fast in Wasserstein-2 distance, with rate proportional to $\vars$. Furthermore,
    \begin{equation*}
        V_\infty(\scale) = V(\scale) + O(\std^2 _*) \eqsp, \quad \text{as \, $\vars \to 0$}
        \eqsp.
    \end{equation*}
\end{proposition}
\end{mdframed}
Thus, in this simplified setting, the limiting distribution of our procedure converges to the target distribution as the parameter $\sigma_*$ tends to zero. However, as highlighted in \Cref{prop:variance-bias}, a bias--variance trade-off emerges when the number $R$ of iterations is finite. More specifically, selecting a very small value of $\stds$ may result in slow mixing, since each noising step induces only a minor change in the state $\iX{r}{\stds}$. On the other hand, selecting a larger $\stds$ can enhance mixing speed, albeit at the cost of introducing some bias. This is illustrated numerically in \Cref{apdx-sec:proofs}. 

\paragraph{A different guidance scheme.} We now present a generalization of the score expression \eqref{eq:tilted-score}, offering an alternative approximation of the denoisers \(\denoiser{\std}{\cond; \scale}{}\) in the small-noise regime, tailored for use within \algoname. 
\begin{mdframed}[style=propFrame]
\begin{proposition} 
    \label{prop:delayed_guidance}
    Let $\scale > 1$. For all $\delta > 0$ and $\std > 0$, the scores associated with $\cpdata{0}{\cond; \scale}{}$ are
    \begin{multline*}
        \nabla \log \cpdata{\std}{\cond; \scale}{\bx_\std} = (\scale - 1) \nabla \renyi{w}{\cpdata{0|\stdminus}{\bx_\std, \cond}{}}{\pdata{0|\stdplus}{\bx_\std}{}} \\
        + \scale \nabla \log \cpdata{\stdminus}{\cond}{\bx_\std} + (1 - \scale) \nabla \log \pdata{\stdplus}{}{\bx_\std} \eqsp,
    \end{multline*}
    where $\stdplus \eqdef \std \sqrt{(\scale - 1)/ \delta}$\, and \,$\stdminus \eqdef \std \sqrt{\scale / (1 + \delta)}$.  
\end{proposition}
\end{mdframed}
The proof is provided in \Cref{subsec-apdx:score-expressions}. The key novelty of this generalized formula is that it involves score evaluations at two distinct noise levels, $\stdminus$ and $\stdplus$. By setting $\delta = \scale - 1$, we recover the original formula \eqref{eq:tilted-score}, while selecting $0 < \delta < \scale - 1$ yields the ordering $\stdminus < \std < \stdplus$. Similar to \eqref{eq:tilted-score}, the Rényi divergence term vanishes as $\std \to 0$, leading to the practical approximation
$
\denoiser{\std}{\cond; \scale}{\bx_\std}
\approx \scale \denoiser{\stdminus}{\cond; \scale}{\bx_\std} + (1 - \scale) \denoiser{\stdplus}{\cond; \scale}{\bx_\std}
$ in this regime. 
The idea of performing guidance by combining denoisers at different noise levels has also been recently explored in \cite{sadat2025no, li2024self}. Specifically, these works introduce the modified denoiser
$
\scale \denoiser{\std}{\cond}{\bx_\std} + (1 - \scale) \denoiser{\tilde\std}{\cond}{\bx_\std},
$
where $\tilde\std \eqdef \std + \Delta \std$ for some small increment $\Delta \std$. Remarkably, this formulation avoids the use of the unconditional denoiser employed in CFG, thereby allowing guidance solely through the conditional denoiser. Moreover, \Cref{prop:delayed_guidance} can be applied to provide an alternative expression for the score $\nabla \log \cpdata{\std}{\cond}{}$, offering a theoretically grounded alternative to these recent guidance methods. Additional details and extensions of this approach are discussed in \Cref{apdx:delayed-guidance}. 
\section{Related works} 
\paragraph{Tilted distribution samplers.} Recent works on CFG \cite{bradley2024classifier, chidambaram2024what} have pointed out that $\denoiser{\std}{\cond; \scale}{}[\cfg]$ does not correspond to a valid denoiser for $\cpdata{0}{\cond; \scale}{}$. To address this shortcoming, \cite{bradley2024classifier} introduces a hybrid predictor--corrector approach, in which each update \eqref{eq:ddim} is followed by a few Langevin dynamics steps targeting the intermediate distribution $\cpdata{\std}{\cond; \scale}{}[\cfg]$; see \cite[Algorithm 2]{bradley2024classifier}. The key insight is that, although the family $(\cpdata{\std}{\cond; \scale}{}[\cfg])_{\std \in [0, \stdmax]}$ are not the marginals of a valid diffusion process,  
it nonetheless defines an annealing path connecting the initial distribution $\pdata{\stdmax}{}{}$ and $\cpdata{0}{\cond; \scale}{}$, and can therefore be used within an MCMC framework. The recent approach of \cite{skreta2025feynman} proposes using sequential Monte Carlo methods \cite{doucet2001sequential} to iteratively construct empirical approximations of the distributions $\cpdata{\std}{\cond; \scale}{}[\cfg]$ via a set of $N$ weighted particles. An simple derivation of this sampler is provided in \Cref{subsec:further-related-works}. Concurrently, \cite{lee2025debiasing} proposed a similar approach for discrete diffusion models. The present work falls within the same class of methods, sharing the goal of generating approximate samples from $\cpdata{0}{\cond; \scale}{}$. However, in contrast to particle-based approaches, we employ a Gibbs sampling-like 
approach and do not rely on using multiple particles to produce a single sample. 

\paragraph{Adaptive CFG methods.} Other approaches in the literature do not explicitly aim to sample from the tilted distribution. Instead, they employ various heuristics aimed at enhancing sample quality, diversity, or both simultaneously. To enhance sample diversity under guidance, \cite{chang2023muse, sadat2024cads} propose a time-dependent guidance scale that prioritizes the unconditional model in the early stages of the diffusion process, gradually transitioning toward standard CFG as sampling proceeds.  In \cite{kynkanniemi2024applying}, guidance is activated only when $\std$ is within a specified noise interval $[\std\textsubscript{lo}, \std\textsubscript{hi}]$,  and empirical results indicate that this strategy can significantly improve over the vanilla CFG. \cite{xianalysis} provides a systematic empirical study of adaptive CFG schedulers and arrives at the conclusion that increasing the guidance scale throughout the iterations improves the performance, aligning with the conclusions of the previous works. Finally, by reformulating text guidance as an inverse problem, \cite{chung2025cfg} arrives at the dynamic CFG schedule $\scale_t = \lambda \std_t / (\std_t - \std_{t-1})$, where $\lambda \in [0,1]$. 

\paragraph{Guidance through different mechanisms.} Alternatives to the CFG denoiser have also been proposed in the literature. 
 In \cite{karras2024guiding}, it is proposed using $\widetilde{D}^\param _\std(\bx_\std | \cond; \scale) = \scale \denoiser{\std}{\cond}{\bx_\std}[\param] + (1 - \scale) \denoiser{\std}{\cond}{\bx_\std}[\param_{-}]$, where $\denoiser{\std}{\cond}{}[\param_{-}]$ is a smaller or undertrained denoiser. 
 In terms of scores this is equivalent to substracting the score of a more spread-out density and results in a score of a more peaked density, 
 amplifying sharpness and conditioning alignment. The same effect is achieved by substracting the score at a higher noise level, as in \cite{sadat2025no,li2024self}, and with the score approximation resulting from \Cref{prop:delayed_guidance} when $\delta < \scale - 1$. 
 Moreover, \cite{pavasovic2025understanding} develops non-linear guidance mechanisms that, among other benefits, automatically reduce or switch off guidance when the difference between the conditional and unconditional scores becomes small. Our method also employs a distinct guidance strategy: it begins by generating an initial sample using no or moderate guidance, then repeatedly applies the forward noising process up to a predefined noise level, followed by denoising using CFG. This noising–denoising procedure, originally introduced in SDEdit \cite{meng2021sdedit} for image editing, enables the progressive refinement of an initial coarse sample while preserving its overall structure.
Therefore, it both maintains the diversity of the conditional model and produces high-quality samples.

%% file: experiments.tex
\begin{figure}[t]
    \vspace*{-3mm}
    \centering
    \begin{minipage}[t]{0.49\textwidth}
        \centering
        \resizebox{\textwidth}{!}{%
        \begin{tabular}[t]{lrrrrrr}
            \toprule
             & \multicolumn{6}{c}{\textbf{Quality metrics}} \\ \cmidrule(lr){2-7}
            Algorithm & FID $\downarrow$ & FD$_\text{DINOv2}$ $\downarrow$
                      & Precision $\uparrow$ & Recall $\uparrow$
                      & Density $\uparrow$ & Coverage $\uparrow$ \\
            \midrule
            \textbf{EDM2-S} \\ \cmidrule(lr){1-7}
            \CFG         & 2.30 & 88.70 & 0.61 & 0.57 & 0.58 & 0.54 \\
            \limitedCFG  & \textbf{1.71} & \underline{80.75} & \underline{0.61}
                         & \textbf{0.61} & \underline{0.58} & \underline{0.56} \\
            \CFGPP       & 2.89 & 95.52 & 0.60 & 0.54 & 0.57 & 0.52 \\
            \rowcolor{AlgoHighlight}
            \algoname    & \underline{1.78} & \textbf{75.38} & \textbf{0.64}
                         & \underline{0.59} & \textbf{0.63} & \textbf{0.58} \\
            \cmidrule(lr){1-7}
            \textbf{EDM2-XXL} \\ \cmidrule(lr){1-7}
            \CFG         & 1.81 & 56.82 & 0.67 & 0.65 & 0.71 & 0.65 \\
            \limitedCFG  & \underline{1.50} & \textbf{40.08} & \textbf{0.70}
                         & \textbf{0.68} & \textbf{0.78} & \textbf{0.70} \\
            \CFGPP       & 2.30 & 65.32 & 0.66 & 0.63 & 0.69 & 0.62 \\
            \rowcolor{AlgoHighlight}
            \algoname    & \textbf{1.48} & \underline{42.87} & \textbf{0.70}
                         & \textbf{0.68} & \underline{0.77} & \underline{0.69} \\
            \bottomrule
        \end{tabular}}
        \captionsetup{font=small}
        \captionof{table}{Comparison of average FID, FD$_\text{DINOv2}$,
                          Precision/Recall, and Density/Coverage on
                          \imagenet-$512$ for EDM2-S and EDM2-XXL.}
        \label{tab:imagenet512_edm2_full}
    \end{minipage}%
    \hfill
    \begin{minipage}[t]{0.49\textwidth}
        \centering
        \includegraphics[width=\textwidth, valign=t]{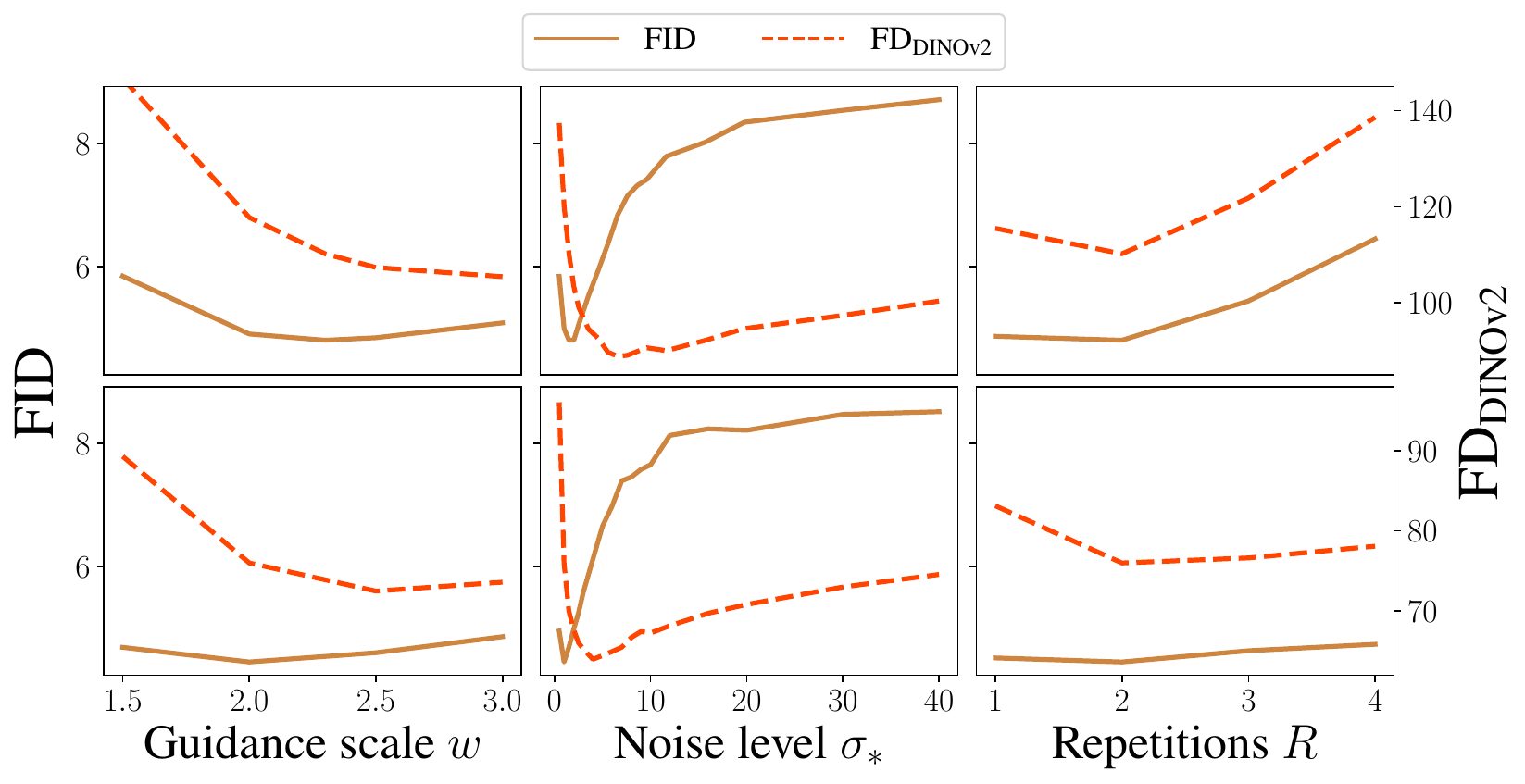}
        \vspace{-.20cm}
        \captionsetup{font=small}
        \caption{Impact of the hyperparameters for the EDM2-S (top) and EDM2-XXL (bottom) models. The metrics are computed with 10k generated samples.}
        \label{fig:imagenet_sensitivity}
    \end{minipage}
\end{figure}
    \section{Experiments}
We now present experimental results for both image and audio generation tasks, investigating two NFE setups and employing two different samplers: Heun for images and DDIM for audio generation.

\paragraph{Image experiments.}
We evaluate the algorithms on the \imagenet-$512$ dataset using EDM2 small (EDM2-S) and largest (EDM2-XXL) models which operate in the latent space \citep{karras2024edm2}. 
We follow the setting in \cite{kynkanniemi2024applying} and run all the algorithms with the 2\textsuperscript{nd} order Heun sampler and $32$ deterministic steps \cite{karras2022elucidating}. For the sake of completeness, this sampler is provided in \Cref{algo:heun}. 
We use FID \citep{heusel2017fid} and FD\textsubscript{DINOv2} \citep{stein2023fd_dinvo2} to assess the perceptual quality of the generated images. For both models, $\std_0=2 \times 10^{-3}$ and $\stdmax=80$. 
We also provide finer assessment of the algorithm in terms of fidelity and diversity by computing Precision/Recall and Density/Coverage \citep{naeem2020prdc}. We use the model parameters for which the EMA length has been tuned to yield the best FID under CFG, as reported in \cite{karras2024edm2}. 
As a result and unlike \cite{kynkanniemi2024applying}, we report both FID and FD\textsubscript{DINOv2} using the same set of model parameters. We compare \algoname\ against three competitors: \CFG\ \cite{ho2022classifier}, limited interval CFG (\limitedCFG) \citep{kynkanniemi2024applying}, and \CFGPP\ \citep{chung2025cfg}. For \CFG\ and \limitedCFG, we use the optimal parameters reported in \cite{karras2024edm2,kynkanniemi2024applying}, which optimize FID. For \algoname\ and \CFGPP, we performed a grid search over the parameters for each model, optimizing with respect to FID. The results are provided in \Cref{tab:imagenet512_edm2_full}. The metrics are calculated using 50k samples, and the reported values represent the average of three independent runs. We provide a qualitative assessment in \Cref{apdx:qualitative}. 

\begin{figure}[t]
    \vspace*{-4mm}
    \centering
    \begin{minipage}[t]{0.49\textwidth}
        \centering
        \resizebox{0.8\textwidth}{!}{
        \begin{tabular}[t]{lccc}
          \toprule
             & \multicolumn{3}{c}{\textbf{Quality metrics}} \\
          \cmidrule(lr){2-4}
            Algorithm & FAD$\downarrow$ & KL$\downarrow$ & IS$\uparrow$ \\
          \midrule
          \CFG                                & 1.78 & 1.59 & 7.07 \\
          \limitedCFG                         & 1.74 & 1.61 & 6.93 \\
          \CFGPP                              & 1.88 & \bf{1.55} & \underline{7.37} \\
          \rowcolor{AlgoHighlight} \algoname\, $(\scale_0=1.0, R=1)$     & \underline{1.71} & 1.65 & 7.05 \\
          \rowcolor{AlgoHighlight} \algoname\, $(\scale_0=1.5, R=1)$     & \bf{1.61} & 1.58 & 7.31 \\
          \rowcolor{AlgoHighlight} \algoname\, $(\scale_0=1.5, R=2)$     & 1.74 & \underline{1.56} & \bf{7.64} \\
          \bottomrule
        \end{tabular}
        }
        \captionsetup{font=small}
        \captionof{table}{Comparison of FAD, KL, and IS on \audiocaps\ test set for AudioLDM 2-Full-Large model.}
        \label{tab:mini_audioldm_full}
    \end{minipage}%
    \hfill
    \begin{minipage}[t]{0.49\textwidth}
        \centering
        \vspace{.3cm}
        \includegraphics[width=\textwidth, valign=t]{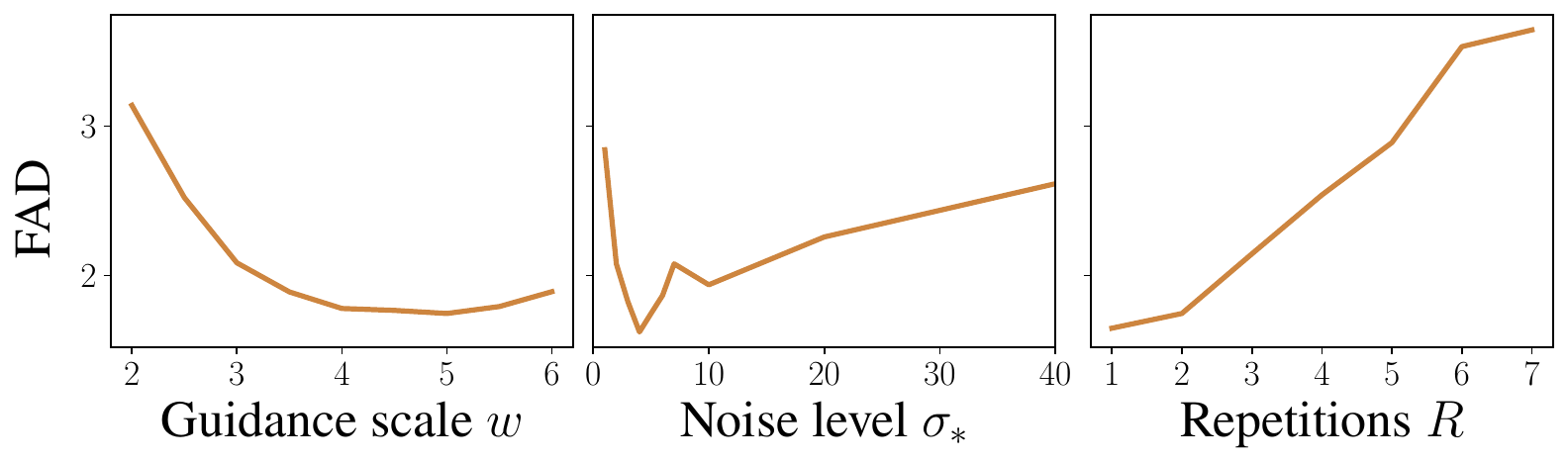}
        \vspace{.20cm}
        \captionsetup{font=small}
        \caption{Impact of the hyperparameters on the FAD with the AudioLDM 2-Full-Large model. The metrics are computed with 1k prompts.}
        \label{fig:audio_sensitivity}
    \end{minipage}
    \vspace*{-3mm}
\end{figure}

\emph{Results.\, } On both models, \algoname\ consistently achieves the best or near-best performance across all metrics. Both \algoname\ and \limitedCFG\ significantly outperform \CFG\ and \CFGPP\ on FD\textsubscript{DINOv2}, indicating improved perceptual quality as captured by DINOv2 features, as well as better fidelity and diversity (Precision/Recall and Density/Coverage). Overall, these results confirm that \algoname\ enhances visual quality while preserving high sample diversity. On the S model, we use an initial guidance scale $\scale_0 = 1$, \emph{i.e.}, we sample from the conditional model, and $\scale = 2.3$, $\stds = 2$ with $R=2$ repetitions and $T_0 = 12$ initial steps. For the XXL model, we also use $\scale_0 = 1$ and then $\scale=2$, $\stds=1$ and $R=2$. As a comparison, \limitedCFG\, uses $\std\textsubscript{lo}=0.28$, $\std\textsubscript{hi}=2.9$ and $\scale = 2.1$ for the S model, and $\std\textsubscript{lo}=0.19$, $\std\textsubscript{hi}=1.61$ and $\scale=2$ for the XXL. 
\begin{wrapfigure}{r}{0.3\textwidth}   
    \centering
    \includegraphics[width=0.29\textwidth]{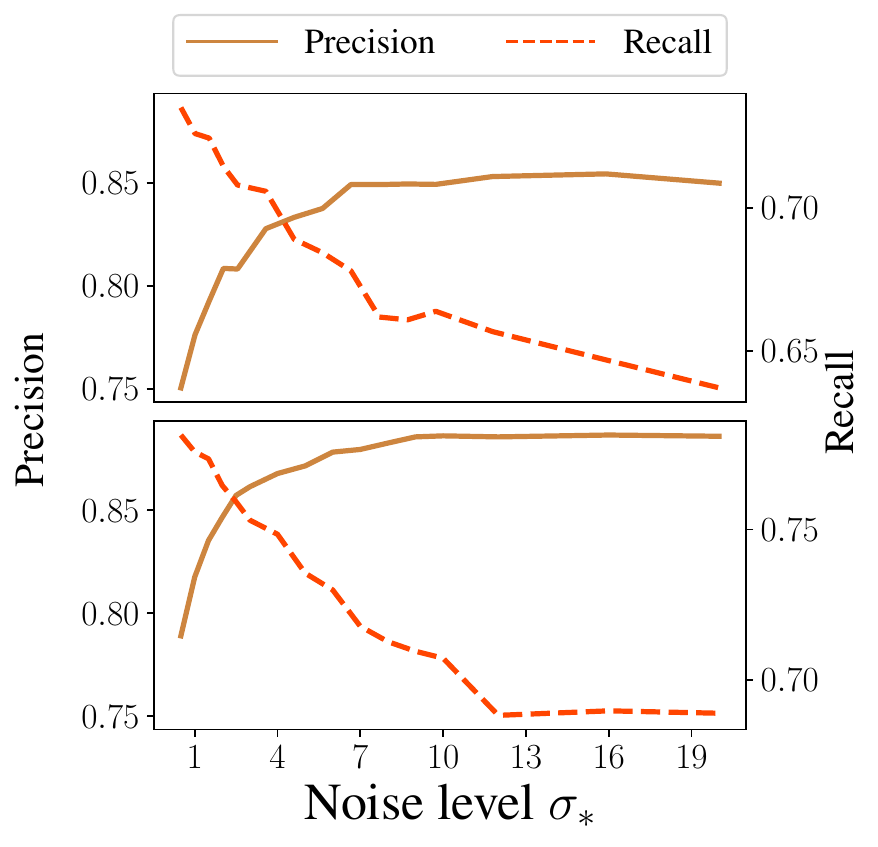}
    \captionsetup{font=small}
    \caption{Precision and recall as a function of $\stds$ for EDM2-S (top) and EDM2-XXL (bottom).}
    \label{fig:prdc}
    \vspace*{-3mm}
\end{wrapfigure}

\emph{Ablations.\,}  In Figure \ref{fig:imagenet_sensitivity} we investigate the impact of the hyperparameters $(\scale,\stds,R)$. For each model, we vary each parameter while keeping the others fixed at the values used to generate the results in \Cref{tab:imagenet512_edm2_full}.
Using $R=2$ repetitions improves upon $R=1$. With 50k samples and $R=1$ (and the other parameters optimized), the S and XXL models achieved FID/FD\textsubscript{DINOv2} of 1.84/81.67 and 1.57/49.74, respectively.  
These results are slightly worse than those obtained with $R=2$ as reported in Table \ref{tab:imagenet512_edm2_full}.
However, increasing $R$ beyond 2 worsens both FID and FD\textsubscript{DINOv2} scores, since more repetitions involve fewer ODE integration steps per repetition (to maintain the same NFE), yielding the loss of high-frequency details and degradation of semantic content. 
Increasing $\stds$ does not yield the same behavior however. FID drops to a minimum and then deteriorates, whereas FD\textsubscript{DINOv2} continues to improve beyond that turning point before eventually slightly degrading at higher noise levels. 
Figure \ref{fig:prdc} shows how Precision and Recall vary with the noise level $\sigma_\ast$. As anticipated, higher $\sigma_\ast$ diminishes diversity—Recall falls—because the subsequent samples $\iX{r}{0}$ ($r\geq1$) diverge from the initial, unguided sample $\iX{0}{0}$ and \algoname\ essentially behaves like standard CFG. On the other hand, Precision increases similarly to FD\textsubscript{DINOv2}. 
\paragraph{Audio experiments.}
We now assess \algoname\ on a text-to-audio task using the \audiocaps\ \citep{kim2019audiocaps} test set. We use the model AudioLDM 2-Full-Large of \citet{liu2024audioldm2} and adopt their experimental setup.
1k prompts were randomly selected from \audiocaps, and the following quality metrics were computed for each algorithm: Fréchet Audio Distance (FAD), Kullback--Leibler divergence over softmax outputs (KL), and Inception Score (IS). The KL is computed by applying Softmax to extracted features from the generated and groundtruth samples.
Following \cite{liu2024audioldm2}, for each prompt, the best out of 3 samples were selected.
In this setting, $\std_0=3.88 \times 10^{-2}$ and $\stdmax=83.33$ and all competitors are used with 200 steps and the DDIM sampler.
We run a grid search over the hyperparameter space of each algorithm and choose the parameters that optimize the FAD.
We provide more details about the models, the hyperparameters used for each algorithm, as well as the evaluation in \Cref{apdx:exp-details-audio}.
The results are reported in \Cref{tab:mini_audioldm_full}.

\emph{Results.\, } The configuration of \algoname\ with $\scale_0 = 1.5$ and $R=2$ achieves the best overall balance across the three evaluated metrics. Interestingly, reducing repetitions to $R=1$ significantly improves the FAD metric. While this scenario closely resembles \limitedCFG, \algoname\ consistently outperforms it, highlighting fundamental differences between these methods. In all these settings, we set $T_0 = 100$, $\stds=5$, and $\scale=5$. For \limitedCFG, we use $\std_\text{hi}=8.5$, $\scale=5$, and set $\std_\text{lo}=\std_0$, as further optimization of this parameter showed no measurable improvements beyond runtime efficiency. For \CFG, we choose $\scale=4.5$, and for \CFGPP, we set $\lambda = 0.1$. We also observed that omitting initial guidance ($\scale_0=1$) yields better FAD scores compared to \CFG, \limitedCFG, and \CFGPP, but negatively impacts the other metrics. Overall, both text-to-audio and text-to-image models require a moderate degree of guidance, since when unguided they frequently produce samples of poor perceptual quality. Finally, in \Cref{fig:audio_sensitivity} we provide an ablation study of the parameters, where we observe trends similar to those reported in \Cref{fig:imagenet_sensitivity} for the FID metric.

\begin{figure}[t]
    \vspace*{-8mm}
    \centering
    \includegraphics[width=0.8\textwidth]{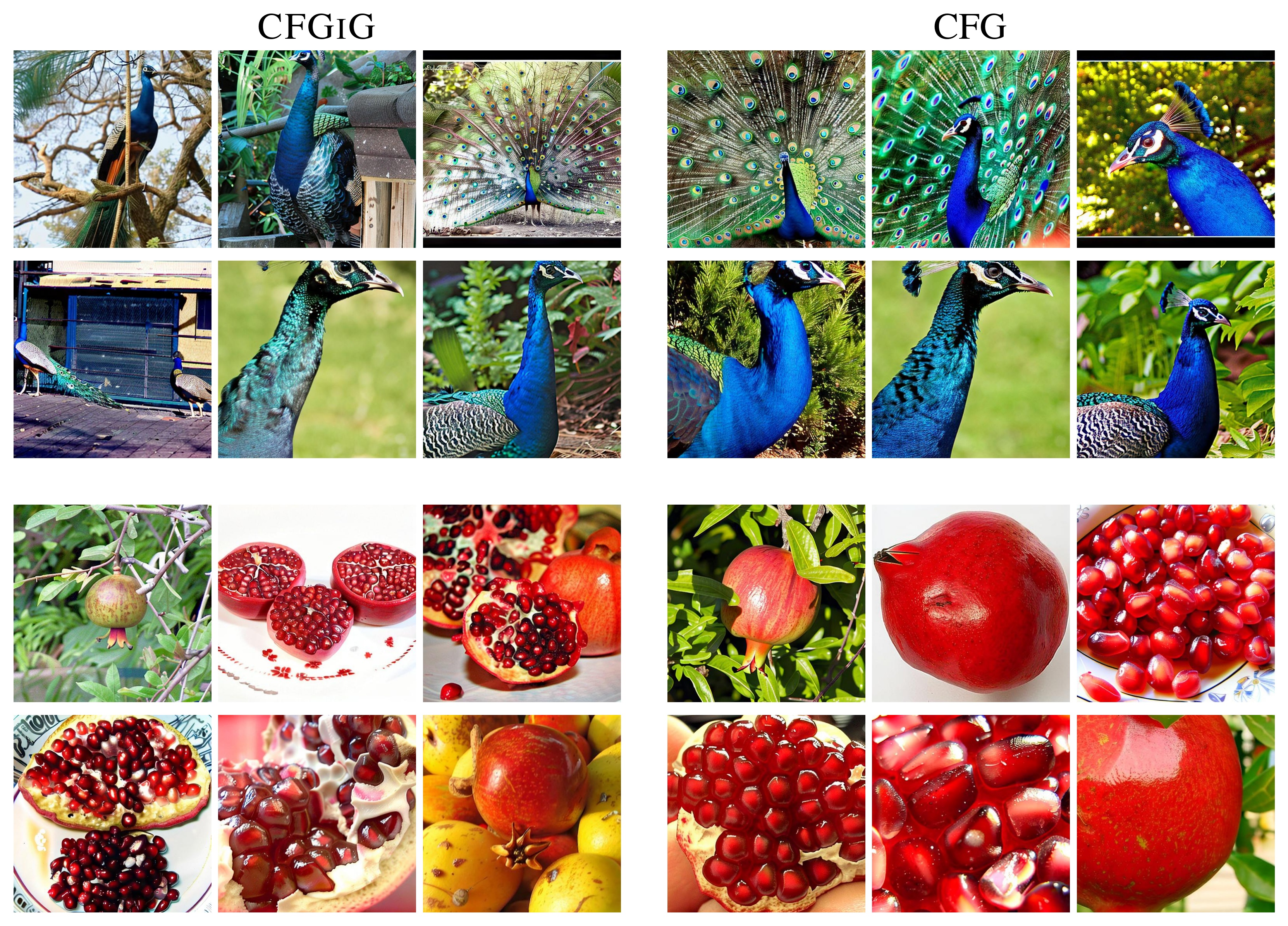}
    \captionsetup{font=small}
    \caption{Comparison of sample diversity between \algoname\ and \CFG\ across a batch of $6$ samples 
    generated using EDM-XXL model for two \imagenet\ classes: $84$ (top) $957$ (bottom).
    Both algorithms are run with a total of $32$ Heun steps and the same seed.
    \CFG\ is run with a guidance scale $\scale=3.5$.
    \algoname\ is run with $12$ initial steps at guidance scale $\scale_0=1$,
    followed by two Gibbs iterations starting at $\stds=1$ and using $\scale=3.5$.
    }
    \vspace*{-3mm}
\end{figure}


%% file: appendix/proofs.tex
\subsection{Score expressions}
\label{subsec-apdx:score-expressions}
In this section, we adopt measure-theoretic notations. Denote by $\mathcal{P}(\rset^d)$ the set of probability distribution on $\rset^d$ endowed with its Borel $\sigma$-algebra. We first restate the definition of the $\scale$-Rényi divergence between two arbitrary probability distributions $\mu$ and $\nu$ in $\mathcal{P}(\rset^\dimx)$:
\begin{equation} 
      \renyi{\scale}{\mu}{\nu} \eqdef \begin{cases} \frac{1}{\scale - 1} \log \int \big[ \frac{\rmd \mu}{\rmd \nu} \big]^\scale \rmd \nu \quad & \text{if} \, \mu \ll \nu, \\
      \infty \quad & \text{otherwise}.\end{cases}
\end{equation}

Let $\pdata{0}{}{} \in \mathcal{P}(\rset^\dimx)$ and denote, for any $\cond \in \mathcal{C}$, $\constNc \eqdef \int \clf{0}{\tilde\bx_0}{\cond} \pdata{0}{}{\rmd \tilde\bx_0}$. We impose the following assumptions on the likelihood function $\bx_0 \mapsto \clf{0}{\bx_0}{\cond}$ 
in order to make $\cpdata{0}{\cond}{\rmd \bx_0} = \clf{0}{\bx_0}{\cond} \pdata{0}{}{\rmd \bx_0} \big/ \constNc$ a well-defined probability distribution. 
\begin{hypA} 
      \label{assp:post-def}
      For every $\bx \in \rset^\dimx$ and $\cond \in \mathcal{C}$, $\clf{0}{\bx_0}{\cond} \geq 0$ and  $0 < \constNc < \infty$. 
\end{hypA}
Moreover, we recall the definitions
\begin{align} 
      \label{eq:pdata-cond}
      \pdata{0|\std}{\bx_\std}{\rmd \bx_0} & \eqdef \frac{\pdata{0}{}{\rmd \bx_0} \fw{\std|0}{\bx_0}{\bx_\std}}{\pdata{\std}{}{\bx_\std}} \eqsp,  \\
      \cpdata{0|\std}{\bx_\std, \cond}{\rmd \bx_0} & \eqdef \frac{\cpdata{0}{\cond}{\rmd \bx_0} \fw{\std|0}{\bx_0}{\bx_\std}}{\cpdata{\std}{\cond}{\bx_\std}} \label{eq:cpdata-cond} \eqsp.
\end{align}
\begin{mdframed}[style=propFrame]
      \begin{proprestate}[Restatement of \Cref{prop:tilted-score}]
          Assume 
         \assp{\ref{assp:post-def}}. Then for every $\std > 0$, $\bx_\std \in \rset^\dimx$, and $\cond \in \mathcal{C}$, 
          \begin{equation*} 
              \nabla_{\bx_{\std}} \log \cpdata{\std}{\cond; \scale}{\bx_\std} = (\scale - 1) \nabla_{\bx_{\std}} \renyi{\scale}{\cpdata{0|\std}{\bx_\std, \cond}{}}{\pdata{0|\std}{\bx_\std}{}} + \nabla_{\bx_{\std}} \log \cpdata{\std}{\cond; \scale}{\bx_\std}[\cfg] \eqsp. 
          \end{equation*} 
      \end{proprestate}
      \end{mdframed}
\begin{proof}[Proof of \Cref{prop:tilted-score}]
    \label{proof:tilted-score}
Combining definitions \eqref{eq:pdata-cond} and \eqref{eq:cpdata-cond} yields  
$$ 
      \cpdata{0|\std}{\bx_\std, \cond}{\rmd \bx_0} = \frac{1}{\constNc} \frac{\clf{0}{\bx_0}{\cond} \pdata{\std}{}{\bx_\std}}{\cpdata{\std}{\cond}{\bx_\std}} \pdata{0|\std}{\bx_\std}{\rmd \bx_0} \eqsp,
$$ 
from which it follows that $\cpdata{0|\std}{\bx_\std, \cond}{} \ll \pdata{0|\std}{}{}$ and $\frac{\rmd \cpdata{0|\std}{\bx_\std, \cond}{}}{\rmd \pdata{0|\std}{\bx_\std}{}} (\bx_0) = \frac{\clf{0}{\bx_0}{\cond} \pdata{\std}{}{\bx_\std}}{\constNc \cpdata{\std}{\cond}{\bx_\std}}$. Hence, 
\begin{align*} 
      & \int \bigg[ \frac{\rmd \cpdata{0|\std}{\bx_\std, \cond}{}}{\rmd \pdata{0|\std}{\bx_\std}{}}\bigg]^\scale (\bx_0) \, \pdata{0|\std}{\bx_\std}{\rmd \bx_0} \\
      & \hspace{1.5cm} = \frac{1}{\constNc^{\scale}} \frac{\pdata{\std}{}{\bx_\std}^\scale}{\cpdata{\std}{\cond}{\bx_\std}^\scale} 
      \int \clf{0}{\cond}{\bx_0}^\scale \, \pdata{0|\std}{\bx_\std}{\rmd \bx_0} 
      \\
      & \hspace{1.5cm} = \frac{1}{\constNc^{\scale}} \frac{\pdata{\std}{}{\bx_\std}^{\scale-1}}{\cpdata{\std}{\cond}{\bx_\std}^\scale} 
      \int \clf{0}{\cond}{\bx_0}^\scale \fw{\std|0}{\bx_0}{\bx_\std} \, \pdata{0}{}{\rmd \bx_0} 
      \\
      &  \hspace{1.5cm} =  \frac{\int \clf{0}{\tilde\bx_0}{\cond}^\scale \, \pdata{0}{}{\rmd \tilde\bx_0}}{\constNc^{\scale}} \frac{\pdata{\std}{}{\bx_\std}^{\scale-1}}{\cpdata{\std}{\cond}{\bx_\std}^\scale} \cpdata{\std}{\cond; \scale}{\bx_\std} 
      \eqsp,
\end{align*}  
where we used, in the last step, the definition 
$$
\cpdata{\std}{\cond; \scale}{\bx_\std} \eqdef \int \fw{\std|0}{\bx_0}{\bx_\std} \, \frac{\clf{0}{\bx_0}{\cond}^\scale \, \pdata{0}{}{\rmd \bx_0}}{\int \clf{0}{\tilde\bx_0}{\cond}^\scale \, \pdata{0}{}{\rmd \tilde\bx_0}} \eqsp.
$$  
Taking the logarithm of both sides of the identity and then differentiating with respect to $\bx_\std$ yields 
\begin{multline*}
      (\scale - 1) \nabla_{\bx_{\std}} \renyi{\scale}{\cpdata{0|\std}{\bx_\std, \cond}{}}{\pdata{0|\std}{\bx_\std}{}} \\
      = \nabla_{\bx_{\std}} \log \cpdata{\std}{\cond; \scale}{\bx_\std} - \scale \nabla_{\bx_{\std}} \log \cpdata{\std}{\cond}{\bx_\std} + (\scale - 1) \nabla_{\bx_{\std}} \log \pdata{\std}{}{\bx_\std}
      \eqsp,
\end{multline*}
which is the desired result upon rearranging.
\end{proof}

Similarly, we establish the following generalization of the identity in \Cref{prop:tilted-score}, involving the scores at two different noise levels.
\begin{mdframed}[style=propFrame]
\begin{proprestate}[Restatement of \Cref{prop:delayed_guidance}]
      Assume \assp{\ref{assp:post-def}} and let $\scale > 1$.
      Then for every $\delta > 0$, $\std > 0$, $\bx_\std \in \rset^\dimx$, and $\cond \in \mathcal{C}$, 
      \begin{multline*}
            \nabla_{\bx_{\std}} \log \cpdata{\std}{\cond; \scale}{\bx_\std} = (\scale - 1) \nabla_{\bx_{\std}} \renyi{w}{\cpdata{0|\stdminus}{\bx_\std, \cond}{}}{\pdata{0|\stdplus}{\bx_\std}{}} \\
            + \scale \nabla_{\bx_{\std}} \log \cpdata{\stdminus}{\cond}{\bx_\std} + (1 - \scale) \nabla_{\bx_{\std}} \log \pdata{\stdplus}{}{\bx_\std} \eqsp,
      \end{multline*}
      where $\stdplus \eqdef \std \sqrt{(\scale - 1)/ \delta}$\, and \,$\stdminus \eqdef \std \sqrt{\scale / (1 + \delta)}$.  
\end{proprestate}
\end{mdframed}
\begin{proof}[Proof of \Cref{prop:delayed_guidance}]
Since $\scale > 1$ and $\delta > 0$, the two noise levels $\stdplus$ and $\stdminus$ are well defined and by combining definitions \eqref{eq:pdata-cond} and \eqref{eq:cpdata-cond} we obtain that 
$$
      \cpdata{0|\stdminus}{\bx_\std, \cond}{\rmd \bx_0} = \frac{1}{\constNc}  \frac{\clf{0}{\bx_0}{\cond} \pdata{\stdplus}{}{\bx_\std} \fw{\stdminus|0}{\bx_0}{\bx_\std}}{\cpdata{\stdminus}{\cond}{\bx_\std} \fw{\stdplus|0}{\bx_0}{\bx_\std}} \, \pdata{0|\stdplus}{\bx_\std}{\rmd \bx_0} \eqsp.
$$
Thus,
\begin{align*} 
      \lefteqn{\int \bigg[ \frac{\rmd \cpdata{0|\stdminus}{\bx_\std, \cond}{}}{\rmd \pdata{0|\stdplus}{\bx_\std}{}}\bigg]^\scale (\bx_0) \, \pdata{0|\stdplus}{\bx_\std}{\rmd \bx_0}} \hspace{10mm} \\
      & = \frac{1}{\constNc^{\scale}}  \frac{\pdata{\stdplus}{}{\bx_\std}^\scale}{\cpdata{\stdminus}{\cond}{\bx_\std}^\scale} \int \bigg[
            \frac{\fw{\stdminus|0}{\bx_0}{\bx_\std}}{\fw{\stdplus|0}{\bx_0}{\bx_\std}}
      \bigg]^\scale \clf{0}{\bx_0}{\cond}^\scale  \pdata{0|\stdplus}{\bx_\std}{\rmd \bx_0} \\
      & = \frac{1}{\constNc^{\scale}}  \frac{\pdata{\stdplus}{}{\bx_\std}^{\scale-1}}{\cpdata{\stdminus}{\cond}{\bx_\std}^\scale} \bigg[
            \int 
        \frac{\fw{\stdminus|0}{\bx_0}{\bx_\std}^\scale}{\fw{\stdplus|0}{\bx_0}{\bx_\std}^{\scale-1}}
            \clf{0}{\bx_0}{\cond}^\scale  \pdata{0}{}{\rmd \bx_0} 
      \bigg] \eqsp.
\end{align*} 
Here, by the definition of $\stdminus$, 
\begin{align*}
      \fw{\stdminus|0}{\bx_0}{\bx_\std}^\scale 
            & = \normpdf\big(\bx_\std; \bx_0, \std^2 \scale / (\delta+1) \Id\big)^\scale \\
            & \propto \normpdf\big(\bx_\std; \bx_0, \std^2 \Id \big)^{(\delta+1)} \\
            & \propto \fw{\std|0}{\bx_0}{\bx_\std}^{\delta+1} \eqsp, 
\end{align*}
where the constant of proportionality is independent of $(\bx_0, \bx_\std)$. Similarly, 
$
\fw{\stdplus|0}{\bx_0}{\bx_\std}^{\scale-1} \propto \fw{\std|0}{\bx_0}{\bx_\std}^{\delta}
$. Hence, $\fw{\stdminus|0}{\bx_0}{\bx_\std}^{\scale} / \fw{\stdplus|0}{\bx_0}{\bx_\std}^{\scale-1} \propto \fw{\std|0}{\bx_0}{\bx_\std}$, which implies that  
\begin{align*}
      & \int \bigg[ \frac{\rmd \cpdata{0|\stdminus}{\bx_\std, \cond}{}}{\rmd \pdata{0|\stdplus}{\bx_\std}{}}\bigg]^\scale (\bx_0) \, \pdata{0|\stdplus}{\bx_\std}{\rmd \bx_0} \\
      & \hspace{1.5cm} \propto \frac{1}{\constNc^{\scale}}  \frac{\pdata{\stdplus}{}{\bx_\std}^{\scale-1}}{\cpdata{\stdminus}{\cond}{\bx_\std}^\scale} \int \fw{\std|0}{\bx_0}{\bx_\std} \clf{0}{\bx_0}{\cond}^\scale \pdata{0}{}{\rmd \bx_0} \\
      & \hspace{1.5cm} \propto \frac{\int \clf{0}{\tilde\bx_0}{\cond}^\scale \pdata{0}{}{\rmd \tilde\bx_0}}{\constNc^{\scale}}  \frac{\pdata{\stdplus}{}{\bx_\std}^{\scale-1}}{\cpdata{\stdminus}{\cond}{\bx_\std}^\scale} \cpdata{\std}{\cond; \scale}{\bx_\std}  \eqsp.
\end{align*}
Finally, taking the logarithm of both sides and then the gradient with respect to $\bx_\std$ yields 
\begin{multline*}
      (\scale - 1) \nabla_{\bx_{\std}} \renyi{\scale}{\cpdata{0|\stdminus}{\bx_\std, \cond}{}}{\pdata{0|\stdplus}{\bx_\std}{}} \\
      = \nabla_{\bx_{\std}} \log \cpdata{\std}{\cond; \scale}{\bx_\std} - \scale \nabla_{\bx_{\std}} \log \cpdata{\stdminus}{\cond}{\bx_\std} + (\scale - 1) \nabla_{\bx_{\std}} \log \pdata{\stdplus}{}{\bx_\std} \eqsp, 
\end{multline*}
which is the desired expression upon rearranging.
\end{proof}

A direct consequence of \Cref{prop:delayed_guidance} is that the score of the unconditional distribution can be expressed in terms of the scores at two different noise levels.
\begin{mdframed}[style=propFrame]
\begin{corollary}
      \label{cor:delayed_score}
Let $\pdata{0}{}{} \in \mathcal{P}(\rset^\dimx)$ and $\scale > 1$.
Then for every $\delta > 0$, $\std > 0$, and $\bx_\std \in \rset^\dimx$, 
\begin{multline*}
      \nabla \log \pdata{\std}{}{\bx_\std} = (\scale - 1) \nabla \renyi{w}{\pdata{0|\stdminus}{\bx_\std}{}}{\pdata{0|\stdplus}{\bx_\std}{}} \\
      + \scale \nabla \log \pdata{\stdminus}{}{\bx_\std} + (1 - \scale) \nabla \log \pdata{\stdplus}{}{\bx_\std} \eqsp,
\end{multline*}
where $\stdplus \eqdef \std \sqrt{(\scale - 1)/ \delta}$\, and \,$\stdminus \eqdef \std \sqrt{\scale / (1 + \delta)}$.
\end{corollary}
\end{mdframed}
\begin{proof}
The result follows by applying \Cref{prop:delayed_guidance} with $g_0$ being the constant mapping $\bx \mapsto 1$.
\end{proof}
\subsection{Score approximations}

Throughout this section, we will often assume that the relevant densities belong to the following class. 

\begin{definition}
    For $m \in \nset^*$, denote by $\pclass{m}$ the set of functions $f_0: \rset^m \to \rset$ satisfying 
    \begin{itemize}
        \item[(i)] $f_0$ is strictly positive,  \label{assp:positive}
      \item[(ii)] $f_0$ is  three times continuously differentiable with third-order partial derivatives of at most polynomial growth,  \emph{i.e.}, there exist $C \geq 0$ and $m \geq 0$ such that for any $(i, j, k) \in \intset{1}{\dimx}^3$, $| \partial_{i} \partial_{j} \partial_k f_0(\bx) | \leq C (1 + \| \bx \|_2)^m$. \label{assp:growth}
    \end{itemize} \label{def:pclass}
\end{definition}
The proof of \Cref{prop:tilted-score} relies on the following general result, which is proven subsequently. For any $\std > 0$ and $\bx \in \rset^\dimx$, define 
$$
f_\std(\bx) \eqdef \int f_0(\bx_0) \fw{\std|0}{\bx_0}{\bx} \, \rmd \bx_0 \eqsp.
$$
\begin{mdframed}[style=propFrame]
\begin{lemma}
      \label{lem:convolution-expansion}
Let $f_0 \in \pclass{\dimx}$.  
Then for every $\bx \in \rset^\dimx$, 
$$ \nabla \log f_\sigma(\bx) = \nabla \log f_0(\bx) + O(\std^2) \quad \text{as \, $\std \to 0$} \eqsp.$$
\end{lemma}
\end{mdframed}
\begin{mdframed}[style=propFrame]
      \begin{proprestate}[Restatement of \Cref{prop:renyi-expansion}]
      Assume that $\pdata{0}{}{} $, $\clf{0}{\cdot}{\cond}$ and $\clf{0}{\cdot}{\cond}^{w}$ belong to $\pclass{\dimx}$ for some $\cond \in \mathcal{C}$ and $w >1$. Then for any fixed $\bx \in \rset^\dimx$, 
      $$
      \nabla \renyi{\scale}{\cpdata{0|\std}{\bx}{}}{\pdata{0|\std}{\bx}{}} = O(\std^2) \quad \text{as \, $\std \to 0$} \eqsp.
      $$ 
  \end{proprestate}
  \end{mdframed}

  \begin{proof} 
      Since the third derivatives of $\pdata{0}{}{}$ and $\clf{0}{\cdot}{\cond}$ satisfy the polynomial growth assumption (ii) in \Cref{def:pclass}, so does $\cpdata{0}{\cond}{}$. Thus, by \Cref{lem:convolution-expansion} it holds that  
      \begin{equation*}
            \scale \nabla \log \cpdata{\std}{\cond}{\bx} + (1 - \scale) \nabla \log \pdata{\std}{}{\bx} \\
            = \scale \nabla \log \cpdata{0}{\cond}{\bx} + (1 - \scale) \nabla \log \pdata{0}{}{\bx} + O(\std^2)
      \end{equation*}
      as $\std \to 0$.
      Recall that we assume that  $ \clf{0}{\bx}{\cond}^\scale$ and $\pdata{0}{}{}$ belong to $\pclass{\dimx}$.  Consequently, $\cpdata{0}{\cond; \scale}{}$ satisfies the same condition. Thus, since $\cpdata{0}{\cond; \scale}{\bx} \propto \cpdata{0}{\cond}{\bx}^\scale \pdata{0}{}{\bx}^{1 - \scale}$, we have that
      $$ 
            \nabla \log \cpdata{\std}{\cond; \scale}{\bx} =  \scale \nabla \log \cpdata{0}{\cond}{\bx} + (1 - \scale) \nabla \log \pdata{0}{}{\bx} + O(\std^2) 
      $$ 
      as $\std \to 0$. 
      Finally, by \Cref{prop:tilted-score}, as $\std \to 0$, 
      \begin{align*} 
            (\scale - 1) \nabla \renyi{\scale}{\cpdata{0|\std}{\bx}{}}{\pdata{0|\std}{\bx}{}} & = \nabla \log \cpdata{\std}{\cond; \scale}{\bx} -         \nabla \log        \cpdata{\std}{\cond; \scale}{\bx_\std}[\cfg] \\
            & = O(\std^2) \eqsp.
      \end{align*}
  \end{proof}     

\begin{proof}[Proof of \Cref{lem:convolution-expansion}]
      For every $\bx \in \rset^\dimx$, 
\begin{align} 
      \nabla \log f_\std(\bx) 
      &= \frac{\int \nabla_\bx \fw{\std|0}{\bx_0}{\bx} f_0(\bx_0)\, \rmd \bx_0}{\int f_0(\tilde\bx_0) \fw{\std|0}{\tilde\bx_0}{\bx} \, \rmd \tilde\bx_0} \nonumber \\ &= \frac{\int \nabla_\bx \log \fw{\std|0}{\bx_0}{\bx}  f_0(\bx_0) \fw{\std|0}{\bx_0}{\bx} \, \rmd \bx_0}{\int f_0(\tilde\bx_0) \fw{\std|0}{\tilde\bx_0}{\bx} \, \rmd \tilde\bx_0} \eqsp. \label{eq:score_ratio_latex} 
\end{align}
Since $\fw{\std|0}{\bx_0}{\bx} = \normpdf(\bx; \bx_0, \std^2 \Id)$ and $\nabla \log \fw{\std|0}{\bx_0}{\bx} = (\bx_0 - \bx) / \std^2$, making the change of variables $\bz = \bx_0 - \bx$ yields the expression
\begin{equation} 
      \nabla \log f_\std(\bx) = \frac{1}{\std^2} \frac{\int \bz \, f_0(\bz + \bx) \varphi_\dimx(\bz; \std) \, \rmd \bz}{\int f_0(\tilde\bz + \bx) \varphi_\dimx(\tilde\bz; \std) \, \rmd \tilde\bz},  
\end{equation}
where $\varphi_\dimx(\cdot; \std)$ denotes the zero-mean $\dimx$-dimensional Gaussian density with covariance $\std^2 \Id$. Define $N(\bx) \eqdef  \int \bz \, f_0(\bz + \bx) \varphi_\dimx(\bz; \std) \rmd \bz$ and $D(\bx) \eqdef \int f_0(\bz + \bx) \varphi_\dimx(\bz; \std) \, \rmd \bz$, so that $\nabla \log f_\std(\bx) = \frac{1}{\std^2} N(\bx) / D(\bx)$. 

We start with the numerator $N(\bx)$ and expand $f_0$ around $\bx$ using the Taylor--Lagrange formula. More precisely, there exists $\xi(\bx, \bz) \in [0, 1]$ such that
$$ 
f_\std(\bz + \bx) = \std^2 f_0(\bx) + \bz^\intercal \nabla f_0(\bx) + \frac{1}{2} \bz^\intercal \nabla^2 f_0(\bx) \bz + \frac{1}{6} \sum_{i, j, k} \bz_i \bz_j \bz_k \partial_i \partial_j \partial_k f_0(\bx + \xi(\bx, \bz) \cdot \bz) \eqsp.
$$
Hence, since $\int \bz \, \varphi_\dimx(\bz; \std) \, \rmd \bz = 0$ and $\int \bz_i \bz_j \, \varphi_\dimx(\bz; \std) \, \rmd \bz = \delta_{ij}$, $\int \bz_i \bz_j \bz_k  \, \varphi(\bz; \std) \, \rmd \bz = 0$, 
\begin{align}
N(\bx) & = \int \bigg[ f_0(\bx) \bz + (\bz^\intercal \nabla f_0(\bx)) \bz + \bigg(\frac{1}{2} \bz^\intercal \nabla^2 f_0(\bx + \xi(\bx, \bz) \cdot \bz) \bz \bigg) \bz  \nonumber \\
& \hspace{.5cm} + \frac{1}{6} \sum_{i, j, k} \left( \bz_i \bz_j \bz_k \partial_i \partial_j \partial_k f_0(\bx + \xi(\bx, \bz) \cdot \bz) \right) \bz \bigg] \, \varphi(\bz; \std) \, \rmd \bz \nonumber \\
& = \std^2 \nabla f_0(\bx) + \int \bigg[ \frac{1}{6} \sum_{i, j, k} \left( \bz_i \bz_j \bz_k \partial_i \partial_j \partial_k f_0(\bx + \xi(\bx, \bz) \cdot \bz) \right) \bz \bigg] \, \varphi_\dimx(\bz; \std) \, \rmd \bz \eqsp. \label{eq:taylor-lagrange-exp}
\end{align} 
Next, for the second term in \eqref{eq:taylor-lagrange-exp}, it holds that 
\begin{multline*} 
      \int \bigg[ \sum_{i, j, k} \left( \bz_i \bz_j \bz_k \partial_i \partial_j \partial_k f_0(\bx + \xi(\bx, \bz) \cdot \bz) \right) \bz \bigg] \, \varphi_\dimx(\bz; \std) \, \rmd \bz \\
       = \std^4 \int \bigg[ \sum_{i, j, k} \left( \bz_i \bz_j \bz_k \partial_i \partial_j \partial_k f_0(\bx + \std \xi(\bx, \bz) \cdot \bz) \right) \bz \bigg] \, \varphi_\dimx(\bz; 1) \, \rmd \bz\eqsp.
\end{multline*}
By continuity of the third derivatives of $f_0$,
$
\lim_{\std \downarrow 0} \partial_i \partial_j \partial_k f_0(\bx + \std \xi(\bx, \bz) \cdot \bz) = \partial_i \partial_j \partial_k f_0(\bx)
$ for any $\bz \in \rset^\dimx$ and $(i, j, k) \in \intset{1}{d}^3$. Moreover, since the second derivatives are assumed to grow at most polynomially 
and we integrate against $\varphi_\dimx(\cdot; 1)$, having finite moments of all orders,  we can combine the dominated convergence theorem with the fact that $\int \bz_i \bz_j \bz_k \bz_\ell \, \varphi_\dimx(\bz; \std) \, \rmd \bz = \std^4 (\delta_{ij} \delta_{kl} + \delta_{ik}\delta_{jl} + \delta_{il} \delta_{jk})$ to obtain 
\begin{equation*} 
      \lim_{\std \downarrow 0} \frac{1}{\std^4} \int \bigg[ \sum_{i, j, k} \left( \bz_i \bz_j \bz_k \partial_i \partial_j \partial_k f_0(\bx + \std \xi(\bx, \bz) \cdot \bz) \right) \bz \bigg] \, \varphi_\dimx(\bz; \std) \, \rmd \bz = 3 \sum_{i, j} \partial^2 _i \partial_j f_0(\bx) \eqsp. 
\end{equation*}
As a result, $N(\bx) = \std^2 \nabla f_0(\bx) + O(\std^4).$ 

We now deal with the denominator $D(\bx)$ in a similar fashion. Again, by the Taylor--Lagrange formula, there exists $\xi(\bx; \bz) \in [0, 1]$ such that
\begin{align*}
      D(\bx) & = \int \big[ f_0(\bx) + \bz^\intercal \nabla f_0(\bx) + \frac{1}{2} \bz^\intercal \nabla^2 f_0(\bx + \xi(\bx, \bz) \cdot \bz) \bz\big] \, \varphi_\dimx(\bx; \std) \, \rmd \bz \\
      & = f_0(\bx) + \std^2 \int \bz^\intercal \nabla^2 f_0(\bx + \std \xi(\bx, \bz) \cdot \bz) \bz \, \varphi(\bz; 1) \, \rmd \bz \eqsp.
\end{align*}
Using again the polynomial-growth assumption (\Cref{def:pclass}(ii)) 
yields $D(\bx) = f_0(\bx) + O(\std^2)$. Combining the expansions for the numerator and denominator, we obtain, as $f_0$ is strictly positive (\Cref{def:pclass}(i)), 
\begin{align*} 
      \nabla \log f_\std(\bx) & = \frac{1}{\std^2} \frac{N(\bx)}{D(\bx)} \\ 
      &= \frac{\nabla f_0(\bx) + O(\std^2)}{f_0(\bx) + O(\std^2)} \\
      & = \frac{1}{f_0(\bx)} ( \nabla f_0(\bx) + O(\std^2)) (1 + O(\std^2)) \\ 
      & = \nabla \log f_0(\bx) + O(\std^2) \eqsp. 
\end{align*}
\end{proof}
\subsection{Classifier-free Gibbs-like guidance}
\paragraph{Uniqueness of the stationary distribution in the ideal case.} Recall that the Markov chain under consideration is given by $\iX{r+1}{0} = F_{0|\stds}\big(\iX{r}{0} + \stds Z^{(r+1)}| \cond; \scale\big)$, where $(Z^{(r)})_{r \in \nset}$ is a  sequence of i.i.d. standard Gaussian random variables and $F_{0|\stds}(\bx_\stds|\cond; \scale)$ is the solution to the PF-ODE \eqref{eq:pf-ode} with  denoiser $\denoiser{\std}{\cond; \scale}{}$ and initial condition $\bx_\stds$. By definition of this denoiser, it is associated with $\cpdata{0}{\cond; \scale}{}$, \ie, if $X_{\stds} \sim \cpdata{\stds}{\cond; \scale}{}$, then $F_{0|\stds}(X_\stds| \cond, \scale) \sim \cpdata{0}{\cond; \scale}{}$; see \Cref{sec:gibbs-like-guidance}.

Moreover, by \ref{ref:step_1_gibbs_like}, if $\iX{r}{0} \sim \cpdata{0}{\cond; \scale}{}$, then $\iX{r}{\stds} \sim \cpdata{\stds}{\cond; \scale}{}$. Hence $\iX{r+1}{0} = F_{0|\stds}(\iX{r}{\stds}| \cond; \scale) \sim \cpdata{0}{\cond; \scale}{}$, showing that $\cpdata{0}{\cond; \scale}{}$ is indeed a stationary distribution. In order to show that this invariant distribution is unique, we now establish that the chain $(\iX{r}{0})_{r \in \nset}$ is $\Leb$-irreducible, where $\Leb$ is the Lebesgue measure. To facilitate the analysis, we make the following simplifying assumption.
\begin{hypA} 
      \label{assp:diffeo}
      For some fixed $\stds > 0$, $\scale > 1$, and $\cond \in \mathcal{C}$, the flow map $\bz \mapsto F_{0|\stds}(\bz | \cond; \scale)$ is a continuously differentiable diffeomorphism on $\rset^\dimx$. 
\end{hypA}
\begin{mdframed}[style=propFrame]
\begin{proposition} 
      Assume \assp{\ref{assp:diffeo}}. Then the Markov chain $(\iX{r}{0})_{r \in \nset}$ is $\Leb$-irreducible and has $\cpdata{0}{\cond; \scale}{}$ as its unique invariant distribution. 
\end{proposition}
\end{mdframed}
\begin{proof} 
  Let $\msa$ be a Borel set of $\rset^d$ such that $\Leb(\msa) > 0$. It is enough to show that for any $\bx \in \rset^\dimx$,
  \begin{equation}
    \label{eq:proof_irred_1}
    \int \indic_{\msa}(F_{0|\stds}(\bz | \cond; \scale)) \normpdf(\bz; \bx, \Id) \, \rmd \bz > 0\eqsp.
  \end{equation}
  
  Denote by $\bz \mapsto F^{\inv} _{0|\stds}(\bz|\cond; \scale)$ the inverse of $\bz \mapsto F _{0|\stds}(\bz|\cond; \scale)$. In addition, denote by $\det \, \mbox{Jac} \, \Phi$, the Jacobian determinant of any function $\Phi : \rset^d \to \rset^d$ and note that since $\bz \mapsto F _{0|\stds}(\bz|\cond; \scale)$ is a diffeomorphism, it holds that $\bz \mapsto \det \, \mbox{Jac}  F_{0|\stds}(\bz|\cond; \scale)$ and $\bz \mapsto \det \, \mbox{Jac} F^{\inv} _{0|\stds}(\bz|\cond; \scale)$ are both non-zero. Then a change a variables yields 
  \begin{equation*}
    \label{eq:2}
    \int \indic_{\msa}(F_{0|\stds}(\bz | \cond; \scale)) \normpdf(\bz; \bx, \Id) \, \rmd \bz = \int \indic_{\msa}(\by) \normpdf(F^{\inv} _{0|\stds}(\by|\cond; \scale); \bx, \Id) \, |\det \, \mbox{Jac} F^{\inv} _{0|\stds}(\by|\cond; \scale)| \,  \rmd \by\eqsp,
  \end{equation*}
  which implies \eqref{eq:proof_irred_1}.
  Finally, the fact that the Markov chain $(\iX{r}{0})_{r \in \nset}$ has $\cpdata{0}{\cond; \scale}{}$ as its unique invariant distribution follows from \cite[Proposition~10.1.1. and Theorem~10.4.4]{meyn2012markov}.
\end{proof}
\paragraph{\Cref{ex:gaussian-cfg} details.}

Recall that $\cpdata{\std}{\cond; \scale}{\bx_\std}[\cfg] \propto 
     \clf{\std}{\bx_\std}{\cond}^\scale \pdata{\std}{}{\bx_\std}$, where $\clf{\std}{\bx_\std}{\cond} \eqdef \int \clf{0}{\bx_0}{\cond}\,\pdata{0|\std}{\bx_\std}{\bx_0}\,\rmd \bx_0$. In addition, note that since $\pdata{0}{}{} = \gauss(0, 1)$, it holds that $\pdata{0|\std}{\bx_\std}{\bx_0} = \mathrm{N}(0; \bx_\std/(1+\sigma^2),\sigma^2/(1+\sigma^2))$. Since $\clf{0}{\bx_0}{\cond} = \normpdf(\cond; \bx_0, \gamma^2)$, this implies that $\clf{\std}{\bx_\std}{\cond} = \mathrm{N}(\cond; \bx_\std/(1+\sigma^2),\gamma^2+\sigma^2/(1+\sigma^2))$. Therefore, we get $\cpdata{\std}{\cond; \scale}{\bx_\std}[\cfg] = \mathrm{N}(\bx_{\std}; \mathrm{v}(\scale,\sigma^2) w \cond / ((1+\sigma^2)\gamma^2+ \sigma^2),\mathrm{v}(\scale,\sigma^2))$, where  
$$
\mathrm{v}(\scale,\sigma^2) \eqdef (1 + \std^2) \frac{(1 + \std^2)\gamma^2 + \std^2}{\scale + \gamma^2 (1 + \std^2) + \sigma^2} \eqsp.
$$

The limit part follows by a simple characteristic-function argument, since $\cpdata{\std}{\cond; \scale}{}[\cfg]$ is a Gaussian distribution and we make the assumption that it is the result of a Gaussian convolution. It remains to show that $\mathrm{v}(\scale,\sigma^2) < \pV(X_\std)$, where we recall that 
 $\pV(X_\std) = \std^2 + \gamma^2 / (\gamma^2 + \scale)$. However, since $\scale > 1$, we have that 
\begin{align*}
      (1 + \std^2)^{-1} \pV(X_\std) & = \frac{\gamma^2 (1 + \std^2) + \std^2 \scale}{\scale + \gamma^2 (1 + \std^2) + \scale \std^2}  \\
      & = 1 - \frac{\scale}{\scale + \gamma^2 (1 + \std^2) + \scale \std^2} \\
      &> 1 - \frac{\scale}{\scale + \gamma^2 (1 + \std^2) + \std^2} \\
      & = (1 + \std^2)^{-1} \mathrm{v}(\scale,\sigma^2) \eqsp.
\end{align*}

\paragraph{Proofs for \Cref{prop:variance-bias}.}
\label{subsec:gaussian_case}
We assume that $\pdata{0}{}{} = \gauss(\zero, 1)$ and $\potg{0}{\bx_0}{\cond} = \normpdf(\cond; \bx_0, \gamma^2)$. Then it holds that 
\begin{equation*}
      \denoiser{\std}{}{\bx_\std} = \frac{1}{1 + \std^2} \bx_\std \eqsp,
      \quad \denoiser{\std}{\cond}{\bx_\std} = \frac{\gamma^2}{\gamma^2 (1 + \std^2) + \std^2} \bx_\std + \frac{\std^2}{\gamma^2 (1 + \std^2) + \std^2} \cond \eqsp. 
\end{equation*}
The next lemma provides the expression of the distribution obtained by integrating exactly the PF-ODE \eqref{eq:pf-ode} from $\std$ to zero using the CFG denoiser \eqref{eq:cfg-denoiser}.
We assume $\cond=0$, as the solution to the PF-ODE cannot, in general, be expressed using elementary functions when $\cond \neq 0$. 
In that setting, we have the following standard result. 
\begin{mdframed}[style=propFrame]
\begin{lemma} 
      \label{lem:ode-solution}
Let $\scale >1$ and consider the first-order linear homogeneous ODE 
\begin{equation}
      \label{eq:homog-ODE}
\frac{\rmd \bx(\std)}{\rmd \std} = \left[ \frac{\scale (1 + \gamma^2) \std}{\gamma^2 + (1 + \gamma^2) \std^2} + \frac{(1- \scale)\std}{1 + \std^2} \right] \bx(\std) \eqsp, \quad \std \ge 0. 
\end{equation}
Let $\bx$ be the unique solution to \eqref{eq:homog-ODE} on the interval $[0, \std_*]$, for some $\std_* > 0$, satisfying the condition $\bx(\stds) = X_{\stds}$. Then the value of the solution at $\std=0$ is given by
\begin{equation}
X_0 = \frac{\gamma^\scale (1 + \std^2 _*)^{(\scale - 1) / 2}}{(\gamma^2 + (1 + \gamma^2) \std^2 _*)^{\scale / 2}} X_\stds \eqsp.
\end{equation}
\end{lemma}
\end{mdframed}
\begin{proof}
      We define
      \begin{equation} 
      P(\sigma) = \frac{w(1+\gamma^2)\sigma}{\gamma^2 + (1+\gamma^2)\sigma^2} + \frac{(1-w)\sigma}{1+\sigma^2} \eqsp.
      \end{equation}
      The function $P$ is continuous on $[0, \stds]$ and thus, by the Cauchy--Lipschitz theorem, it admits a unique solution satisfying $\bx(\stds) = X_{\stds}$. If $\bx(\stds) = 0$, then the unique solution to \eqref{eq:homog-ODE} with $\bx(\stds) = 0$ satisfies $\bx(\std) = 0$ for any $\std \in [0, \stds]$. 
      
      Now, assume $\bx({\stds}) \neq 0$. Then $\bx(\std) \neq 0$ for any $\std \in [0, \stds]$, and integrating from $\sigma_*$ to zero yields 
      \begin{equation*} 
      \int_{X_{\sigma_*}}^{X_0} \frac{\rmd \bx(\std)}{\bx(\std)} = \int_{\stds}^0 P(s) \, \rmd s = F(0) - F(\stds),
      \end{equation*}     
      where $F(s) = \frac{w}{2} \log(\gamma^2 + (1+\gamma^2)s^2) + \frac{1-w}{2} \log(1+s^2)$ for any $s \in [0, \stds]$. Note that $F(0) = \log \gamma^\scale$ and 
      $$ 
 F(\sigma_*) = \frac{\scale}{2} \log(\gamma^2 + (1+\gamma^2)\sigma_*^2) + \frac{1-\scale}{2} \log(1+\sigma_*^2) \eqsp; 
$$    
      thus, 
      $$ 
            \log \left|\frac{X_0}{X_{\sigma_*}}\right| = \log \frac{\gamma^\scale}{(\gamma^2 + (1+\gamma^2)\sigma_*^2)^{\scale/2} (1+\sigma_*^2)^{(1-\scale)/2}} \eqsp. 
      $$
      Applying the exponential function to both sides and rearranging terms leads to
      $$ 
      \left|\frac{X_0}{X_{\sigma_*}}\right| = \frac{\gamma^\scale (1+\sigma_*^2)^{(\scale - 1)/2}}{(\gamma^2 + (1+\gamma^2)\sigma_*^2)^{\scale/2}} \eqsp. 
      $$
      By continuity of the solution and the fact that $\bx(\std) \neq 0$ for any $\std$, $\bx$ has a fixed sign on $[0, \stds]$, which yields the result.
\end{proof}
As a sanity check, note that when $\scale = 0$, it holds that $X_0 = X_{\std_*} / \sqrt{1 + \std^2 _*}$. Thus, if $X_\stds \sim \pdata{\stds}{}{} = \gauss(0, 1 + \vars)$, then $X_0 \sim \pdata{0}{}{}$. If $\scale = 1$, we get $X_0 = \gamma X_\stds / \sqrt{\gamma^2 + (1 + \gamma^2) \std^2 _*}$, and hence, if $X_\std \sim \cpdata{\stds}{\cond}{} = \gauss(0, \vars + \gamma^2 / (1 + \gamma^2))$, then $X_0 \sim \cpdata{0}{\cond}{}$. 

\paragraph{Analysis as $\stds \to 0$ and $R \to \infty$.} We now consider the iterative procedure defined by, for any $r \geq 1$,
\begin{equation}
      \label{eq:recursion}
      \iX{r}{0} = c(\stds) \big( \iX{r-1}{0} + \stds Z^{(r)}) \eqsp, 
\end{equation}
where $c(\stds) \eqdef \gamma^\scale (1 + \std^2 _*)^{(\scale - 1) / 2} \big / (\gamma^2 + (1 + \gamma^2) \std^2 _*)^{\scale / 2}$ and $(Z^{(r)}) _{r \in \nset}$ is a sequence of i.i.d. standard Gaussian random variables. Define $V_\infty(\scale) \eqdef \vars c(\stds)^2 / (1 - c(\stds)^2)$.

\begin{figure} 
      \center
      \includegraphics[width=.6\textwidth]{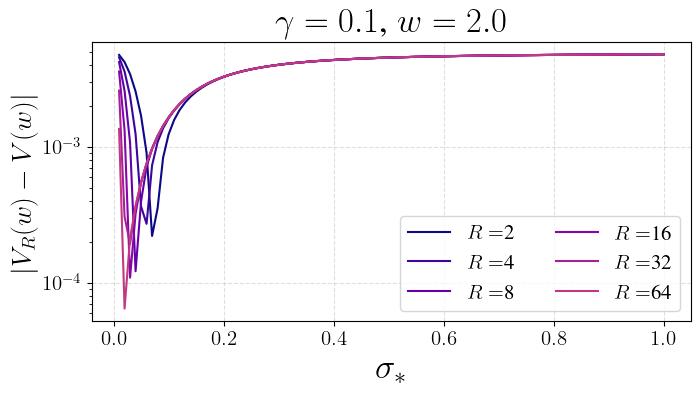}
      \includegraphics[width=.6\textwidth]{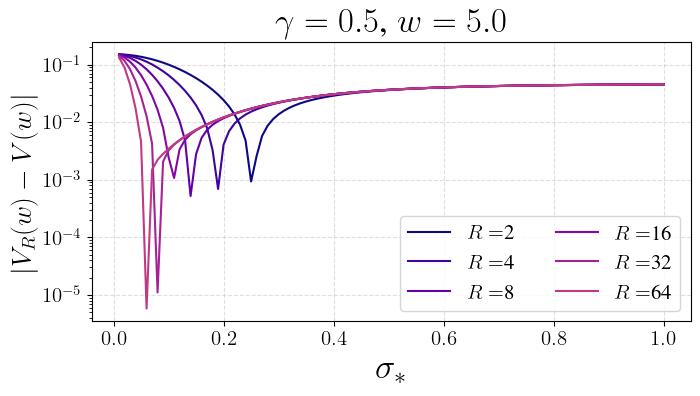}
      \caption{The difference $|V_R(\scale) - V(\scale)|$ as a function of $\stds$ for various values of $R$.}
      \label{fig:mixing}
\end{figure}

We denote by $\mathrm{W}_2(\mu, \nu)$ the 2-Wasserstein distance between distributions $\mu$ and $\nu$ in $\mathcal{P}(\rset^\dimx)$. 
\begin{mdframed}[style=propFrame]
\begin{proprestate}[Restatement of \Cref{prop:variance-bias}]
      Assume that $\pV(\iX{0}{0}) < \infty$. For any $\scale > 1$, $(\iX{r}{0})_{r\in \nset}$ converges to $\gauss(0, V_\infty(\scale))$ exponentially fast in the 2-Wasserstein distance, with rate proportional to $\vars$. Furthermore,
      \begin{equation*}
          V_\infty(\scale) = V(\scale) + O(\std^2 _*) \quad \text{as \, $\vars \to 0$}
          \eqsp.
      \end{equation*}
\end{proprestate}
\end{mdframed}
\begin{proof} 
      We let $\mu_r \eqdef \law(\iX{r}{0})$ and $\mu_\infty \eqdef  \gauss(\zero, V_\infty(\scale))$. Defining the transition kernel $k(\bx|\tilde\bx) = \normpdf(c(\stds) \tilde\bx, c(\stds)^2 \vars)$, it holds that $\mu_r(\bx) \eqdef \int k(\bx | \tilde\bx) \, \mu_{r-1}(\rmd \tilde\bx)$.
      Furthermore, it is easily checked that $\mu_\infty$ is the stationary distribution of the chain \eqref{eq:recursion}. Hence, $\mu_\infty(\bx) = \int k(\bx|\tilde\bx) \, \mu_\infty(\rmd \tilde\bx)$. Then, 
      \begin{align*}
            \mathrm{W}_2( \mu_r, \mu_\infty) & = \mathrm{W}_2\left( \int k(\cdot|\tilde\bx) \, \mu_{r-1}(\rmd \tilde\bx), \int k(\cdot|\tilde\bx) \, \mu_{\infty}(\rmd \tilde\bx) \right)\\
            & \leq c(\stds) \mathrm{W}_2\left(\int \fw{\stds|0}{\tilde\bx}{} \, \mu_{r-1}(\rmd \tilde\bx), \int \fw{\stds|0}{\tilde\bx}{} \, \mu_\infty(\rmd \tilde\bx)\right) \\
            & \leq c(\stds) \mathrm{W}_2(\mu_{r-1}, \mu_\infty \big) \eqsp,
      \end{align*}
where we used, in the second step, the fact that for any $c > 0$, if $X \sim \mu$ and $Y \sim \nu$, then $\mathrm{W}_2(\law(c X), \law(cY)) \leq c \mathrm{W}_2(\mu, \nu)$. Moreover, in the third step we used the fact that convolution does not increase the Wasserstein distance \cite[Lemma 5.2]{santambrogio2015optimal}. Iterating the bound we obtain that $\mathrm{W}_2(\mu_r, \mu_\infty) \leq c(\stds)^r \mathrm{W}_2(\mu_0, \mu_\infty)$, and as $\mu_0$ is square-integrable by assumption, $\mathrm{W}_2(\mu_0, \mu_\infty) < \infty$. Thus, since 
      $$ 
      c(\stds) = \frac{(\gamma^2 + \gamma^2 \vars)^{\scale / 2}}{\sqrt{1 + \vars} (\gamma^2 + \gamma^2 \vars + \vars)^{w/2}} < 1 \eqsp,
      $$ 
    we establish the exponential convergence of $(\mu_r)_{r \in \nset}$ to $\mu_\infty$.  
Next, making a Taylor expansion of $V_\infty(\scale)$ around $\vars = 0$ yields that 
\begin{align*} 
      \log c(\stds) & = \scale \log \gamma + \frac{\scale - 1}{2} \log(1 + \vars) - \frac{\scale}{2} \log(\gamma^2 + (1 + \gamma^2) \vars) \\ 
      & = \frac{\scale - 1}{2} \log (1 + \vars) - \frac{\scale}{2} \log \left(1 + \frac{1 + \gamma^2}{\gamma^2} \vars \right) \\
      & = \left( \frac{\scale - 1}{2} - \scale \frac{1 + \gamma^2}{2 \gamma^2} \right) \vars + o(\vars )\\ 
      & = - \frac{\gamma^2 + \scale}{2 \gamma^2} \vars + o(\vars) \eqsp, 
\end{align*} 
which in turn yields the expansion $
      1 - c(\stds)^2 = \frac{\gamma^2 + \scale}{\gamma^2} \vars + o(\vars)$. Substituting the same into  $V_\infty(\scale)$ gives  
\begin{align*}
      V_\infty(\scale) = \frac{\vars \left( 1 - \frac{\gamma^2 + \scale}{\gamma^2} \vars + o(\vars) \right)}{\frac{\gamma^2 + \scale}{\gamma^2} \vars + o(\vars)} 
      = \frac{1 - \frac{\gamma^2 + \scale}{\gamma^2} \vars + o(\vars)}{\frac{\gamma^2 + \scale}{\gamma^2} + o(1)} \eqsp.
\end{align*}
      Developing the right-hand side further, we arrive at
      \begin{align*}
            V_\infty(\scale) &= \left( 1 - \frac{\gamma^2 + \scale}{\gamma^2} \vars + o(\vars) \right) \left( \frac{\gamma^2 + \scale}{\gamma^2} \right)^{-1} \left( 1 + \frac{o(\vars)}{\frac{\gamma^2 + \scale}{\gamma^2}} \right)^{-1} \\
            &= \frac{\gamma^2}{\gamma^2 + \scale} \left( 1 - \frac{\gamma^2 + \scale}{\gamma^2} \vars + o(\vars) \right) (1 + o(\vars)) \\
            &= \frac{\gamma^2}{\gamma^2 + \scale} \left( 1 - \frac{\gamma^2 + \scale}{\gamma^2} \vars + o(\vars) \right) \\
            &= \frac{\gamma^2}{\gamma^2 + \scale} - \vars + o(\vars) \eqsp, 
            \end{align*}
            from which the claim follows. 
\end{proof}
\paragraph{Analysis as $\stds \to 0$ and $R < \infty$.}
To investigate the behavior at finite R, we initialize the recursion \eqref{eq:recursion} with \(\iX{0}{0} \sim \cpdata{0}{\cond}{} = \gauss(\zero, V_0)\), where \(V_0 \coloneqq \gamma^2 / (\gamma^2 + 1)\). For any R > 0, the final iterate satisfies \(\iX{R}{0} \sim \gauss(\zero, V_R(\scale))\), where
\[
V_R(\scale) = c(\stds)^{2R} V_0 + \frac{c(\stds)^2 \, \std^2\, (1 - c(\stds)^{2R})}{1 - c(\stds)^2}.
\]
We plot the absolute error \(|V_R(\scale) - V(\scale)|\) as a function of \(\stds\) for various values of R in \Cref{fig:mixing}. We observe that for small R, the error is minimized at higher values of \(\stds\), and that the minimal error achieved for small \(\stds\) is lower than that observed at larger \(\stds\).

%% file: appendix/related.tex
\section{On related works}
\label{subsec:further-related-works}
\paragraph{\CFGPP\, with Heun sampler.}
Let $\lambda \in [0,1]$. Following \cite[Appendix A]{chung2025cfg}, \CFGPP\, with the Euler discretization of the PF-ODE \eqref{eq:ddim} corresponds to the update 
$$ 
    X_t = \denoiser{t+1}{\cond; \lambda}{X_{t+1}}[\cfg] + \frac{\std_t}{\std_{t+1}} \big( X_{t+1} - \denoiser{t+1}{}{X_{t+1}} \big),  
$$ 
which, upon replacing with the definition of the CFG denoiser \eqref{eq:cfg-denoiser}, writes as 
\begin{align*}
    X_t & = \bigg[ (1 - \frac{\std_t}{\std_{t+1}}) \denoiser{t+1}{}{X_{t+1}} + \lambda \big( \denoiser{t+1}{\cond}{X_{t+1}} - \denoiser{t+1}{}{X_{t+1}}\big) \bigg] + \frac{\std_t}{\std_{t+1}} \denoiser{t+1}{}{X_{t+1}} \\
    & = \frac{\std_t}{\std_{t+1}} X_{t+1}  +  \big(1 - \frac{\std_t}{\std_{t+1}}\big) \denoiser{t+1}{\cond; \scale^{_{++}} _{t+1}}{X_{t+1}}[\cfg], 
\end{align*}
with $\scale^{_{++}} _{t+1} = \lambda \std_{t+1} / \big( \std_{t+1} - \std_t \big)$. In short, \CFGPP\, can be viewed as a DDIM update based on the CFG denoiser with a $\sigma$-dependent guidance scale. Hence, in our experiment, we implement \CFGPP\ with Heun sampler by using \Cref{algo:heun} with the CFG denoiser \eqref{eq:cfg-denoiser} and time-dependent guidance factors $(\scale^{_{++}} _{t})_{t=1} ^T$. 

\paragraph{Feynman--Kac correctors.} In \cite{skreta2025feynman}, a sequential Monte Carlo (SMC) algorithm is proposed to target the distribution sequence $(\cpdata{\std}{\cond; \scale}{}[\cfg])_{\std}$. In what follows, we provide a simple derivation of the (approximate) recursion relating two consecutive distributions $\cpdata{\std_t}{\cond; \scale}{}[\cfg]$ and $\cpdata{\std_{t+1}}{\cond; \scale}{}[\cfg]$, for $\std_t < \std_{t+1}$. This serves as the basis for the SMC algorithm resulting from \cite[Proposition 3.1]{skreta2025feynman}. Write   
\begin{align} \label{eq:fk_correctors_approx}
    \cpdata{\std_t}{\cond; \scale}{\bx_t}[\cfg] & \propto \int \cpdata{\std_t}{\cond; \scale}{\bx_t}[\cfg] \fw{\std_{t+1}|\std_t}{\bx_t}{\bx_{t+1}}  \, \rmd \bx_{t+1} \nonumber \\
    & \propto \int \cpdata{\std_t}{\cond}{\bx_t}^\scale \pdata{\std_t}{}{\bx_t}^{1 - \scale} \fw{\std_{t+1}|\std_t}{\bx_t}{\bx_{t+1}}  \, \rmd \bx_{t+1} \nonumber \\
    &  \propto \int \big[ \cpdata{\std_t}{\cond}{\bx_t} \fw{\std_{t+1}|\std_t}{\bx_t}{\bx_{t+1}}\big]^\scale [\pdata{\std_t}{}{\bx_t} \fw{\std_{t+1}|\std_t}{\bx_t}{\bx_{t+1}}]^{1 - \scale}  \, \rmd \bx_{t+1}  \nonumber\\
    & \propto \int \cpdata{\std_t|\std_{t+1}}{\bx_{t+1}, \cond}{\bx_t}^\scale \pdata{\std_t|\std_{t+1}}{\bx_{t+1}}{\bx_t}^{1 - \scale} \cpdata{\std_{t+1}}{\cond; \scale}{\bx_{t+1}}[\cfg] \, \rmd \bx_{t+1} \\
    & \propto \int \mathcal{Z}^\cfg _{t+1}(\bx_{t+1}; \scale) \cpdata{\std_t|\std_{t+1}}{\bx_{t+1}, \cond; \scale}{\bx_t}[\cfg] \cpdata{\std_{t+1}}{\cond; \scale}{\bx_{t+1}}[\cfg] \, \rmd \bx_{t+1}, 
\end{align}
where we have defined 
\begin{align}
    \label{eq:cpdata_def}
    \cpdata{\std_t|\std_{t+1}}{\bx_{t+1}, \cond}{\bx_t} & \eqdef \frac{\cpdata{\std_t}{\cond}{\bx_t} \fw{\std_{t+1}|\std_t}{\bx_t}{\bx_{t+1}}}{\cpdata{\std_{t+1}}{\cond}{\bx_{t+1}}} \eqsp, \\
    \pdata{\std_t|\std_{t+1}}{\bx_{t+1}}{\bx_t} & \eqdef \frac{\pdata{\std_t}{}{\bx_t} \fw{\std_{t+1}|\std_t}{\bx_t}{\bx_{t+1}}}{\pdata{\std_{t+1}}{}{\bx_{t+1}}} \eqsp, \label{eq:pdata_def} \\
    \cpdata{\std_t|\std_{t+1}}{\bx_{t+1}, \cond}{\bx_t}[\cfg] & \propto  \cpdata{\std_t|\std_{t+1}}{\bx_{t+1}, \cond}{\bx_t}^\scale \pdata{\std_t|\std_{t+1}}{\bx_{t+1}}{\bx_t}^{1 - \scale} \eqsp, \label{eq:pi:cfg:kernel} 
\end{align}
and $\mathcal{Z}_{t+1}^\cfg(\bx_{t+1}; \scale) \eqdef \int \cpdata{\std_t|\std_{t+1}}{\bx_{t+1}, \cond}{\bx_t}^\scale \pdata{\std_t|\std_{t+1}}{\bx_{t+1}}{\bx_t}^{1 - \scale} \, \rmd \bx_{t+1}$. 

The recursion in \cite{skreta2025feynman} approximates \eqref{eq:fk_correctors_approx} by replacing \eqref{eq:cpdata_def} and \eqref{eq:pdata_def} with their Gaussian approximations. More precisely, consider, \emph{e.g.}, the Gaussian approximations
\begin{align*} 
    \cpdata{\std_t|\std_{t+1}}{\bx_{t+1}, \cond}{\bx_t} & \approx \normpdf(\bx_t; \mu_{t|t+1}(\bx_{t+1} | \cond),\std^2 _{t|t+1} \Id) \eqsp,\\
    \pdata{\std_t|\std_{t+1}}{\bx_{t+1}}{\bx_t} & \approx \normpdf(\bx_t; \mu_{t|t+1}(\bx_{t+1}),\std^2 _{t|t+1} \Id)
\end{align*}
stemming from DDPM, where 
\begin{align*}
    \mu_{t|t+1}(\bx_{t+1}|\cond) & \eqdef \frac{\std^2_t}{\std^2_{t+1}}X_{t+1}  + \left(1 - \frac{\std^2_t}{\std^2_{t+1}}\right) \denoiser{\std_{t+1}}{\cond}{X_{t+1}} \eqsp,\\
    \mu_{t|t+1}(\bx_{t+1}) & \eqdef \frac{\std^2_t}{\std^2_{t+1}}X_{t+1}  + \left(1 - \frac{\std^2_t}{\std^2_{t+1}}\right) \denoiser{\std_{t+1}}{}{X_{t+1}} \eqsp, 
\end{align*}
and $\std^2 _{t|t+1} \eqdef  \std^2 _t  (\std^2 _{t+1} - \std^2 _t) / \std^2 _{t+1}$. The Gaussian approximation of $\cpdata{\std_t|\std_{t+1}}{\bx_{t+1}, \cond}{}[\cfg]$ is formed by plugging the two previous Gaussian approximations into \eqref{eq:pi:cfg:kernel}. When two Gaussian densities $\mathcal{N}(\bx_t; m_1, \Sigma)$ and $\mathcal{N}(\bx_t; m_2, \Sigma)$ are raised to powers $\scale$ and $1-\scale$, respectively, their product is proportional to another Gaussian density with the same covariance matrix $\Sigma$ and a new mean $\scale m_1 + (1-\scale) m_2$; hence, 
$$ 
\cpdata{\std_t|\std_{t+1}}{\bx_{t+1}, \cond}{\bx_t}[\cfg] \approx \normpdf\left(\bx_t; \frac{\std^2 _t}{\std^2 _{t+1}} \bx_{t+1} + \left(1 - \frac{\std^2 _t}{\std^2 _{t+1}}\right) \denoiser{\std_{t+1}}{\cond}{\bx_{t+1}}[\cfg], \std^2 _{t|t+1} \Id\right) \eqsp.
$$ 
It remains to derive an approximation of the weighting term $\mathcal{Z}^\cfg _{t+1}(\bx_{t+1}; \scale)$. Plugging again the two previous Gaussian approximations into the definition of the latter yields 
\begin{align*}
\mathcal{Z}^\cfg _{t+1}(\bx_{t+1}; \scale) & \approx \int \normpdf(\bx_t; \mu_{t|t+1}(\bx_{t+1} | \cond), \std^2 _{t|t+1} \Id)^\scale \normpdf(\bx_t; \mu_{t|t+1}(\bx_{t+1}),\std^2 _{t|t+1} \Id)^{1 - \scale} \, \rmd \bx_{t} \\
& \propto \mbox{exp}\left(\frac{\scale(\scale - 1)(\std^2 _{t+1} - \std^2 _t)}{2 \std^2 _{t+1} \std^2 _t} \big\| \denoiser{\std_{t+1}}{\cond}{\bx_{t+1}} - \denoiser{\std_{t+1}}{}{\bx_{t+1}} \big\|^2 _2 \right) \eqsp.
\end{align*}

%% file: appendix/exp_details.tex
\label{apdx:exp-details-images}
\begin{algorithm}
    \caption{Heun sampler}
    \begin{algorithmic}
       \STATE {\bfseries Require:} Noise levels $\{ \std_t \}_{t = 0} ^T$
       \STATE $X_T \sim \gauss(\zero, \std_T ^2 \Id)$
       \FOR{$t = T-1$ {\bfseries to} $1$}
            \STATE $D_{t+1} \gets \denoiser{\std_{t+1}}{}{X_{t+1}}$
            \STATE $X^\prime _t \gets (\std_t / \std_{t+1}) X_{t+1} + (1 - \std_t / \std_{t+1}) D_{t+1}$
            \STATE $X_t \gets X_{t+1} + \frac{\std_t - \std_{t+1}}{2} \big[ \frac{X^\prime _t - \denoiser{\std_t}{}{X^\prime _t}}{\std_t} + \frac{X_{t+1} - D_{t+1}}{\std_{t+1}}\big]$
       \ENDFOR
     \STATE {\bfseries Output:} $\denoiser{\std_1}{}{X_1}$
    \end{algorithmic}
    \label{algo:heun}
\end{algorithm}

\subsection{Algorithms}

\emph{Implementation.}
We implement \algoname\ as described in \Cref{algo:algo}. Notably, since at each Gibbs sampling iteration, we simulate the PF-ODE over a newly redefined interval $[0, \stds]$, it is possible to adapt the sequence of noise levels for it.
We use the noise-scheduling method proposed in~\cite[Eqn.~(269)]{karras2022elucidating}, \emph{i.e.}, for some $\rho \geq 1$, 
\begin{equation*}
    \std_t = \bigg( \std_0^{1/\rho} + \frac{t}{T-1} (\std_{\max}^{1/\rho} - \std_0^{1/\rho}) \bigg)^{\rho}
    \enspace.
\end{equation*}
For \imagenet, we use $\rho = 7$ and recompute the discretization steps accordingly for the Gibbs iterations starting from $\std_{\max}=\stds$.
For AudioLDM and Stable Diffusion, which are DDMs defined through the variance-preserving (VP) formulation, we found that selecting uniformly over the diffusion steps and then mapping these diffusion steps to noise levels yields better performance.
For \CFGPP, we leverage the equivalence between the proposed algorithm and standard CFG with a dynamic guidance scale, as described in~\cite[Appendix B]{chung2025cfg}.
For \limitedCFG, we implement the procedure as defined in~\cite[Equations 5 and 6]{kynkanniemi2024applying}.

\emph{Hyperparameters.}
For each algorithm, we perform a grid search over its hyperparameter space and select the configuration that optimizes FID for image generation and FAD for audio generation. The selected hyperparameters are summarized in \Cref{table:hyperparams-algo} for \algoname\ and in \Cref{table:hyperparams-competitors} for the other algorithms.

The experiments were run on Nvidia H100 SXM5 80 GB.
In \Cref{table:time-comparison}, we provide  a runtime comparison between the algorithms.
The runtimes were averaged over a run to generate 50k images with a batch size of 500 images.

\subsection{Image experiments}
\emph{Models description.\,}
We employ the class conditional latent diffusion models introduced in~\cite{karras2024edm2} and trained on \imagenet-$512$ dataset.
These model operate in a latent space, with dimension $4 \times 64 \times 64$, defined by the pre-trained VAE~\cite{rombach2022high}.
The experiments are performed using two model sizes: small (EDM-S) and largest (EDM-XXL); see~\cite[Table 2]{karras2024edm2} for their model sizes.
We use the publicly available weights\footnote{\url{https://github.com/NVlabs/edm2}}, more specifically the models under the pseudos \texttt{edm2-img512-s-guid-fid} and \texttt{edm2-img512-xxl-guid-fid}.
These weights were obtained by tuning the EMA length to optimize FID as detailed in \cite[Section 3]{karras2024edm2}.

\emph{Evaluation.\,} We assess the overall quality of the generated samples by computing the Fréchet distance between the reference and generated samples.
We follow the recommandations in \cite{stein2023fd_dinvo2} and report, in addition to the FID, the FD\textsubscript{DINOv2}; in the former the images representations are computed using InceptionV3~\cite{szegedy2016inceptionv3}, whereas in the latter they are computed with DINOv2 ViT-L/14~\cite{oquab2023dinov2}.
We reuse the precomputed statistics provided in EDM2 repository\footnotemark[1] for the reference distribution, and use, following common practices~\cite{stein2023fd_dinvo2}, 50k samples to compute the statistics of the generative distribution.
We also evaluate individually the fidelity and diversity of the generated samples using Precision/Recall and Density/Coverage~\cite{naeem2020prdc}.
We follow the setting in~\cite{kynkanniemi2024applying}, where the manifold of the data is estimated using 50k samples with representations computed using DINOv2 ViT-L/14~\cite{oquab2023dinov2} and using 3 neighbors, except that for the reference images, where we use \imagenet\ validation set instead of training set.

\subsection{Audio experiments}
\label{apdx:exp-details-audio}

\emph{Model description.\,} AudioLDM 2-Full-Large~\cite{liu2024audioldm2} is a latent diffusion model tailored for general-purpose audio generation. It operates on compressed representations of mel-spectrograms obtained through a variational autoencoder. The model uses a self-supervised AudioMAE~\cite{xu2022masked} to extract a semantic embedding known as the Language of Audio (LOA), which captures both acoustic and semantic information. A GPT-2 language model \cite{radford2019language} translates text prompts into LOA features, which then condition the diffusion model for audio generation in a computationally efficient latent space.
We use the publicly available pre-trained model at HuggingFace\footnote{\url{https://huggingface.co/cvssp/audioldm2}} under the pseudo name \texttt{audioldm2-large}.

\emph{Dataset and evaluation.\,} We sampled 1k prompts randomly from the \audiocaps~\cite{kim2019audiocaps} test set.
Following \cite{liu2024audioldm2}, each algorithm was run with a negative prompt \textit{``low quality''} along side the conditioning prompt and the best out of 3 samples was selected for each conditioning prompt.
The evaluation was performed on 10 seconds audios at 16 kHz.
The VGGish model~\cite{hershey2017cnn} was used to extract the features to be used to compute FAD.
Similarly, PANNs~\cite{kong2020panns} was used to compute the features for the KL and IS metrics.

\begin{table}[t]
    \centering
    \captionsetup{font=small}
    \caption{\algoname\ hyperparameters for the considered models.
    The symbol \# stands for \emph{``number of''}.}
    \renewcommand{\arraystretch}{2.} 
    \resizebox{\textwidth}{!}{
    \begin{tabular}{l ccccccc}
        \toprule
        & \parbox[m]{16em}{\centering Sampler} & \# Total steps & \# Initial steps & \parbox[m]{10em}{\centering Initial\\  guidance scale} & \parbox[m]{6em}{\centering \# Gibbs\\repetitions} & \parbox[m]{8em}{\centering Gibbs\\ noise level} & \parbox[m]{8em}{\centering Gibbs\\Guidance scale} \\
        \midrule
        EDM-S & \multirow{2}{*}{
                Heun with $\begin{cases}
                \std_0 & = 2\times 10^{-3}\\
                \std_{\max} & = 80 \\
                \end{cases}$
            } & \multirow{2}{*}{$T=32$} & $T_0=12$ & $\scale_0=1$ & $R=2$ & $\stds=2$ & $\scale=2.3$
        \\
        EDM-XXL & & & $T_0=12$ & $\scale_0=1$ & $R=2$ & $\stds=1$ & $\scale=2$
        \\
        \midrule
        AudioLDM 2-Full-Large &  DDIM with $\begin{cases}
            \std_0 & = 3.88\times 10^{-2}\\
            \std_{\max} & = 83.33 \\
            \end{cases}$ & $T=200$ & $T_0=100$ & $\scale_0=1.5$ & $R=2$ & $\stds=5$ & $\scale=5$
        \\
        \bottomrule
    \end{tabular}
    \label{table:hyperparams-algo}
    }
\end{table}

\begin{table}[t]
    \centering
    \captionsetup{font=small}
    \caption{Competitors' hyperparameters for the considered models.
    The symbol \# stands for \emph{``number of''}.}
    \renewcommand{\arraystretch}{1.5} 
    \resizebox{\textwidth}{!}{
    \begin{tabular}{l cc c | cc | c}
        \toprule
        & & & \multicolumn{1}{c}{\CFG} & \multicolumn{2}{c}{\limitedCFG} & \CFGPP \\
        \cmidrule{4-7}
        &  Sampler & \# Total steps & Guidance scale & Guidance interval &  Guidance scale &  Guidance factor  \\
        \midrule
        EDM-S &
            \multirow{2}{*}{
                Heun with $\begin{cases}
                \std_0 & = 2\times 10^{-3}\\
                \std_{\max} & = 80 \\
                \end{cases}$
            }
            & \multirow{2}{*}{$T=32$} & $\scale=1.4$ & $[\std_{\text{lo}}, \std_{\text{hi}}] = [0.28, 2.9]$ & $\scale=2.1$ & $\lambda=0.35$
        \\
        EDM-XXL & & & $\scale=1.2$ & $[\std_{\text{lo}}, \std_{\text{hi}}] = [0.19, 1.61]$ & $\scale=2$ & $\lambda=0.35$
        \\
        \midrule
        AudioLDM 2-Full-Large &  DDIM with $\begin{cases}
            \std_0 & = 3.88\times 10^{-2}\\
            \std_{\max} & = 83.33 \\
            \end{cases}$& $T=200$ & $\scale=4.5$ & $[\std_{\text{lo}}, \std_{\text{hi}}] = [0, 8.5]$ & $\scale=5$ & $\lambda=0.1$
        \\
        \bottomrule
    \end{tabular}
    \label{table:hyperparams-competitors}
    }
\end{table}

\begin{table}[t]
    \centering
    \captionsetup{font=small}
    \caption{Runtime in seconds of each algorithm on a batch size of $500$ images.}
    \renewcommand{\arraystretch}{1.5} 
    \resizebox{0.4\textwidth}{!}{
    \begin{tabular}{l cccc}
        \toprule
        & \CFG & \limitedCFG & \CFGPP & \algoname \\
        \midrule
        EDM-S & 907 & 638 & 811 & 760
        \\
        EDM-XXL & 1623 & 1334 & 1480 & 1445
        \\
        \bottomrule
    \label{table:time-comparison}
    \end{tabular}
    }
\end{table}

\input{tables/bootstrap.tex}

%% file: tables/bootstrap.tex
\begin{table}
        \centering
        \captionsetup{font=small}
        \captionof{table}{Comparison of mean $\pm$ standard deviation of FID, FD$\text{DINOv2}$, Precision/Recall, and Density/Coverage on \imagenet-$512$ for EDM2-S and EDM2-XXL across $10$ bootstraps, in each bootstrap, 50k images were subsampled from 150k generated images.
        }
        \resizebox{0.8\textwidth}{!}{%
        \begin{tabular}[t]{lrrrrrr}
            \toprule
             & \multicolumn{6}{c}{\textbf{Quality metrics}} \\ \cmidrule(lr){2-7}
            Algorithm & FID $\downarrow$ & FD$_\text{DINOv2}$ $\downarrow$
                      & Precision $\uparrow$ & Recall $\uparrow$
                      & Density $\uparrow$ & Coverage $\uparrow$ \\
            \midrule
            \textbf{EDM2-S} \\
            \cmidrule(lr){1-7}
            \CFG                     & 2.29 $\pm$ 0.026 & 88.69 $\pm$ 0.433  & 0.61 $\pm$ 0.003 & 0.57 $\pm$ 0.003 & 0.58 $\pm$ 0.002 & 0.54 $\pm$ 0.001 \\
            \limitedCFG              &  1.71 $\pm$ 0.024 & 80.76 $\pm$ 0.533 &  0.62 $\pm$ 0.002 &  0.61 $\pm$ 0.001 &  0.58 $\pm$ 0.003 &  0.56 $\pm$ 0.003 \\
            \CFGPP               & 2.89 $\pm$ 0.019 & 95.52 $\pm$ 0.266  & 0.60 $\pm$ 0.001 & 0.54 $\pm$ 0.002 & 0.57 $\pm$ 0.003 & 0.52 $\pm$ 0.002 \\
            \algoname               &  1.77 $\pm$ 0.021 & 75.43 $\pm$ 0.292 &  0.64 $\pm$ 0.002 &  0.59 $\pm$ 0.003 &  0.62 $\pm$ 0.002 &  0.58 $\pm$ 0.001 \\
            \midrule
            \textbf{EDM2-XXL} \\
            \cmidrule(lr){1-7}
            \CFG                  &  1.81 $\pm$ 0.019 & 57.22 $\pm$ 0.421 &  0.67 $\pm$ 0.002 &  0.66 $\pm$ 0.002 &  0.70 $\pm$ 0.004 &  0.65 $\pm$ 0.003 \\
            \limitedCFG              &  1.49 $\pm$ 0.026 & 39.97 $\pm$ 0.333 &  0.70 $\pm$ 0.002 &  0.68 $\pm$ 0.002 &  0.77 $\pm$ 0.003 &  0.70 $\pm$ 0.001 \\
            \CFGPP                  & 2.30 $\pm$ 0.022 & 65.32 $\pm$ 0.511 & 0.66 $\pm$ 0.004 & 0.63 $\pm$ 0.003 & 0.69 $\pm$ 0.002 & 0.62 $\pm$ 0.002 \\
            \algoname               &  1.47 $\pm$ 0.017 & 42.85 $\pm$ 0.263 &  0.70 $\pm$ 0.001 &  0.68 $\pm$ 0.002 &  0.77 $\pm$ 0.003 &  0.69 $\pm$ 0.002 \\
            \bottomrule
        \end{tabular}}
\end{table}

%% file: appendix/delayed_guidance.tex
In \Cref{prop:delayed_guidance} and \Cref{cor:delayed_score} we have proposed two novels expression for the score of the tilted distribution $\cpdata{0}{\cond; \scale}{}$. Similarly to the CFG score, it can be shown that the Rényi divergence term involved in \Cref{prop:delayed_guidance} is negligible. Thus, we get the following new denoiser approximation: 
\begin{equation} 
    \label{eq:delayed-denoiser}
    \denoiser{\std}{\cond; \scale}{\bx_\std}[\mathsf{del}] \eqdef \scale \denoiser{\stdminus}{\cond}{\bx_\std} + (1 - \scale) \denoiser{\stdplus}{}{\bx_\std}
\end{equation}
where we recall that $\stdplus = \std\sqrt{(\scale-1) / \delta}$ and $\stdminus = \std \sqrt{\scale / (1 + \delta)}$. Note that for the practical implementation, $\delta$ cannot be too small since in this case we would have $\stdplus \gg \std$ and the inputs to the denoiser at noise level $\stdplus$ would be too of this distribution. 

In \Cref{table:delayed_guidance}, we report FID and FD\textsubscript{DINOv2} scores for various combinations of \((\scale, \delta)\). We observe that using a slightly reduced guidance scale in conjunction with the approximation \eqref{eq:delayed-denoiser} substantially improves FD\textsubscript{DINOv2}, albeit at the cost of a higher FID. Conversely, increasing the guidance scale to 2.3—which achieves the best FID in \Cref{tab:imagenet512_edm2_full}—leads to a notable degradation in both FID and FD\textsubscript{DINOv2} performance. We provide a qualitative assessment of \delayedcfg\ in \Cref{subsec:delayed-cfg-quali} using the EDM2-XXL model. Overall, it tends to produce sharp and diverse images, though often with less detailed or rich textures compared to \algoname.

\begin{table}
    \centering
    \captionsetup{font=small}
    \captionof{table}{Comparison of average FID, FD$_\text{DINOv2}$ on \imagenet-$512$ for EDM2-S with $50$k samples. Besides $\scale$, we use the default hyperparameters given in \Cref{table:hyperparams-algo}.}
    \resizebox{0.5\textwidth}{!}{%
    \begin{tabular}[H]{lrr}
        \toprule
        & \multicolumn{2}{c}{\textbf{Quality metrics}} \\ \cmidrule(lr){2-3}
        Algorithm & FID $\downarrow$ & FD$_\text{DINOv2}$ $\downarrow$ \\
        \midrule
        \textbf{EDM2-S} \\
        \cmidrule(lr){1-3}
        \delayedcfg\ ($\scale=2, \delta=0.85)$     & 3.63 & 56.53 \\
        \delayedcfg\ ($\scale=2, \delta=0.90)$      & 3.57 & 54.81 \\
        \delayedcfg\ ($\scale=2, \delta=0.95)$      & 3.53 & 54.13 \\
        \delayedcfg\ ($\scale=2.3, \delta=0.85)$     & 6.35 & 86.79 \\
        \delayedcfg\ ($\scale=2.3, \delta=0.90)$      & 5.51 & 74.31 \\
        \delayedcfg\ ($\scale=2.3, \delta=0.95)$      & 5.02 & 65.63 \\
        \bottomrule
    \end{tabular}}
    \label{table:delayed_guidance}
\end{table}

%% file: appendix/qualitative.tex
\section{Qualitative assessment}
\label{apdx:qualitative}
In this section we qualitatively assess \algoname. We begin by juxtaposing its outputs with those of competing methods, then vary individual parameters to observe their effect on the generated images. All figures share the default hyperparameters from \Cref{table:hyperparams-algo,table:hyperparams-competitors}; only the parameter indicated above each figure is altered.
\subsection{Comparison of \algoname, \limitedCFG, \CFG\ and \CFGPP\ with a high guidance scale}
\begin{figure}[H]
    \center
    \includegraphics[width=\textwidth]{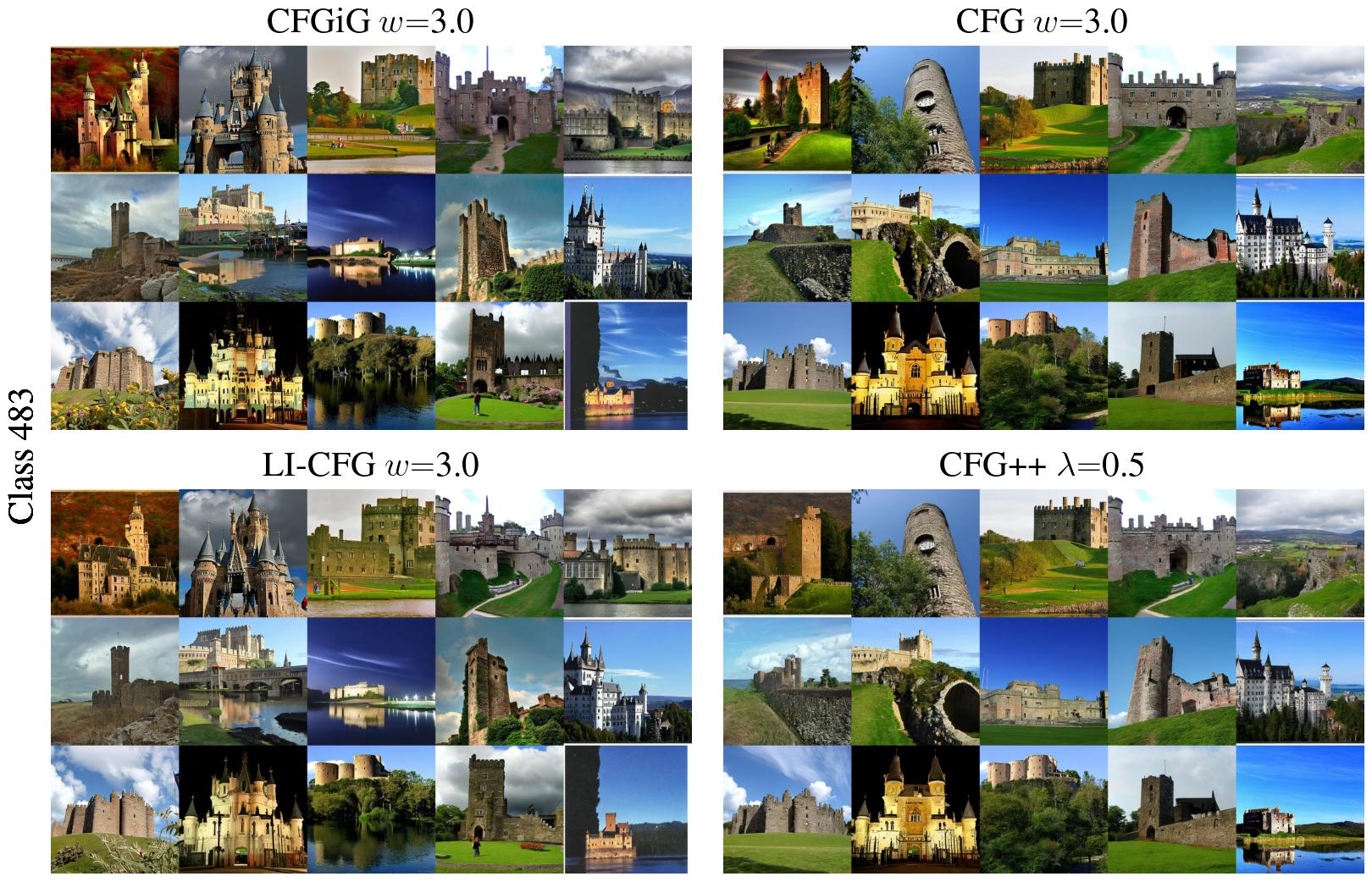}
\end{figure}
\begin{figure}[H]
    \includegraphics[width=\textwidth]{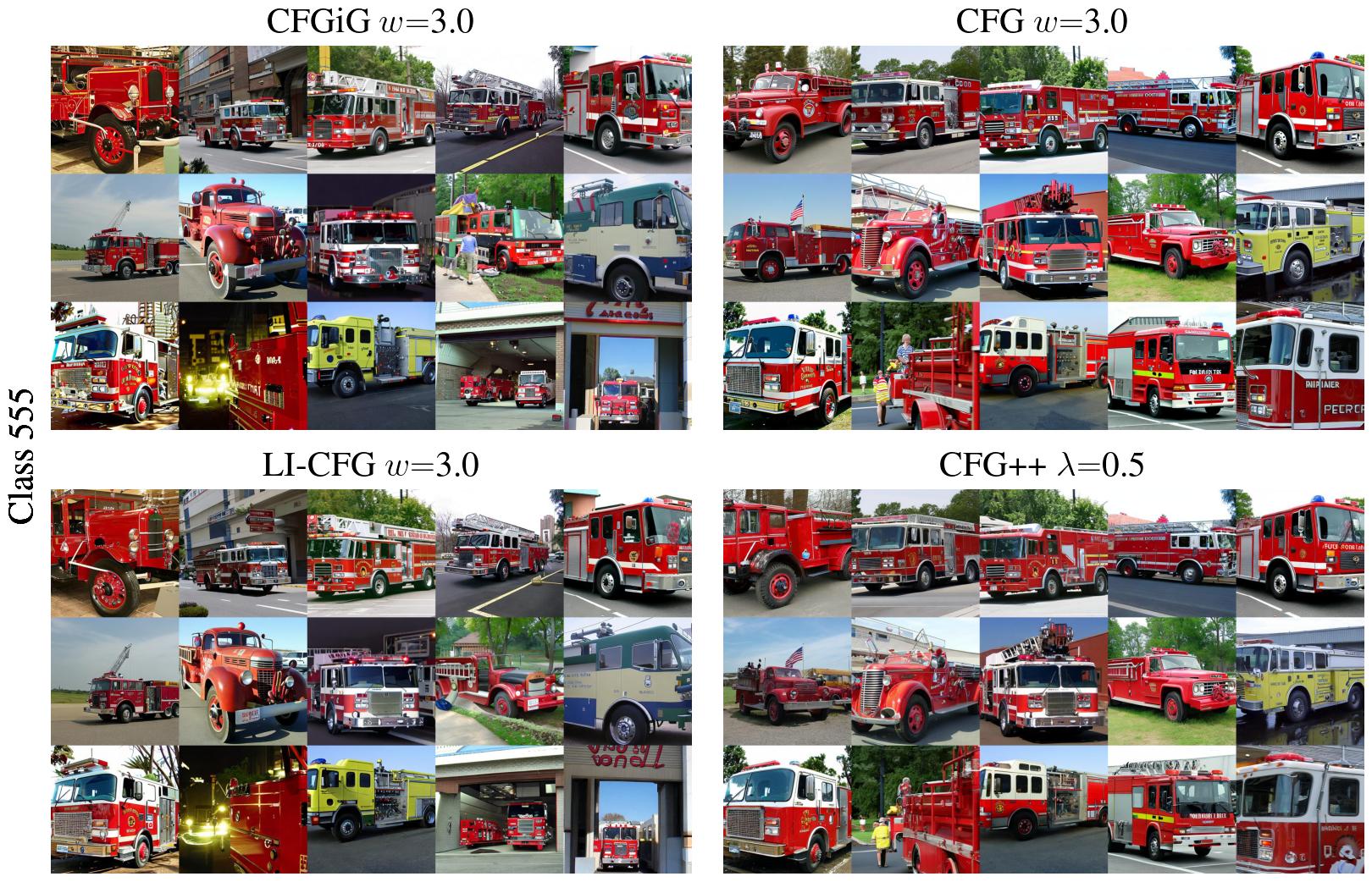}
\end{figure} 

\begin{figure}[H]
    \center
    \includegraphics[width=\textwidth]{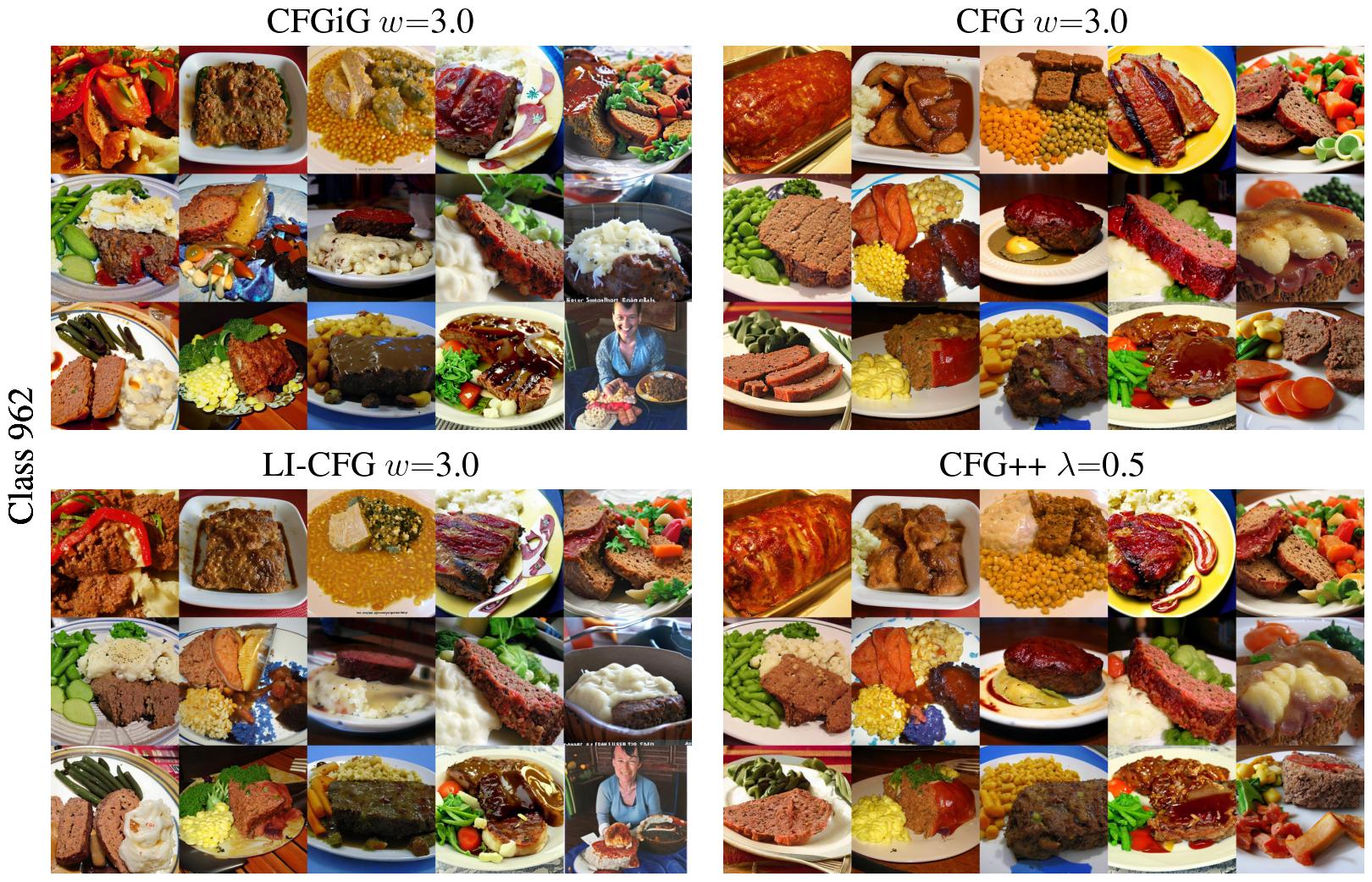}
\end{figure}
\begin{figure}[H]
    \includegraphics[width=\textwidth]{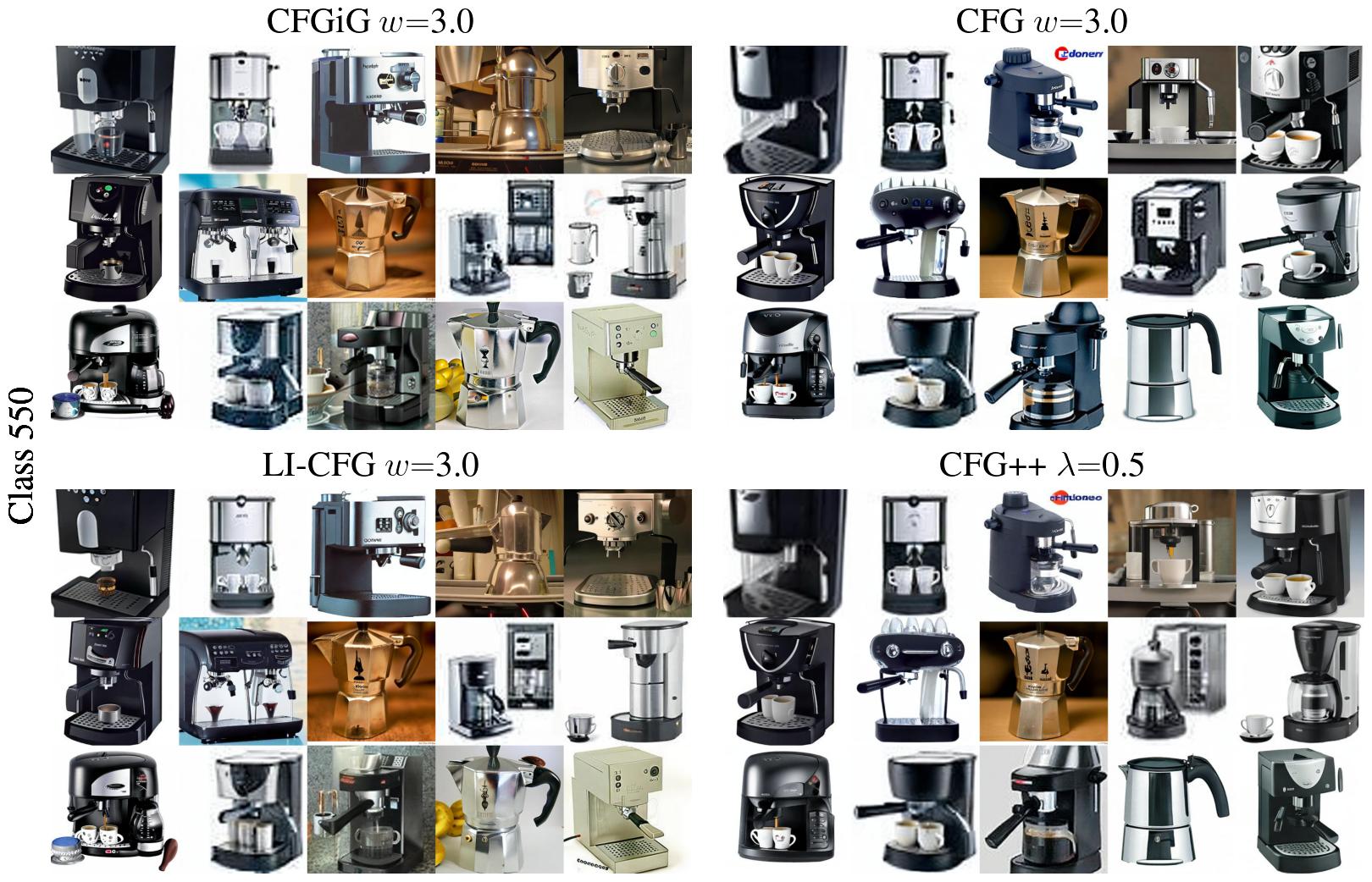}
\end{figure}    

\subsection{Varying $\scale$}
\begin{figure}[H]
    \center
    \includegraphics[width=\textwidth]{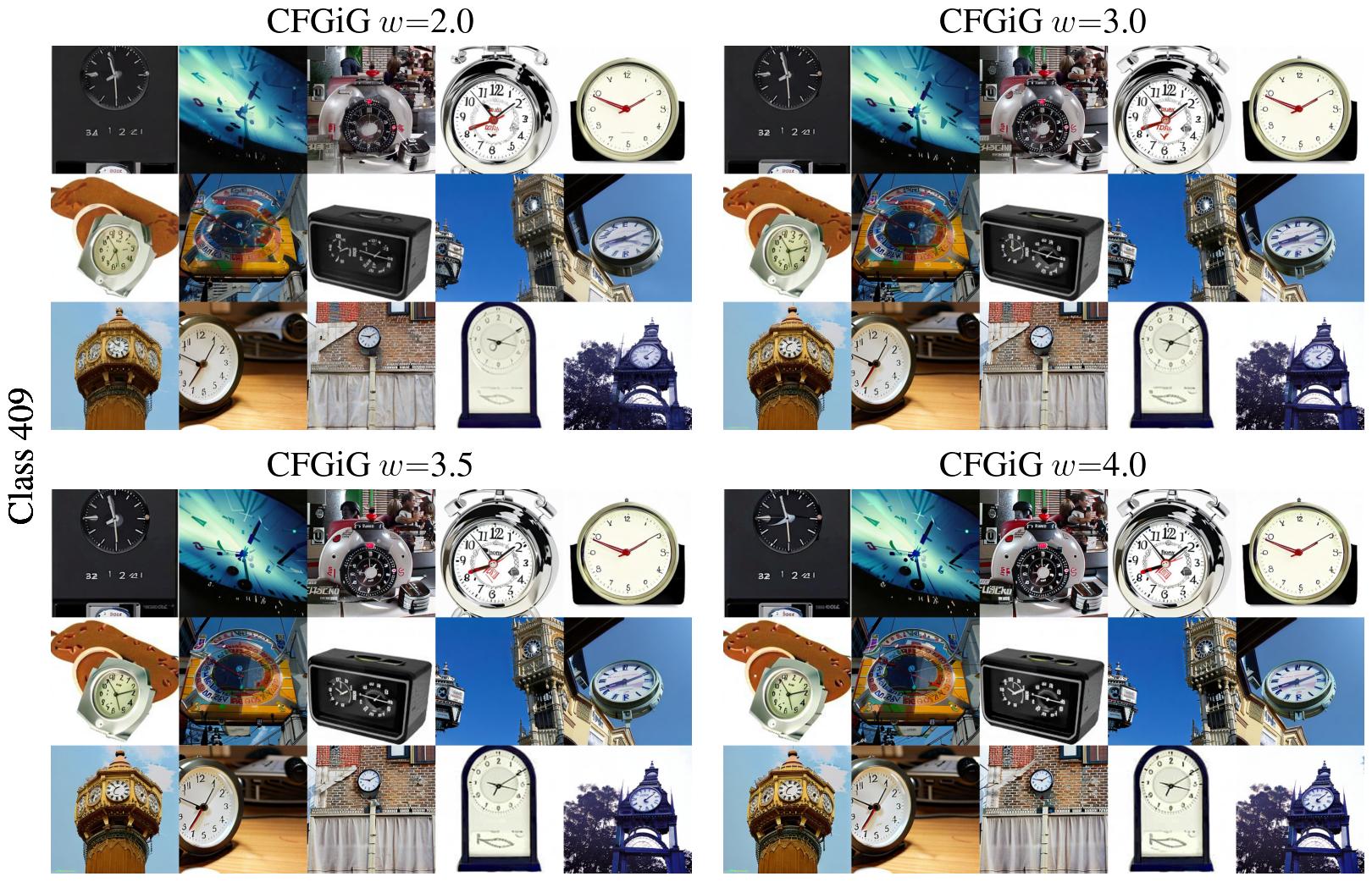}
\end{figure}
\begin{figure}[H]
    \center
    \includegraphics[width=\textwidth]{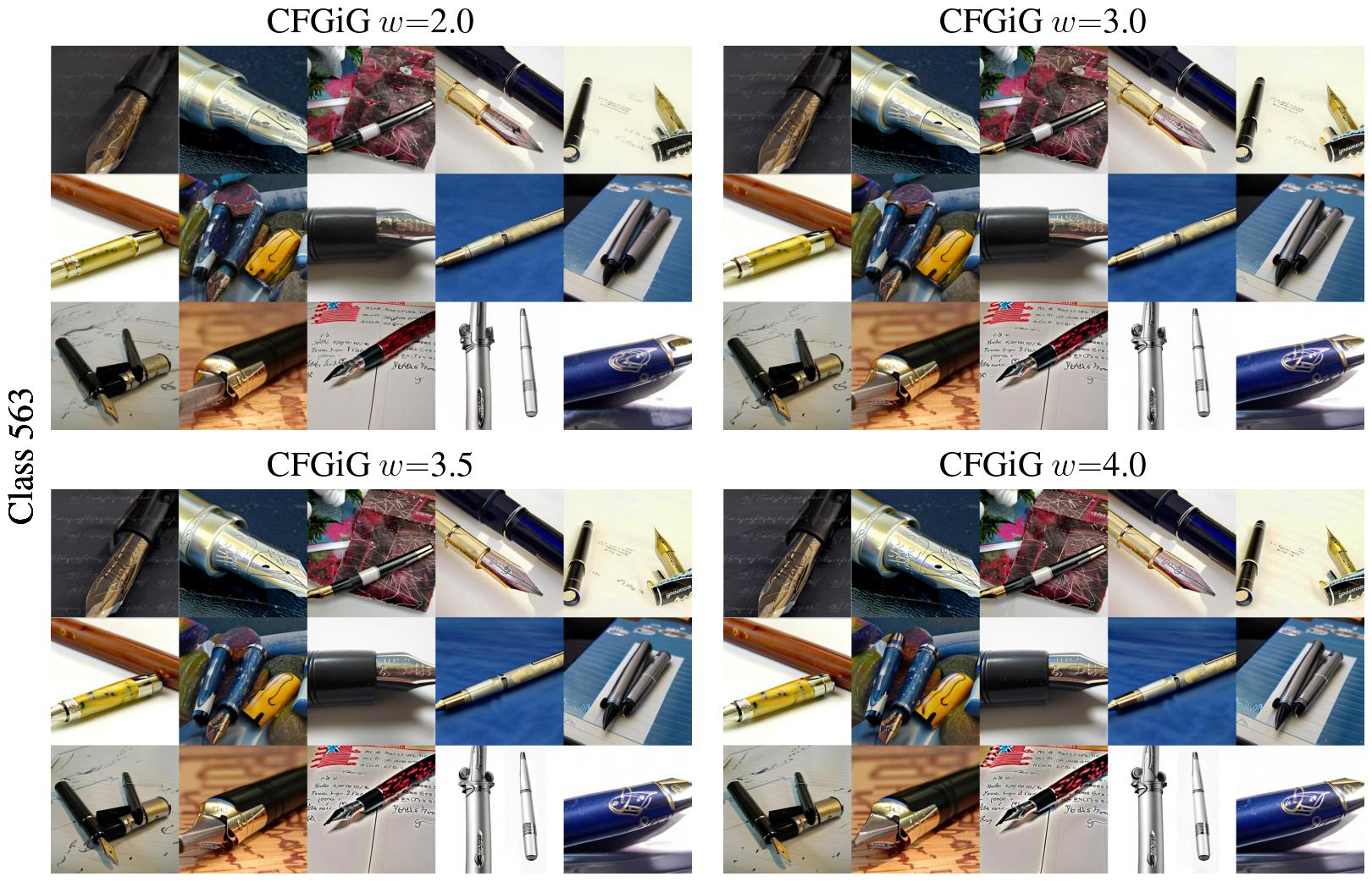}
\end{figure}
\begin{figure}[H]
    \center
    \includegraphics[width=\textwidth]{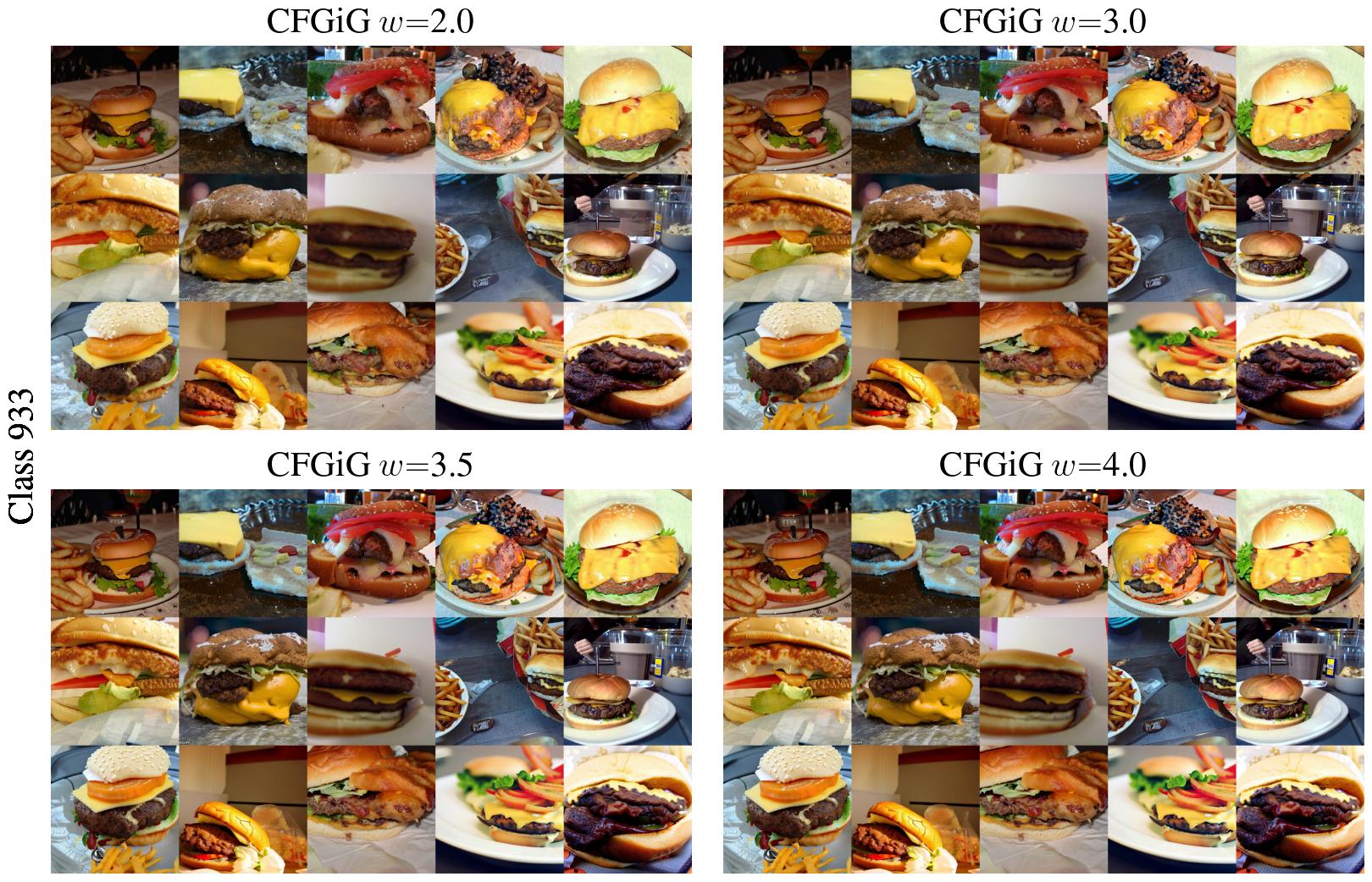}
\end{figure}  

\subsection{Varying $\stds$}
\begin{figure}[H]
    \center
    \includegraphics[width=\textwidth]{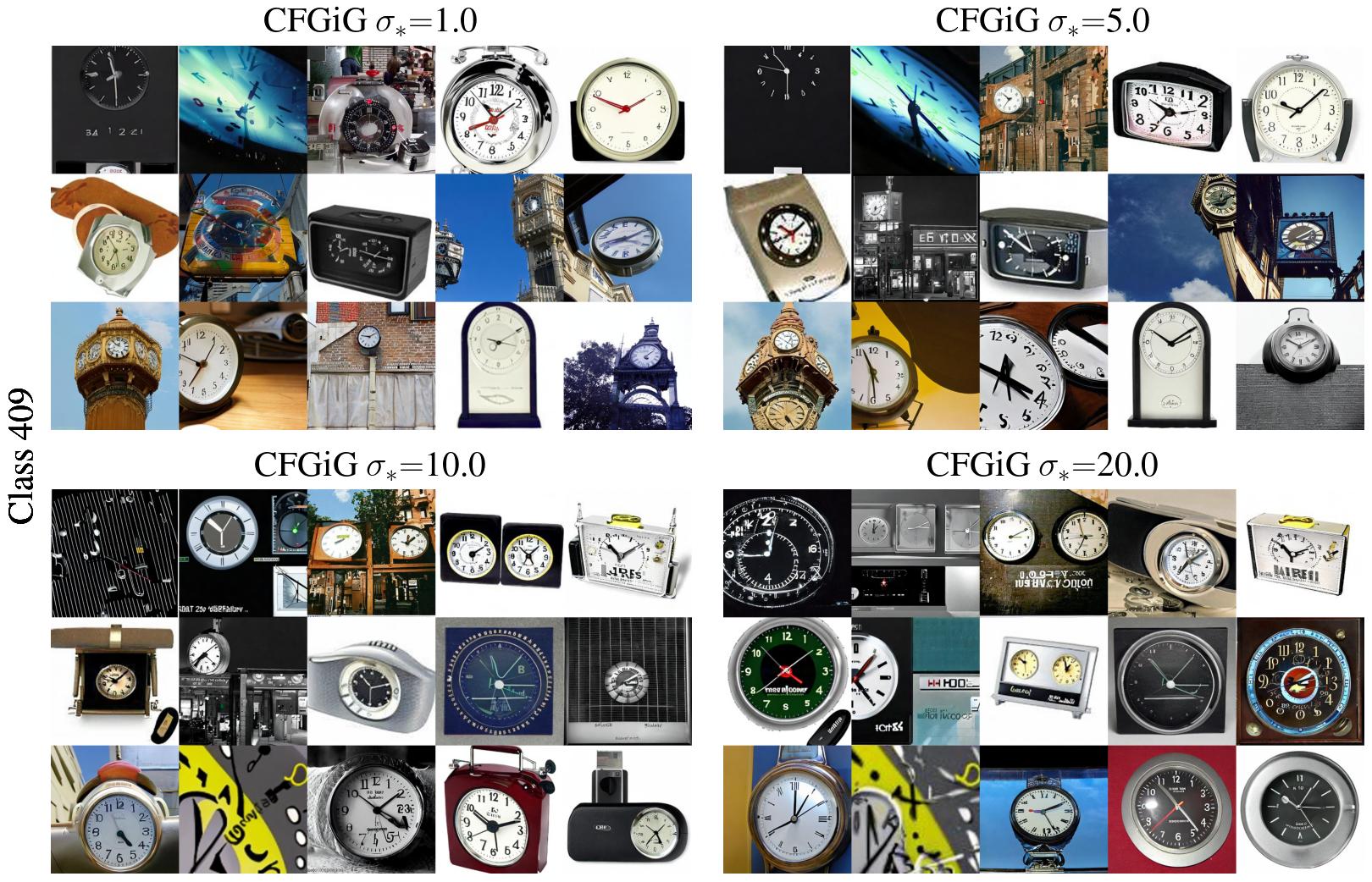}
\end{figure}
\begin{figure}[H]
    \center
    \includegraphics[width=\textwidth]{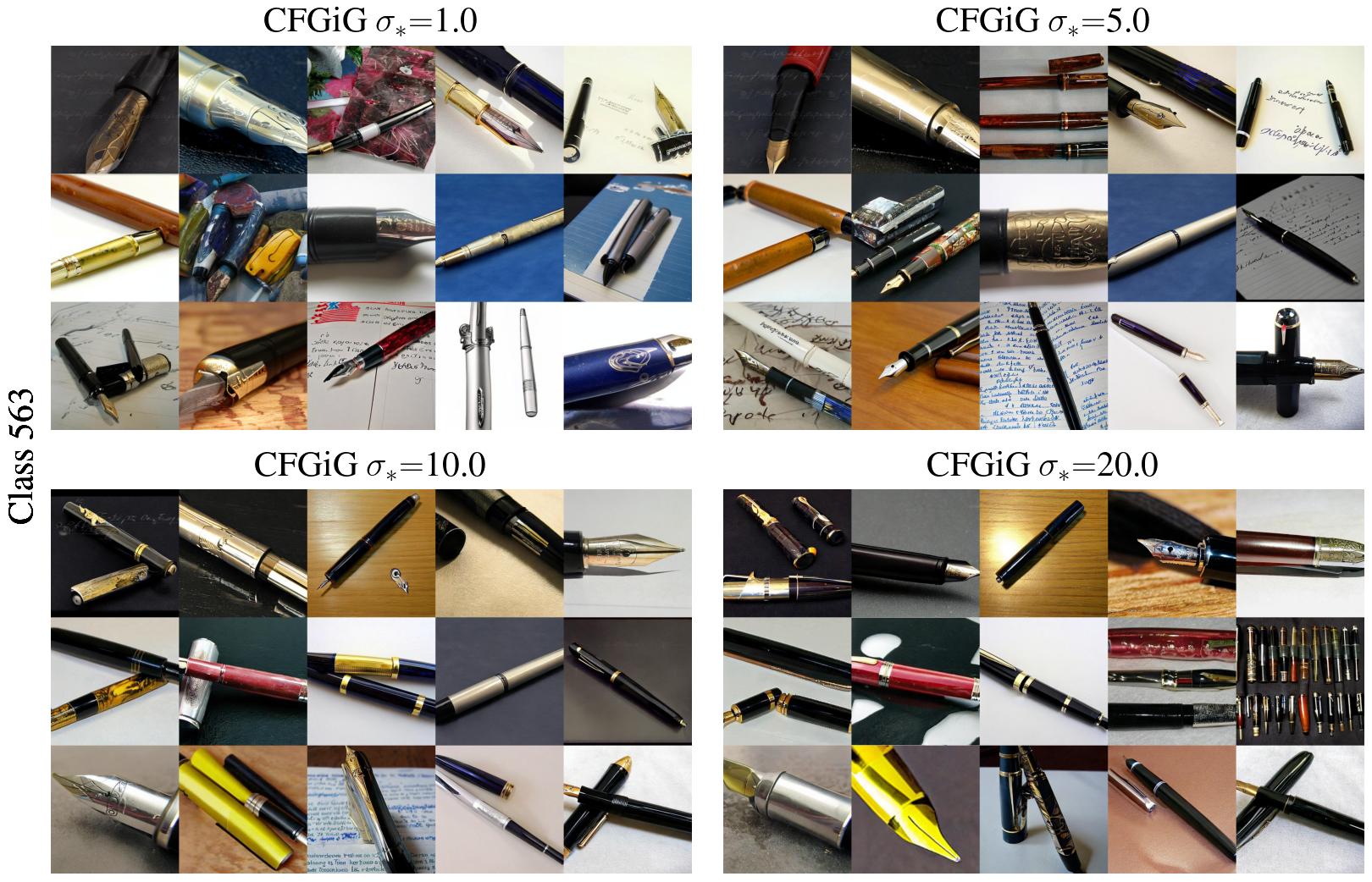}
\end{figure}
\begin{figure}[H]
    \center
    \includegraphics[width=\textwidth]{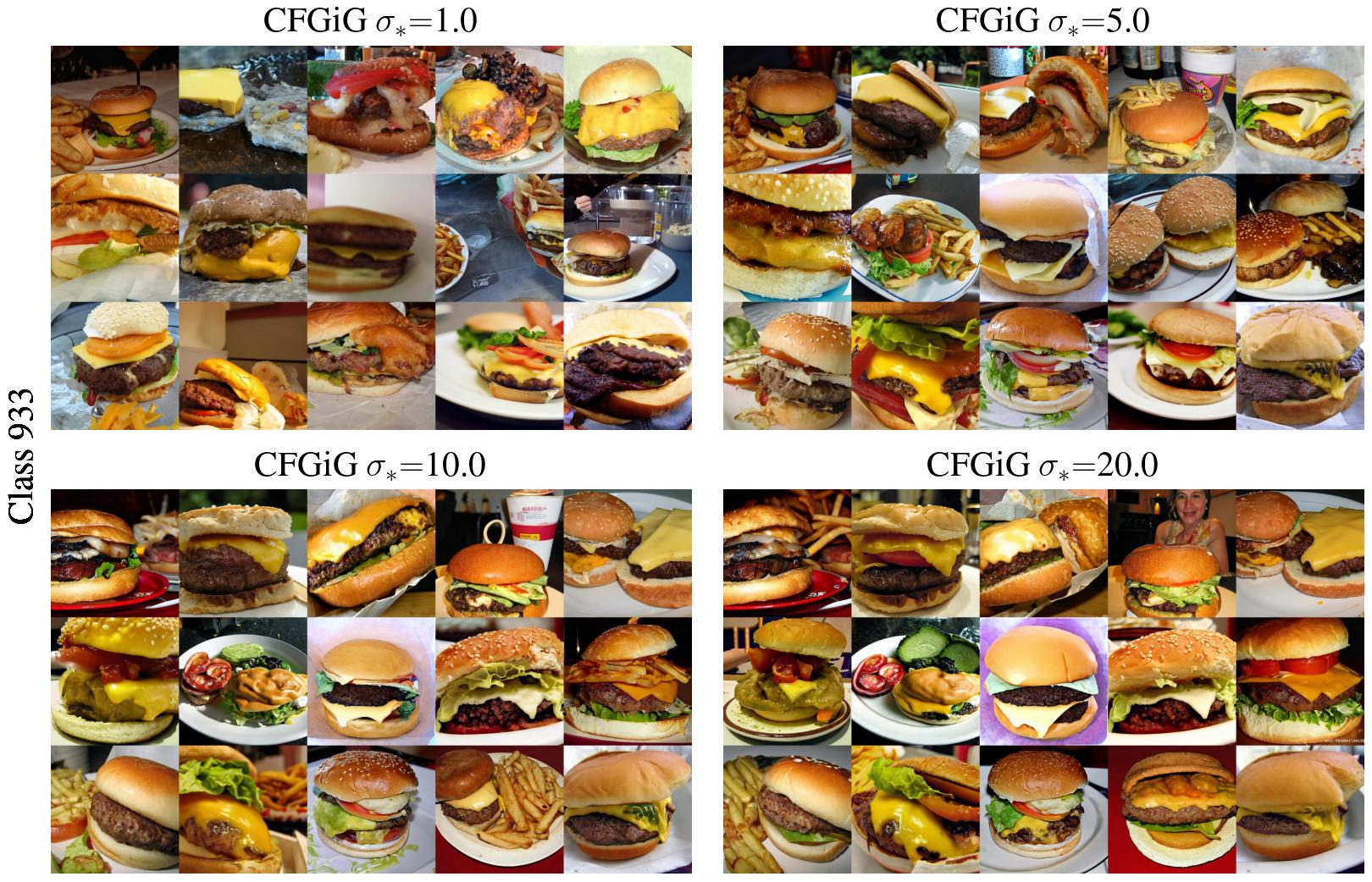}
\end{figure}  

\subsection{Varying $R$}
\begin{figure}[H]
    \center
    \includegraphics[width=\textwidth]{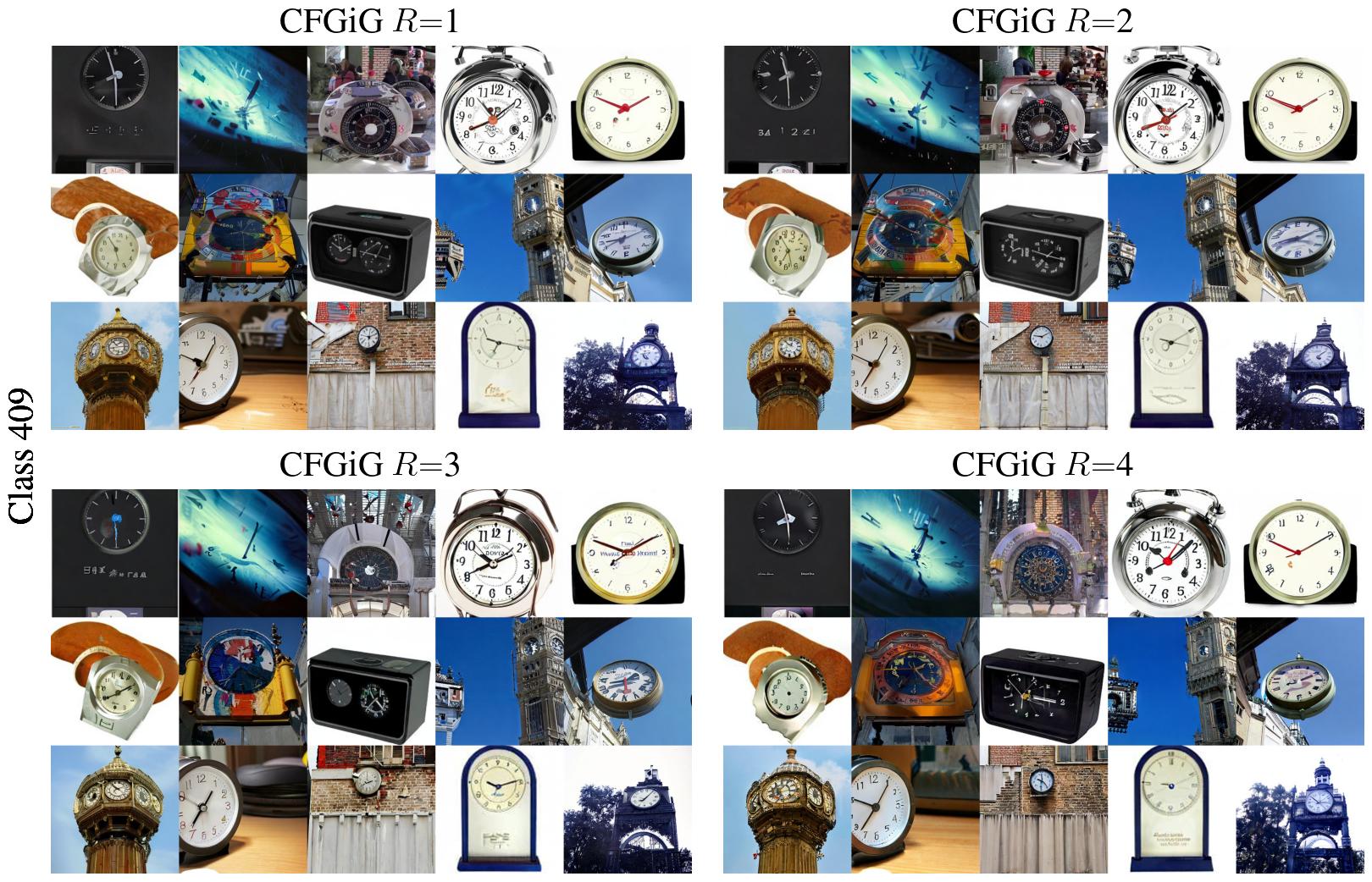}
\end{figure}
\begin{figure}[H]
    \center
    \includegraphics[width=\textwidth]{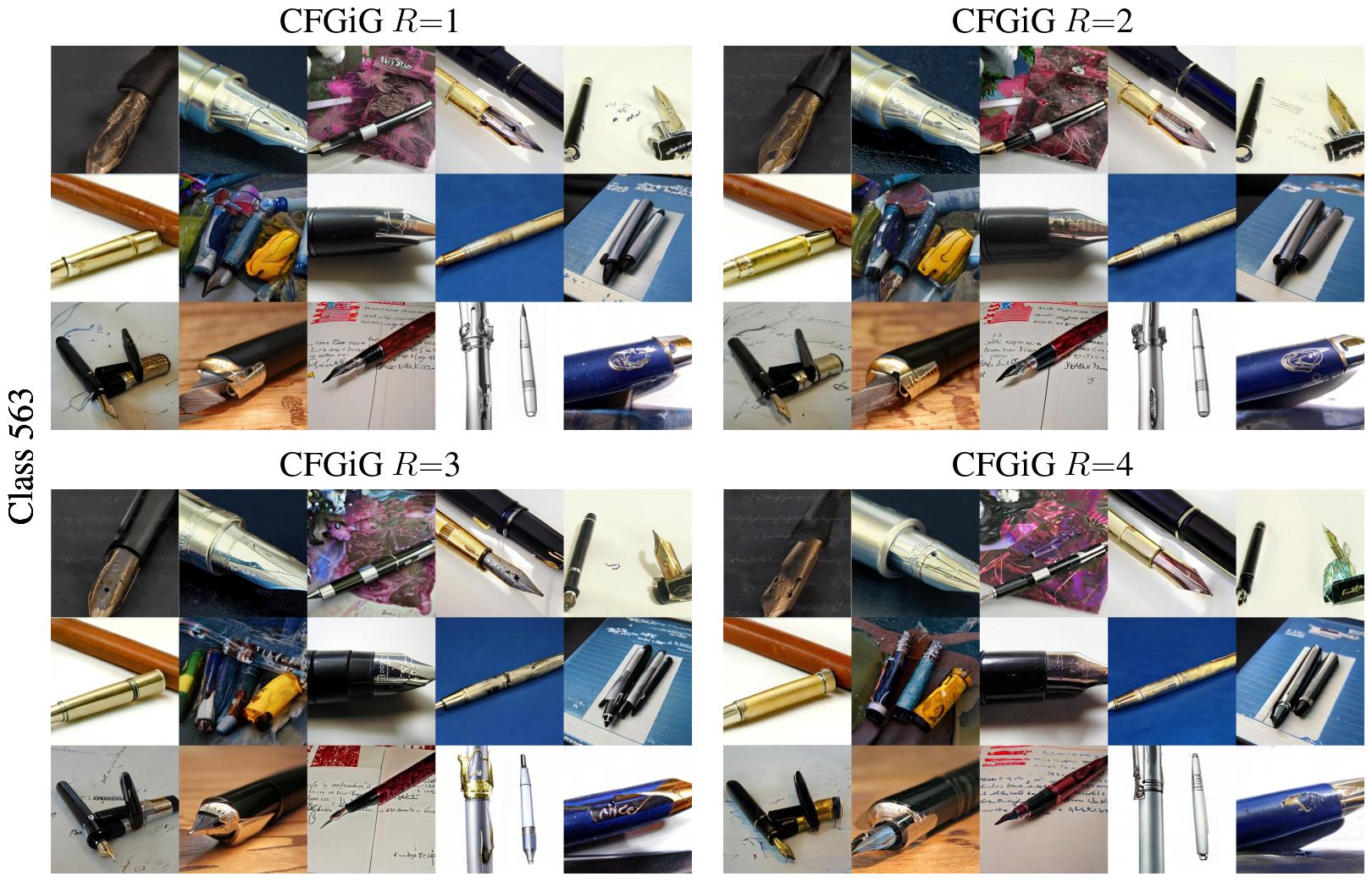}
\end{figure}
\begin{figure}[H]
    \center
    \includegraphics[width=\textwidth]{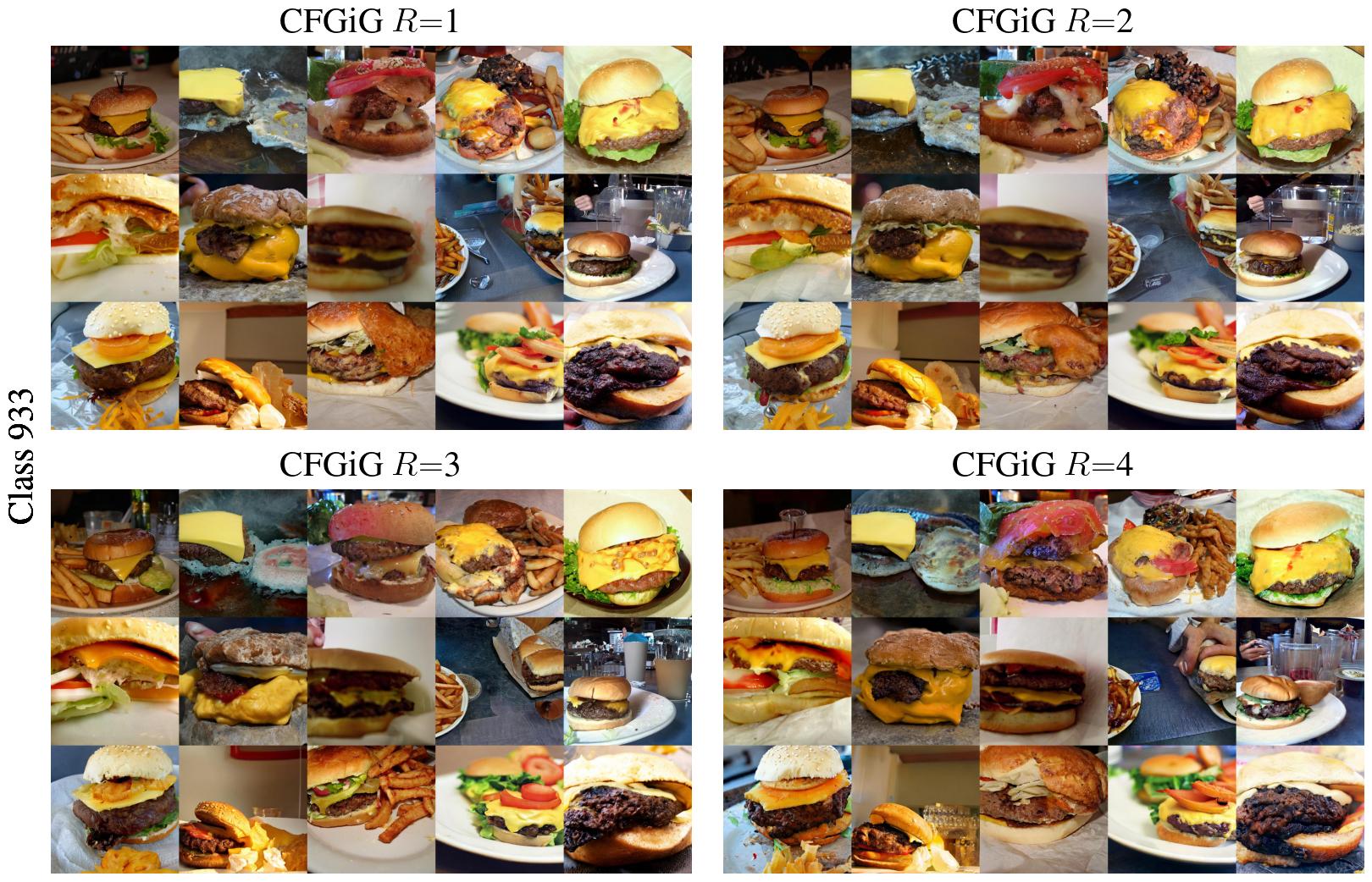}
\end{figure}  

\subsection{Varying $\delta$ in \delayedcfg}
\label{subsec:delayed-cfg-quali}
\begin{figure}[H]
    \center
    \includegraphics[width=\textwidth]{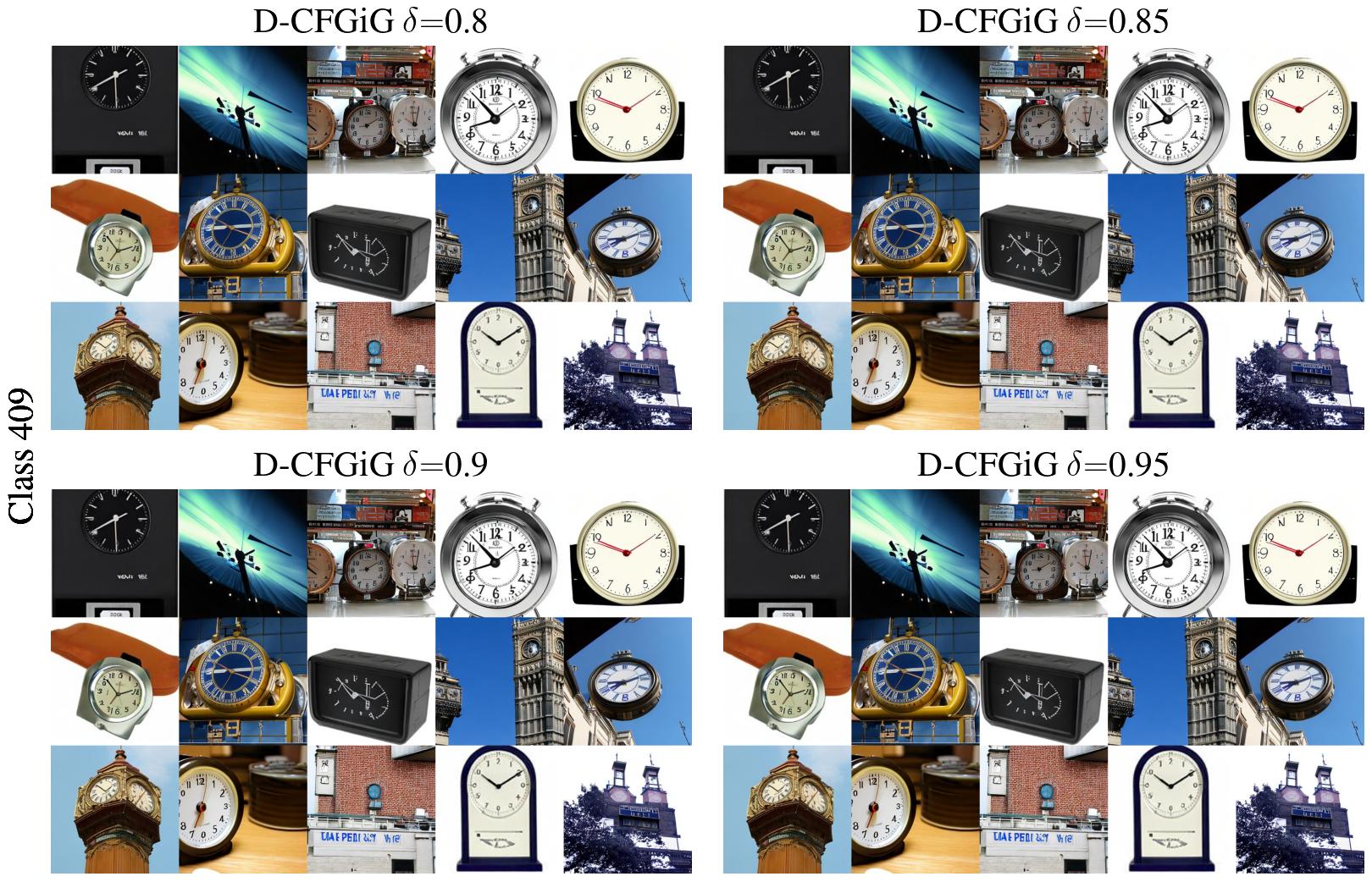}
\end{figure}
\begin{figure}[H]
    \center
    \includegraphics[width=\textwidth]{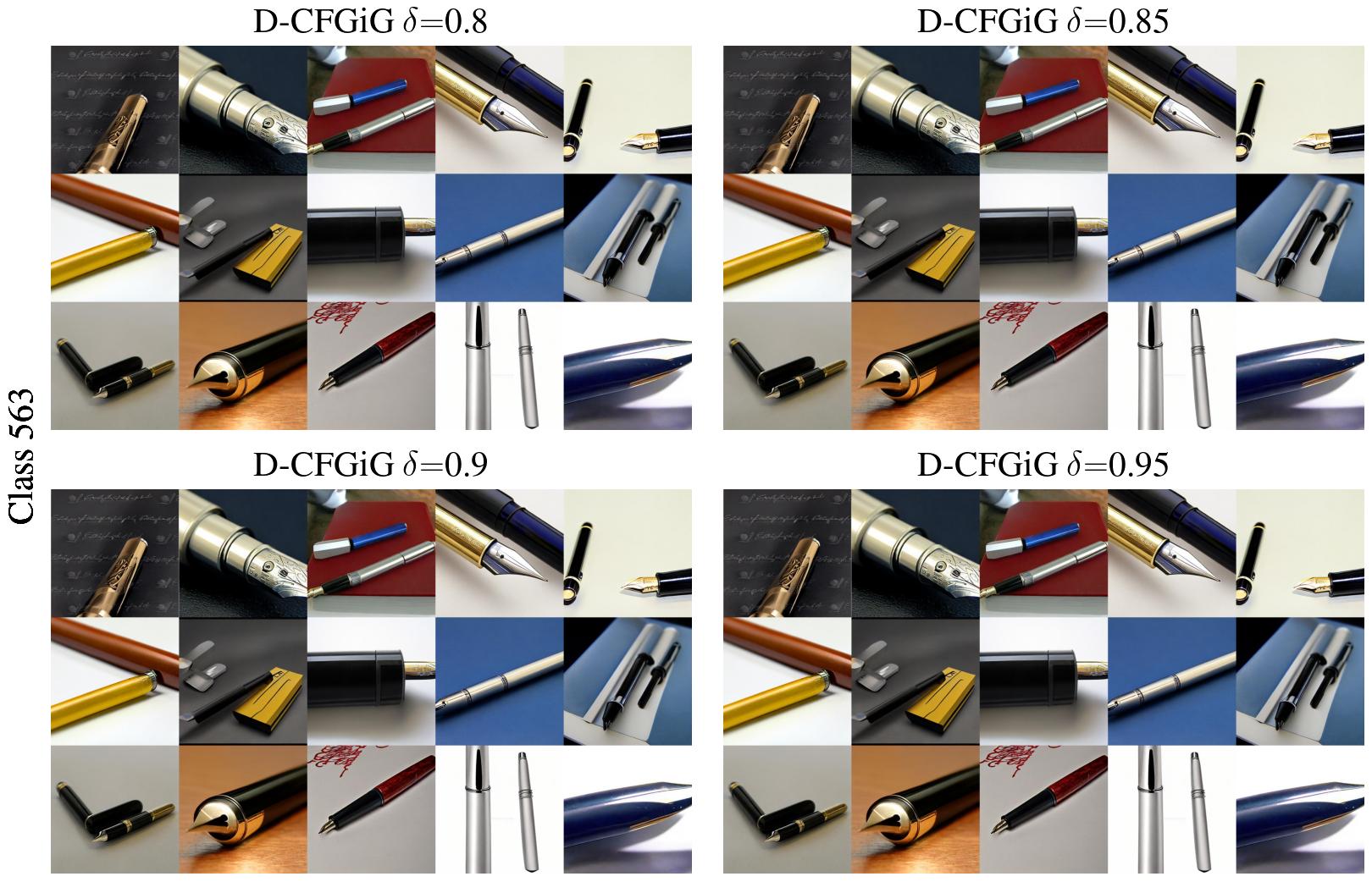}
\end{figure}
\begin{figure}[H]
    \center
    \includegraphics[width=\textwidth]{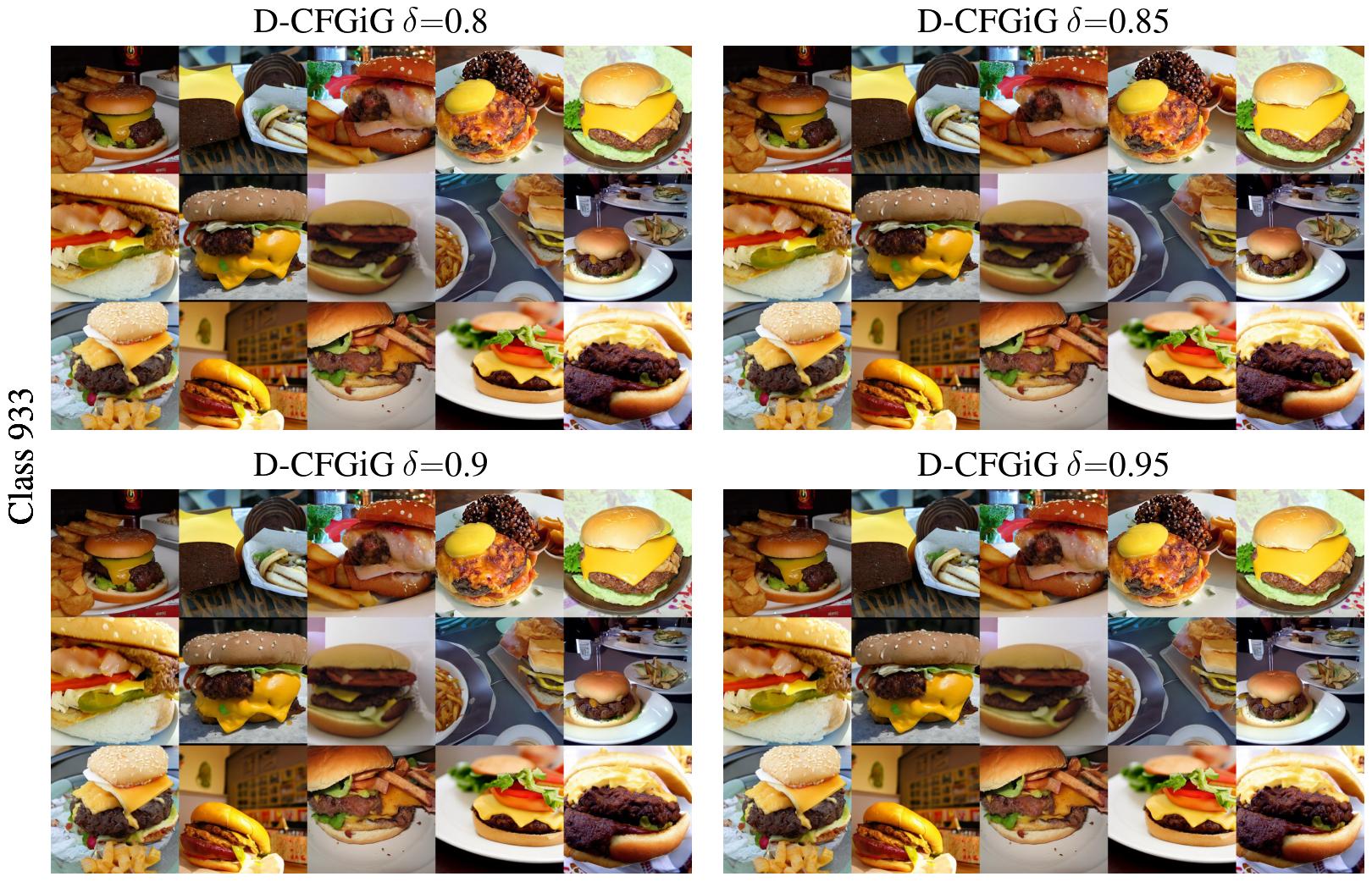}
\end{figure}

%% file: main.bbl
\begin{thebibliography}{10}

\bibitem{blattmann2023align}
Andreas Blattmann, Robin Rombach, Huan Ling, Tim Dockhorn, Seung~Wook Kim,
  Sanja Fidler, and Karsten Kreis.
\newblock Align your latents: High-resolution video synthesis with latent
  diffusion models.
\newblock In {\em Proceedings of the IEEE/CVF conference on computer vision and
  pattern recognition}, pages 22563--22575, 2023.

\bibitem{bradley2024classifier}
Arwen Bradley and Preetum Nakkiran.
\newblock Classifier-free guidance is a predictor-corrector.
\newblock {\em arXiv preprint arXiv:2408.09000}, 2024.

\bibitem{chang2023muse}
Huiwen Chang, Han Zhang, Jarred Barber, Aaron Maschinot, Jose Lezama, Lu~Jiang,
  Ming-Hsuan Yang, Kevin~Patrick Murphy, William~T Freeman, Michael Rubinstein,
  et~al.
\newblock Muse: Text-to-image generation via masked generative transformers.
\newblock In {\em International Conference on Machine Learning}, pages
  4055--4075. PMLR, 2023.

\bibitem{chidambaram2024what}
Muthu Chidambaram, Khashayar Gatmiry, Sitan Chen, Holden Lee, and Jianfeng Lu.
\newblock What does guidance do? a fine-grained analysis in a simple setting.
\newblock In {\em The Thirty-eighth Annual Conference on Neural Information
  Processing Systems}, 2024.

\bibitem{chung2025cfg}
Hyungjin Chung, Jeongsol Kim, Geon~Yeong Park, Hyelin Nam, and Jong~Chul Ye.
\newblock {CFG}++: Manifold-constrained classifier free guidance for diffusion
  models.
\newblock In {\em The Thirteenth International Conference on Learning
  Representations}, 2025.

\bibitem{dieleman2022guidance}
Sander Dieleman.
\newblock Guidance: a cheat code for diffusion models, 2022.

\bibitem{doucet2001sequential}
Arnaud Doucet, Nando De~Freitas, Neil~James Gordon, et~al.
\newblock {\em Sequential {M}onte {C}arlo methods in practice}, volume~1.
\newblock Springer, 2001.

\bibitem{gelfand2000gibbs}
Alan~E Gelfand.
\newblock Gibbs sampling.
\newblock {\em Journal of the American statistical Association},
  95(452):1300--1304, 2000.

\bibitem{hershey2017cnn}
Shawn Hershey, Sourish Chaudhuri, Daniel~PW Ellis, Jort~F Gemmeke, Aren Jansen,
  R~Channing Moore, Manoj Plakal, Devin Platt, Rif~A Saurous, Bryan Seybold,
  et~al.
\newblock Cnn architectures for large-scale audio classification.
\newblock In {\em 2017 ieee international conference on acoustics, speech and
  signal processing (icassp)}, pages 131--135. IEEE, 2017.

\bibitem{heusel2017fid}
Martin Heusel, Hubert Ramsauer, Thomas Unterthiner, Bernhard Nessler, and Sepp
  Hochreiter.
\newblock Gans trained by a two time-scale update rule converge to a local nash
  equilibrium.
\newblock {\em Advances in neural information processing systems}, 30, 2017.

\bibitem{ho2020denoising}
Jonathan Ho, Ajay Jain, and Pieter Abbeel.
\newblock Denoising diffusion probabilistic models.
\newblock {\em Advances in Neural Information Processing Systems},
  33:6840--6851, 2020.

\bibitem{ho2022classifier}
Jonathan Ho and Tim Salimans.
\newblock Classifier-free diffusion guidance.
\newblock {\em arXiv preprint arXiv:2207.12598}, 2022.

\bibitem{karras2022elucidating}
Tero Karras, Miika Aittala, Timo Aila, and Samuli Laine.
\newblock Elucidating the design space of diffusion-based generative models.
\newblock {\em Advances in Neural Information Processing Systems},
  35:26565--26577, 2022.

\bibitem{karras2024guiding}
Tero Karras, Miika Aittala, Tuomas Kynk{\"a}{\"a}nniemi, Jaakko Lehtinen, Timo
  Aila, and Samuli Laine.
\newblock Guiding a diffusion model with a bad version of itself.
\newblock {\em Advances in Neural Information Processing Systems},
  37:52996--53021, 2024.

\bibitem{karras2024edm2}
Tero Karras, Miika Aittala, Jaakko Lehtinen, Janne Hellsten, Timo Aila, and
  Samuli Laine.
\newblock Analyzing and improving the training dynamics of diffusion models.
\newblock In {\em Proceedings of the IEEE/CVF Conference on Computer Vision and
  Pattern Recognition}, pages 24174--24184, 2024.

\bibitem{kim2019audiocaps}
Chris~Dongjoo Kim, Byeongchang Kim, Hyunmin Lee, and Gunhee Kim.
\newblock Audiocaps: Generating captions for audios in the wild.
\newblock In {\em Proceedings of the 2019 Conference of the North American
  Chapter of the Association for Computational Linguistics: Human Language
  Technologies, Volume 1 (Long and Short Papers)}, pages 119--132, 2019.

\bibitem{kong2020panns}
Qiuqiang Kong, Yin Cao, Turab Iqbal, Yuxuan Wang, Wenwu Wang, and Mark~D
  Plumbley.
\newblock Panns: Large-scale pretrained audio neural networks for audio pattern
  recognition.
\newblock {\em IEEE/ACM Transactions on Audio, Speech, and Language
  Processing}, 28:2880--2894, 2020.

\bibitem{kong2020diffwave}
Zhifeng Kong, Wei Ping, Jiaji Huang, Kexin Zhao, and Bryan Catanzaro.
\newblock Diffwave: A versatile diffusion model for audio synthesis.
\newblock {\em arXiv preprint arXiv:2009.09761}, 2020.

\bibitem{kynkanniemi2024applying}
Tuomas Kynk{\"a}{\"a}nniemi, Miika Aittala, Tero Karras, Samuli Laine, Timo
  Aila, and Jaakko Lehtinen.
\newblock Applying guidance in a limited interval improves sample and
  distribution quality in diffusion models.
\newblock In {\em The Thirty-eighth Annual Conference on Neural Information
  Processing Systems}, 2024.

\bibitem{lee2025debiasing}
Cheuk~Kit Lee, Paul Jeha, Jes Frellsen, Pietro Lio, Michael~Samuel Albergo, and
  Francisco Vargas.
\newblock Debiasing guidance for discrete diffusion with sequential {M}onte
  {C}arlo.
\newblock {\em arXiv preprint arXiv:2502.06079}, 2025.

\bibitem{li2024self}
Tiancheng Li, Weijian Luo, Zhiyang Chen, Liyuan Ma, and Guo-Jun Qi.
\newblock Self-guidance: Boosting flow and diffusion generation on their own.
\newblock {\em arXiv preprint arXiv:2412.05827}, 2024.

\bibitem{liu2024audioldm2}
Haohe Liu, Yi~Yuan, Xubo Liu, Xinhao Mei, Qiuqiang Kong, Qiao Tian, Yuping
  Wang, Wenwu Wang, Yuxuan Wang, and Mark~D Plumbley.
\newblock Audioldm 2: Learning holistic audio generation with self-supervised
  pretraining.
\newblock {\em IEEE/ACM Transactions on Audio, Speech, and Language
  Processing}, 2024.

\bibitem{meng2021sdedit}
Chenlin Meng, Yang Song, Jiaming Song, Jiajun Wu, Jun-Yan Zhu, and Stefano
  Ermon.
\newblock Sdedit: Image synthesis and editing with stochastic differential
  equations.
\newblock {\em arXiv preprint arXiv:2108.01073}, 2021.

\bibitem{meyn2012markov}
Sean~P Meyn and Richard~L Tweedie.
\newblock {\em {M}arkov chains and stochastic stability}.
\newblock Springer Science \& Business Media, 2012.

\bibitem{naeem2020prdc}
Muhammad~Ferjad Naeem, Seong~Joon Oh, Youngjung Uh, Yunjey Choi, and Jaejun
  Yoo.
\newblock Reliable fidelity and diversity metrics for generative models.
\newblock In {\em International conference on machine learning}, pages
  7176--7185. PMLR, 2020.

\bibitem{oquab2023dinov2}
Maxime Oquab, Timoth{\'e}e Darcet, Th{\'e}o Moutakanni, Huy Vo, Marc
  Szafraniec, Vasil Khalidov, Pierre Fernandez, Daniel Haziza, Francisco Massa,
  Alaaeldin El-Nouby, et~al.
\newblock Dinov2: Learning robust visual features without supervision.
\newblock {\em arXiv preprint arXiv:2304.07193}, 2023.

\bibitem{pavasovic2025understanding}
Krunoslav~Lehman Pavasovic, Jakob Verbeek, Giulio Biroli, and Marc Mezard.
\newblock Understanding classifier-free guidance: High-dimensional theory and
  non-linear generalizations.
\newblock {\em arXiv preprint arXiv:2502.07849}, 2025.

\bibitem{podell2023sdxl}
Dustin Podell, Zion English, Kyle Lacey, Andreas Blattmann, Tim Dockhorn, Jonas
  M{\"u}ller, Joe Penna, and Robin Rombach.
\newblock Sdxl: Improving latent diffusion models for high-resolution image
  synthesis.
\newblock {\em arXiv preprint arXiv:2307.01952}, 2023.

\bibitem{radford2019language}
Alec Radford, Jeffrey Wu, Rewon Child, David Luan, Dario Amodei, Ilya
  Sutskever, et~al.
\newblock Language models are unsupervised multitask learners.
\newblock {\em OpenAI blog}, 1(8):9, 2019.

\bibitem{robbins1956empirical}
Herbert~E. Robbins.
\newblock An empirical bayes approach to statistics.
\newblock In {\em Proceedings of the Third Berkeley Symposium on Mathematical
  Statistics and Probability, Volume 1: Contributions to the Theory of
  Statistics}, 1956.

\bibitem{rombach2022high}
Robin Rombach, Andreas Blattmann, Dominik Lorenz, Patrick Esser, and Bj{\"o}rn
  Ommer.
\newblock High-resolution image synthesis with latent diffusion models.
\newblock In {\em Proceedings of the IEEE/CVF Conference on Computer Vision and
  Pattern Recognition}, pages 10684--10695, 2022.

\bibitem{sadat2024cads}
Seyedmorteza Sadat, Jakob Buhmann, Derek Bradley, Otmar Hilliges, and Romann~M.
  Weber.
\newblock {CADS}: Unleashing the diversity of diffusion models through
  condition-annealed sampling.
\newblock In {\em The Twelfth International Conference on Learning
  Representations}, 2024.

\bibitem{sadat2025no}
Seyedmorteza Sadat, Manuel Kansy, Otmar Hilliges, and Romann~M. Weber.
\newblock No training, no problem: Rethinking classifier-free guidance for
  diffusion models.
\newblock In {\em The Thirteenth International Conference on Learning
  Representations}, 2025.

\bibitem{santambrogio2015optimal}
Filippo Santambrogio.
\newblock {\em Optimal transport for applied mathematicians}, volume~87.
\newblock Springer, 2015.

\bibitem{skreta2025feynman}
Marta Skreta, Tara Akhound-Sadegh, Viktor Ohanesian, Roberto Bondesan, Al{\'a}n
  Aspuru-Guzik, Arnaud Doucet, Rob Brekelmans, Alexander Tong, and Kirill
  Neklyudov.
\newblock Feynman-kac correctors in diffusion: Annealing, guidance, and product
  of experts.
\newblock {\em arXiv preprint arXiv:2503.02819}, 2025.

\bibitem{sohl2015deep}
Jascha Sohl-Dickstein, Eric Weiss, Niru Maheswaranathan, and Surya Ganguli.
\newblock Deep unsupervised learning using nonequilibrium thermodynamics.
\newblock In {\em International Conference on Machine Learning}, pages
  2256--2265. PMLR, 2015.

\bibitem{song2021ddim}
Jiaming Song, Chenlin Meng, and Stefano Ermon.
\newblock Denoising diffusion implicit models.
\newblock In {\em International Conference on Learning Representations}, 2021.

\bibitem{song2019generative}
Yang Song and Stefano Ermon.
\newblock Generative modeling by estimating gradients of the data distribution.
\newblock {\em Advances in neural information processing systems}, 32, 2019.

\bibitem{song2021score}
Yang Song, Jascha Sohl-Dickstein, Diederik~P Kingma, Abhishek Kumar, Stefano
  Ermon, and Ben Poole.
\newblock Score-based generative modeling through stochastic differential
  equations.
\newblock In {\em International Conference on Learning Representations}, 2021.

\bibitem{stein2023fd_dinvo2}
George Stein, Jesse Cresswell, Rasa Hosseinzadeh, Yi~Sui, Brendan Ross,
  Valentin Villecroze, Zhaoyan Liu, Anthony~L Caterini, Eric Taylor, and
  Gabriel Loaiza-Ganem.
\newblock Exposing flaws of generative model evaluation metrics and their
  unfair treatment of diffusion models.
\newblock {\em Advances in Neural Information Processing Systems},
  36:3732--3784, 2023.

\bibitem{szegedy2016inceptionv3}
Christian Szegedy, Vincent Vanhoucke, Sergey Ioffe, Jon Shlens, and Zbigniew
  Wojna.
\newblock Rethinking the inception architecture for computer vision.
\newblock In {\em Proceedings of the IEEE conference on computer vision and
  pattern recognition}, pages 2818--2826, 2016.

\bibitem{van2014renyi}
Tim Van~Erven and Peter Harremos.
\newblock R{\'e}nyi divergence and kullback-leibler divergence.
\newblock {\em IEEE Transactions on Information Theory}, 60(7):3797--3820,
  2014.

\bibitem{xianalysis}
WANG Xi, Nicolas Dufour, Nefeli Andreou, CANI Marie-Paule, Victoria~Fernandez
  Abrevaya, David Picard, and Vicky Kalogeiton.
\newblock Analysis of classifier-free guidance weight schedulers.
\newblock {\em Transactions on Machine Learning Research}.

\bibitem{xu2022masked}
Hu~Xu, Juncheng Li, Alexei Baevski, Michael Auli, Wojciech Galuba, Florian
  Metze, Christoph Feichtenhofer, et~al.
\newblock Masked autoencoders that listen.
\newblock {\em arXiv e-prints}, pages arXiv--2207, 2022.

\end{thebibliography}
